\documentclass{article}

\usepackage{arxiv}
\usepackage[final,nopatch=footnote]{microtype}
\usepackage[utf8]{inputenc} 
\usepackage[T1]{fontenc}    
\usepackage{hyperref}       
\usepackage{url}            
\usepackage{booktabs}       
\usepackage{amsfonts}       
\usepackage{nicefrac}       
\usepackage{microtype}      
\usepackage{lipsum}		
\usepackage{graphicx}
\graphicspath{{images/}}
\usepackage[numbers]{natbib}
\usepackage{eccvabbrv}
\usepackage{doi}
\usepackage{amsmath}
\usepackage{graphicx}
\usepackage{amsmath}
\usepackage{orcidlink}
\usepackage{graphicx}
\usepackage{booktabs}
\usepackage{subcaption}
\usepackage{rotating}
\usepackage{makecell}
\usepackage{graphicx}
\usepackage{booktabs}
\usepackage{siunitx}
\usepackage{geometry}
\usepackage{rotate}
\usepackage{longtable}
\usepackage{booktabs}
\usepackage{array}
\usepackage{geometry}
\usepackage{pdflscape}
\usepackage{longtable}
\usepackage{array}
\usepackage{booktabs}
\usepackage{longtable}
\usepackage{booktabs}
\usepackage{tabularx}
\usepackage{float}
\usepackage{setspace}
\usepackage{cleveref}
\usepackage{comment}

\usepackage[accsupp]{axessibility} 

\title{Assessing the Capability of YOLO- and Transformer-based Object Detectors for Real-time Weed Detection}

\author{%
\begin{minipage}[t]{0.45\textwidth}
    \centering
    {\fontsize{10}{12}\selectfont
    \href{https://orcid.org/0009-0009-9206-5349}{\includegraphics[scale=0.06]{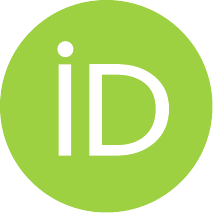}\hspace{1mm} \textbf{Alicia Allmendinger}}\footnotemark[1] \\
    \textnormal{Department of Weed Science} \\
    \textnormal{University of Hohenheim} \\
    }
\end{minipage}
\hspace{8mm} 
\begin{minipage}[t]{0.45\textwidth}
    \centering
    {\fontsize{10}{12}\selectfont
    \href{https://orcid.org/0000-0002-8873-1638}{\includegraphics[scale=0.06]{orcid.pdf}\hspace{1mm} \textbf{Ahmet O\u{g}uz Salt{\i}k}} \\
    \textnormal{Department of Artificial Intelligence in \\ Agricultural Engineering \&} \\
    \textnormal{Computational Science Hub} \\
    \textnormal{University of Hohenheim} \\
    }
\end{minipage}
\\[6em]
\hspace{0mm} 
\begin{minipage}[t]{0.45\textwidth}
    \centering
    {\fontsize{10}{12}\selectfont
    \href{https://orcid.org/0000-0002-8676-1024}{\includegraphics[scale=0.06]{orcid.pdf}\hspace{1mm} \textbf{Gerassimos G. Peteinatos}} \\
    \textnormal{ELGO - “DIMITRA”} \\
    }
\end{minipage}
\hspace{8mm}
\begin{minipage}[t]{0.45\textwidth}
    \centering
    {\fontsize{10}{12}\selectfont
    \href{https://orcid.org/0000-0002-1808-9758}{\includegraphics[scale=0.06]{orcid.pdf}\hspace{1mm} \textbf{Anthony Stein}} \\
    \textnormal{Department of Artificial Intelligence in \\ Agricultural Engineering \&} \\
    \textnormal{Computational Science Hub} \\
    \textnormal{University of Hohenheim} \\
    }
\end{minipage}
\\[6em]
\hspace{0mm} 
\begin{minipage}[t]{0.45\textwidth}
    \centering
    {\fontsize{10}{12}\selectfont
    \href{https://orcid.org/0000-0002-6720-5938}{\includegraphics[scale=0.06]{orcid.pdf}\hspace{1mm} \textbf{Roland Gerhards}} \\
    \textnormal{Department of Weed Science} \\
    \textnormal{University of Hohenheim} \\
    }
\end{minipage}
}

\renewcommand{\shorttitle}{Assessing the Capability of YOLO- and Transformer-based Object Detectors for Real-time Weed Detection}
\begin{document}
\maketitle
\renewcommand\thefootnote{}
\footnotetext{This is the pre-review version of the manuscript submitted to Precision Agriculture (Springer).}
\footnotetext[1]{Corresponding author: \texttt{alicia.allmendinger@uni-hohenheim.de}}
\renewcommand\thefootnote{\arabic{footnote}}
\begin{abstract}
Spot spraying represents an efficient and sustainable method for reducing the amount of pesticides, particularly herbicides, used in agricultural fields. To achieve this, it is of utmost importance to reliably differentiate between crops and weeds, and even between individual weed species in situ and under real-time conditions. To assess suitability for real-time application, different object detection models that are currently state-of-the-art are compared. All available models of YOLOv8, YOLOv9, YOLOv10, and RT-DETR are trained and evaluated with images from a real field situation. The images are separated into two distinct datasets: In the initial data set, each species of plants is trained individually; in the subsequent dataset, a distinction is made between monocotyledonous weeds, dicotyledonous weeds, and three chosen crops. The results demonstrate that while all models perform equally well in the metrics evaluated, the YOLOv9 models, particularly the YOLOv9s and YOLOv9e, stand out in terms of their strong recall scores (66.58 \% and 72.36 \%), as well as mAP50 (73.52 \% and 79.86 \%), and mAP50-95 (43.82 \% and 47.00 \%) in dataset 2. However, the RT-DETR models, especially RT-DETR-l, excel in precision with reaching 82.44 \% on dataset 1 and 81.46 \% in dataset 2, making them particularly suitable for scenarios where minimizing false positives is critical. In particular, the smallest variants of the YOLO models (YOLOv8n, YOLOv9t, and YOLOv10n) achieve substantially faster inference times down to 7.58 ms for dataset 2 on the NVIDIA GeForce RTX 4090 GPU for analyzing one frame, while maintaining competitive accuracy, highlighting their potential for deployment in resource-constrained embedded computing devices as typically used in productive setups.

\end{abstract}

\section{Introduction}
The utilisation of herbicides is a prevalent practice employed to treat weeds in agricultural fields, with the objective of preventing crop losses due to these weeds \cite{korav2018study}. However, in recent years, concerns have emerged regarding the potential site-effects of herbicides on for example the environment, the human health, biodiversity, and so forth \cite{melander2005integrating}.  When considering the heterogeneous distribution of weeds and weed species in agricultural fields, it is obvious that it is not necessary to apply herbicides uniformly across the entire field \cite{marshall1988field}. Indeed, there may be areas with no weeds at all. Furthermore, it is not always beneficial to apply the same herbicide to all weeds, as there are varying degrees of effectiveness. With varying the herbicide the Weed control efficacy can be extended to a higher level compared to one herbicide for all weeds \cite{wiles2009beyond}. As a result, the focus has moved from broadcast application to site-specific weed management \cite{allmendinger2022precision}. Site-specific weed management is the targeted application of herbicides to specific areas where they are required.  In order to implement those site-specific weed management strategies, it is essential to obtain information regarding the specific weed species and their growth patterns within the field \cite{allmendinger2022precision}.
In the event that the herbicide mixture can be varied by a multi-tank sprayer during the crossing of the field, it is additionally necessary to identify the weed species in question \cite{gerhards2006practical}. There are a number of image processing methods for distinguishing between weeds and crops or different species. The advent of machine learning, and in particular, deep learning, and in this sector especially the convolutional neural networks (CNNs), has given rise to a lot of approaches that facilitate image processing, for example image classification and object detection. In image classification, an image is assigned to a class. In object detection, the coordinates within an image are also recognised, which allows for the assignment of several objects within an image to different classes \cite{7112511}. Upon closer examination of object detection, it becomes evident that there are again multiple approaches that can be distinguished. Two distinct categories of object detectors can be identified: two-stage and one-stage detectors \cite{liu2020deep}. Two-stage detectors, such as Faster R-CNN \cite{faster-r-cnn}, utilise a region proposal network in the initial stage, which identifies a limited number of regions of interest within the image. In the second stage, a convolutional neural network (CNN) \cite{lecun2015deep} is employed to encode the extracted features, predict bounding boxes and assign the objects to a class \cite{diwan2023object}. One-stage detectors, such as You Only Look Once (YOLO), integrate the two aforementioned steps into a single process \cite{YOLO1}. The bounding boxes are predicted and assigned directly to a class in one step. The combination of these two steps renders one-stage detectors the optimal choice for real-time applications in the field \cite{diwan2023object}. Since 2017, there has been a new and significant advancement in the field of neural networks, the Transformers. Initially developed for natural language processing (NLP), the transformer has demonstrated remarkable capabilities, particularly in the ability to concentrate on multiple sequences simultaneously through its self-attention mechanism \cite{vaswani2017attention}. The success of transformers in the field of NLP led to the development of an application for image classification. The objective is to divide images into patches that are used as tokens for the transformers \cite{dosovitskiy2020image}. These specific transformers, which are called Vision-based Transformers (ViT), have the capacity to outperform the current state-of-the-art models in image classification \cite{vaswani2017attention}. In the context of object detection tasks, the use of transformers is also a common approach. In particular, detection transformers (DETR) have been shown to be effective in this domain. The typical approach is to utilise CNNs as ResNet as their foundation. This enables the extraction of feature maps, thereby facilitating the spatial hierarchy and feature extraction capabilities \cite{carion2020end}. Furthermore, position encodings are employed in order to utilise the spatial information required for object detection \cite{carion2020end}. A number of studies have already tested the utilisation of transformers, although few have been conducted in an agricultural context, moreover, no studies have yet analysed the use of transformers for weed classification. However, this will change in the future due to the transformers’ demonstrated robustness. Transformers facilitate the real-time perception of field situations, thereby revealing significant variability.

However, a common feature of these approaches is the necessity of providing a suitable dataset for training purposes. Images captured in a laboratory setting are frequently used as datasets \cite{dyrmann2018estimation}. If Deep Neural networks (DNNs) are trained with those images and deployed in the field, their accuracy is likely to be compromised due to the potential for significant variations in real field conditions. In addition to different lighting conditions, the presence of diverse ground conditions can also introduce inconsistencies in the images. On the one hand it is recommended that the data to test DNNs consists of images taken under the same conditions as those used for training and validation\cite{rakhmatulin2021deep}, since the performance on such a testing set can provide insight into the DNNs effectiveness in replicating learned patterns. On the other hand, images under varied conditions should also be included, as testing with diverse scenarios helps to assess the DNNs ability to generalize and adapt to different settings, including those with different plant phenotypes and soil conditions. Furthermore, for real agricultural applications such a system will have to deal with a variety of diverse conditions \cite{machleb2020sensor}.
As mentioned above, there is a considerable variation in the field of agriculture. Therefore, it is imperative that the models demonstrate robust performance \cite{rezaei2024plant} to ensure a precise application of herbicides in the field. However, the issue of computationally expensive training remains a significant challenge, as is the frequent use of systems that are not universally adopted. This can raise issues in the agricultural context, where the use of small devices is required \cite{rezaei2024plant}.

The objective of this study is to compare different state-of-the-art models in terms of efficacy and predictability. All models are trained on images taken under realistic field conditions rather than in a controlled laboratory environment. This enbales the evaluation of the suitability of models for a real-time applications in varying environmental conditions, as well as the identification of models that can be implemented in a realistic manner in the future. In order to ascertain whether a real-time application was feasible, the inference time on a range of devices is also evaluated.

\subsection{Related work}
\paragraph{One-stage detectors}
As the use of image analysis methodologies becomes increasingly prevalent within the agricultural sector, a number of studies have emerged that address this subject. For example, the study by \cite{saleem2022weed}. The dataset contained 17 509 images from 8 classes of the DeepWeeds dataset \cite{saleem2022weed}. They conducted a comparative analysis of different DNNs belonging to the two-stage and one-stage detector categories. The results indicated that the two-stage detectors exhibited superior performance compared to the one-stage detectors, when the default settings were employed. For example, the Faster RCNN ResNet-101 model achieved a mAP50-95 score of 87.64\%, whereas YOLOv4 achieved a mAP50-95 score of 79.68\%. Furthermore, the study explored the use of advanced training techniques, such as image resizing and weight optimization, which led to an improvement in the mAP score by 5.8\% \cite{saleem2022weed}. In comparison to their study,  the present research employed a total of 16 species, representing a twofold increase. The analysis is focused on comparing various YOLO and Detection Transformer models. This approach was selected because both types of detectors are promising for real-time applications.
In another study \cite{ahmad2021performance} evaluated the efficacy of various models in the context of image classification and object detection. Their dataset consisted of 462 images from four classes of early-season weeds found in corn and soybeans. In addition to image classification models, YOLOv3 was analyzed as an object detection model. The results indicated that in an appropriate field situation, the use of YOLOv3 for object detection is possible, provided that a dataset of an appropriate size is available. YOLOv3 achieved an mAP50-95 score of 54.3\%, although it showed difficulty in recognizing grass weeds due to their thin leaves. Another notable finding was that the researchers recommended a minimum of 100 images per class to ensure reliable detection \cite{ahmad2021performance}. In comparison, the present study involved a significantly larger number of species and images. Additionally, the issue of monocots was addressed by classifying them alongside dicots. Furthermore, the YOLO Models up to YOLOv10  represent the current state of research and are demonstrably superior.
\cite{dang2023YOLOweeds} as well focused in their study on the comparison of different YOLO models. They used a dataset of 5648 images taken in cotton fields. The dataset was tested with and without augmentation and the images were taken under natural lighting conditions. The highest mAP50-95 score was achieved with YOLOv4-\textit{P6} with 89.72, the lowest with YOLOv3-tiny with 68.18\% \cite{dang2023YOLOweeds}. Their study employed a smaller number of species than used in the present study. Furthermore, the focus was on the integration of the latest YOLO and RT-DETR models. Another noteworthy aspect is the use of realistic images with diverse species for testing purposes. This approach enables more accurate prediction of the models' precision in field deployment.

\paragraph{Transformer based Detectors}
Numerous studies have been conducted on image analysis in the domain of Transformers. \cite{li2024detr} conducted a study on a semi-supervised object detection method based on a DETR-like transformer. They tested two techniques: low-threshold filtering and decoupled optimization. This should improve class imbalances and multitask optimization. In the context of  plant detection tasks, the use of 5\% of their dataset, comprising 18 images, yielded in a mAP50-95 score of 74.1\%. This has already surpassed current state-of-the-art models in terms of performance \cite{li2024detr}.
\cite{zhao2024detrs} conducted a study using the first end-to-end real-time DETR (RT-DETR), which was trained on the COCO data set and not in an agricultural context. RT-DETR demonstrated superior performance compared to YOLO, mainly due to the fact that YOLO until v9 employs maximum suppression, which significantly impairs the speed and is not used in RT-DETR \cite{zhao2024detrs}.
The application of RT-DETR in an agricultural context is, as yet, uncommon, and similarly, its deployment for the purpose of object detection has yet to be empirically validated. One of the few examples of RT-DETR use in an agricultural context is the application of this technique to ascertain the ripeness of blueberries \cite{aguilera2023comprehensive}. The clustered growth of blueberries presents a challenge for direct detection in images, often resulting in inaccuracies and imprecise bounding boxes. RT-DETR has shown the ability to outperform existing models, achieving greater accuracy in the recognition of blueberries. Furthermore, the model demonstrated inference times comparable to those of the YOLOv7-default model \cite{aguilera2023comprehensive}.
These studies show that transformers are capable of facilitating real-time monitoring and decision-making because of their versatility and adaptability. However, they also have disadvantages. For example, the computational power required for training is very high. Furthermore, configurations that are not typically available on standard systems are often required. Consequently, this can cause problems in agricultural settings, where the deployment of smaller devices is often necessary \cite{rezaei2024plant}. 
As \cite{huang2024small} also report, the DETR algorithm frequently encounters difficulties in identifying small objects. This can be a significant challenge in an agricultural context, as the plants to be detected exhibit a wide range of sizes. Smaller plants are particularly susceptible to being overlooked, yet they must be identified with the same level of accuracy as larger objects, as failure to do so can result in yield losses later on. However, these issues can be mitigated by using data augmentation and adaptive feature fusion algorithms at the underlying level \cite{huang2024small}.

\paragraph{Computer Vision-based Weed Detection}

In the context of weed identification in agriculture, there are numerous approaches beyond the use of CNN. These Computer vision-based methods have been in use for a longer period and are still employed in various systems. The development of computer vision-based weed recognition is rooted in traditional image processing and utilizes machine learning techniques. This process facilitated the detection of weeds by assessing their texture, shape, spectral, color or a combination of these features \cite{wu2021review}. Currently, there are already several robots that have been designed to identify different types of weeds in the field using machine learning algorithms. One example is the Avo robot, produced by Ecorobotix \cite{ecorobotix_website}. It is suitable for use in in row crops. The robot is able to apply herbicides with great precision, which has the potential to reduce the amount of herbicide used by up to 95 \%. 
Another robotic solution is the Robovator from Poulsen \cite{poulsen_website} which uses Machine learning to move blades, that work in the intra-row area, additional to the fixed blades that work inter-row. An additional example is the Kult-iSelect® hoe, which employs image processing and Artificial Intelligence to regulate hydraulic V-shaped blades, which function within and between rows \cite{gerhards2024comparison}.

\section{Materials and Methods}

\subsection{Experimental dataset construction}

The dataset used consists of images taken under natural environmental conditions with a Sony Alpha 7R Mark4 (ILCE7-RM4, Sony Corporation, Tokyo, Japan), a 61-megapixel RGB DSLR camera. The images were taken in 2019 at the Heidfeldhof Research Station of the University of Hohenheim, located in southwestern Germany (48\textdegree 42'$59.0''$N and 9\textdegree 11'$35.4''$E). The images presented were captured by \cite{peteinatos2020weed}. The camera utilizes a shutter speed of 1/2500 s and automated ISO calibration, enabling the attainment of optimal image quality even in fluctuating light conditions. The Zeiss Batis 25mm lens used has a fixed focal length. The camera was mounted on the ``Sensicle'' at a height of 1.2 meters positioned vertically to the ground. The Sensicle is a multisensor platform for precision farming experiments. The speed during the crossing was 4 km h\textsuperscript{-1} and one image was taken every second. Despite the fact that the images were consistently captured at noon, the prevailing conditions at that time varied. This encompasses a range of lighting and ground conditions. The images were captured over a period of 45 days, beginning on the day of emergence and continuing until the eight-leaf stage or the onset of tillering. 
The dataset used consists of 5611 images from 16 classes, on average 350 images per class. Among them, three crop species, namely \textit{Helianthus annuus} L. (sunflower), \textit{Triticum aestivum} L. (winter wheat), and \textit{Zea mays} L. (maize) In addition, nine dicotyledonous weeds \textit{Abutilon theophrasti} Medik., \textit{Amaranthus retroflexus} L., \textit{Chenopodium album} L., \textit{Geranium spp.}, \textit{Lamium purpureum} L., \textit{Fallopia convolvulus} (L.) Á.  Löve, \textit{Solanum nigrum} L.,\textit{Thlaspi arvense} L., \textit{Veronica persica} Poir. and four monocotyledonous weeds \textit{Elymus repens} (L.) Gould, \textit{Alopecurus myosuroides} Huds., \textit{Avena fatua} L., and \textit{Setaria spp.} are represented in the dataset.

\subsection{Preparation of the Dataset}

In the initial phase of the processing procedure, a preliminary sorting of the images is conducted to exclude those of insufficient quality. This includes images where the camera has not been focused, or where the presence of insects has obstructed the view. The images were manually labeled by weed science researchers. The LabelImg tool\footnote{LabelImg:\url{https://docs.ultralytics.com/reference/utils/loss/}, last accessed: 01-Sept-2024} was used to create a rectangular bounding box around each plant, which was then assigned to the corresponding species. In the case of \textit{Setaria} and \textit{Geranium}, it was decided not to select specific species, but to continue with “\textit{spp}.”, as differentiation in the cotyledon stage was not possible in some cases due to the size and the corresponding morphology of the plants, which resulted in a larger number of images for those species. As the images were captured until the plants fully overlapped, a certain degree of overlap is shown in the dataset. The overlaps were incorporated into the bounding boxes, which explains why some bounding boxes can overlap in parts where the outlines of the plants were not visible. This approach was selected to guarantee the identification of all plants. In the absence of labeling, the lack of recognition would reult in missing the plant for treatment, which could potentially have adverse effects, depending on the plant in question.
The images were organized in two different datasets. In dataset 1, each species was considered as a single class, in dataset 2 all monocot weeds were categorized together, as were all dicot weeds. The three crops were still regarded as discrete classes. The selection of this approach is based on the understanding that the differentiation of groups prior to the application of herbicides can result in significant cost savings and enhanced efficacy. By selecting herbicides specifically tailored to each weed group, the previously mentioned benefits can be realized. In light of the prevailing focus on crop plants in an agricultural field, this approach is deemed optimal as it is imperative to be able to differentiate between crop and weed plants.\\

Both data sets were randomly divided into training, validation, and testing images, with a ratio of 68/17/15. This ratio is derived from the initial division, where 85 \% is allocated for training and validation, and 15 \% for testing. To enhance the training process, a five-fold cross-validation was conducted on the training images. This involves splitting the training dataset into five parts. In each fold, four parts are used for training and one part is used for validation.
As illustrated in Fig. \ref{fig:pipeline} the five-fold-cross validation was employed to assess the reliability of the findings and to evaluate whether the results were influenced by the specific images included in the training and validation datasets. This procedure generates five distinct combinations of images. The results presented in the Tables \ref{tab:performance_YOLO_dataset1} to \ref{table:inference_time2} reflect the mean values of these five runs, along with their standard deviations. Importantly, the images used for testing are independent of the training images to prevent data leakage. Table \ref{tab:number_images} provides an overview of the number of images and instances utilized for training and validation. 
The training images were employed for the purpose of training the network, the validation images were utilized for the hyperparameter tuning process, and the test images were used for the implementation of the DNNs and to test their robustness to images that had not been previously encountered. The following results  are those obtained from the DNNs with the test images. The selected models in this study were subjected to training, validation and testing with all the folds.
However, since the images were captured of the same plants in different growth stages, it is possible that the same plants can be included in both the training and the testing sets, although at different growth stages. The images used for training and validation were limited to one species per image. The images utilized for testing can comprise multiple species on a single image, thereby simulating a realistic field situation and detecting how well the models generalize to different field situations. However, the other specified parameters, including image quality and exposure ratio, align with those images used for training and validating of the DNNs. 

\begin{table}[H]
\caption{Number of images and instances used for training and validation averaged for all five folds.}
\label{tab:number_images}
\resizebox{\textwidth}{!}{%
\begin{tabular}{l|c|c|c|c}
\toprule
\textbf{Class} & \multicolumn{2}{c|}{\textbf{Training}} & \multicolumn{2}{c}{\textbf{Validation}} \\
\midrule
& \textbf{Images} & \textbf{Instances} & \textbf{Images} & \textbf{Instances} \\
\midrule
All & 3783 & 45475 & 945 & 11368\\ 
\textit{Abutilon theophrasti} Medik. & 217 & 4500 & 54 & 1125\\ 
\textit{Elymus repens} (L.) Gould & 216 & 1323 & 54 & 330\\ 
\textit{Alopecurus myosuroides} Huds. & 253 & 2501 & 63 & 625\\ 
\textit{Amaranthus retroflexus} L. & 265 & 1186 & 66 & 296\\ 
\textit{Avena fatua} L. & 247 & 3126 & 61 & 781 \\ 
\textit{Chenopodium album} L. & 217 & 873 & 54 & 218 \\ 
\textit{Geranium spp.} & 235 & 4259 & 58 & 1064\\ 
\textit{Helianthus annuus} L. & 244 & 7654 & 61 & 1913 \\ 
\textit{Lamium purpureum} L. & 248 & 2499 & 62 & 624 \\ 
\textit{Fallopia convolvulus} (L.) Á  Löve & 214 & 403 & 53 & 100 \\ 
\textit{Setaria spp.} & 237 & 764 & 59 & 191 \\ 
\textit{Solanum nigrum} L. & 222 & 1496 & 55 & 374 \\ 
\textit{Thlaspi arvense} L. & 220 & 2685 & 55 & 671 \\ 
\textit{Triticum aestivum} L. & 248 & 6068 & 62 & 1517 \\ 
\textit{Veronica persica} Poir. & 231 & 1054 & 57 & 263 \\ 
\textit{Zea mays} L. & 262 & 5080 & 65 & 1270 \\

\bottomrule
\end{tabular}

}
\end{table}

\begin{figure}[H]
    \centering
    \begin{subfigure}{0.3\textwidth}
        \centering
        \includegraphics[width=\textwidth]{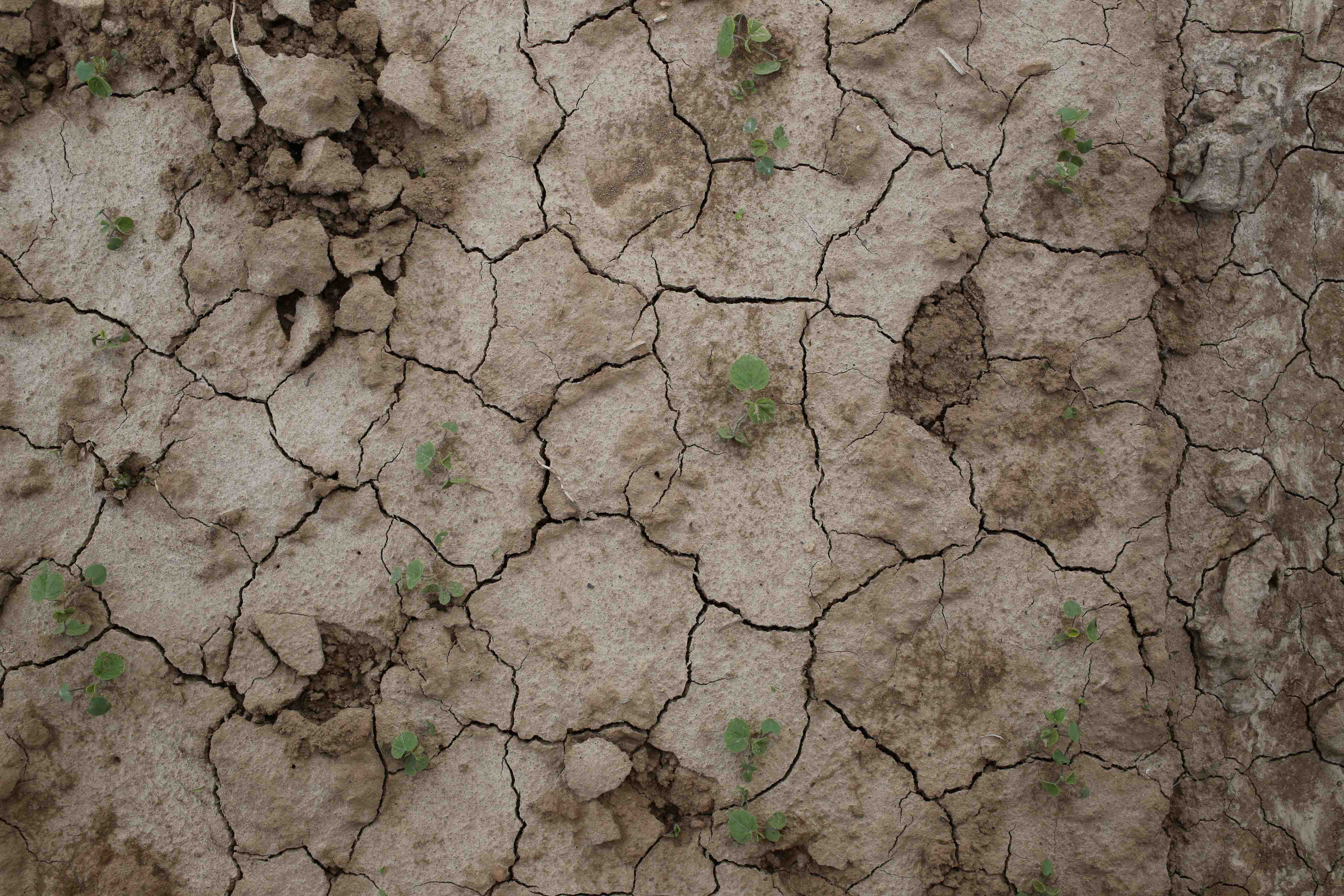}
        \caption{\textit{Abutilon theophrasti} Medik.}
    \end{subfigure}
    \begin{subfigure}{0.3\textwidth}
        \centering
        \includegraphics[width=\textwidth]{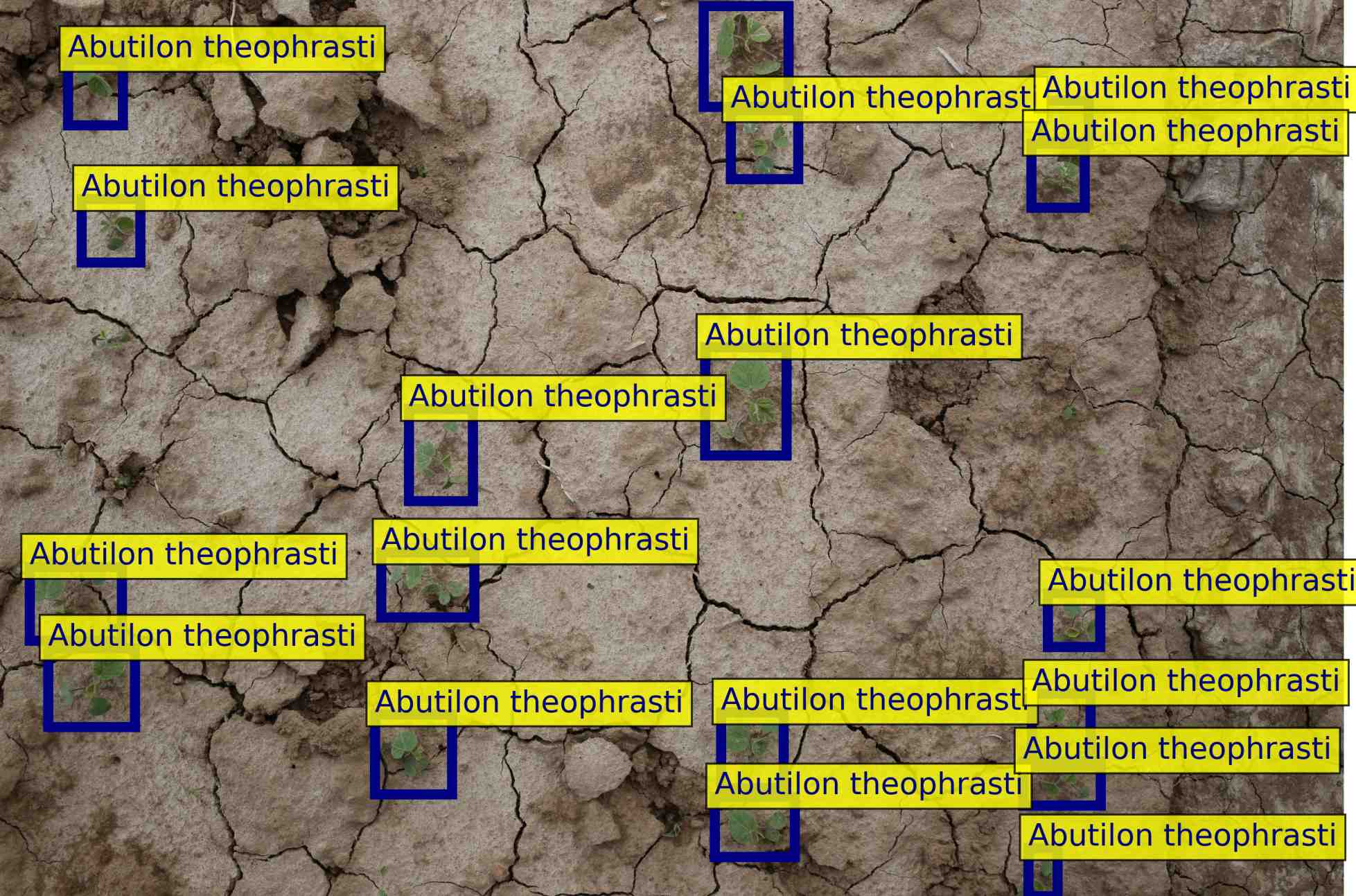}
        \caption{Dataset 1}
    \end{subfigure}
    \begin{subfigure}{0.3\textwidth}
        \centering
        \includegraphics[width=\textwidth]{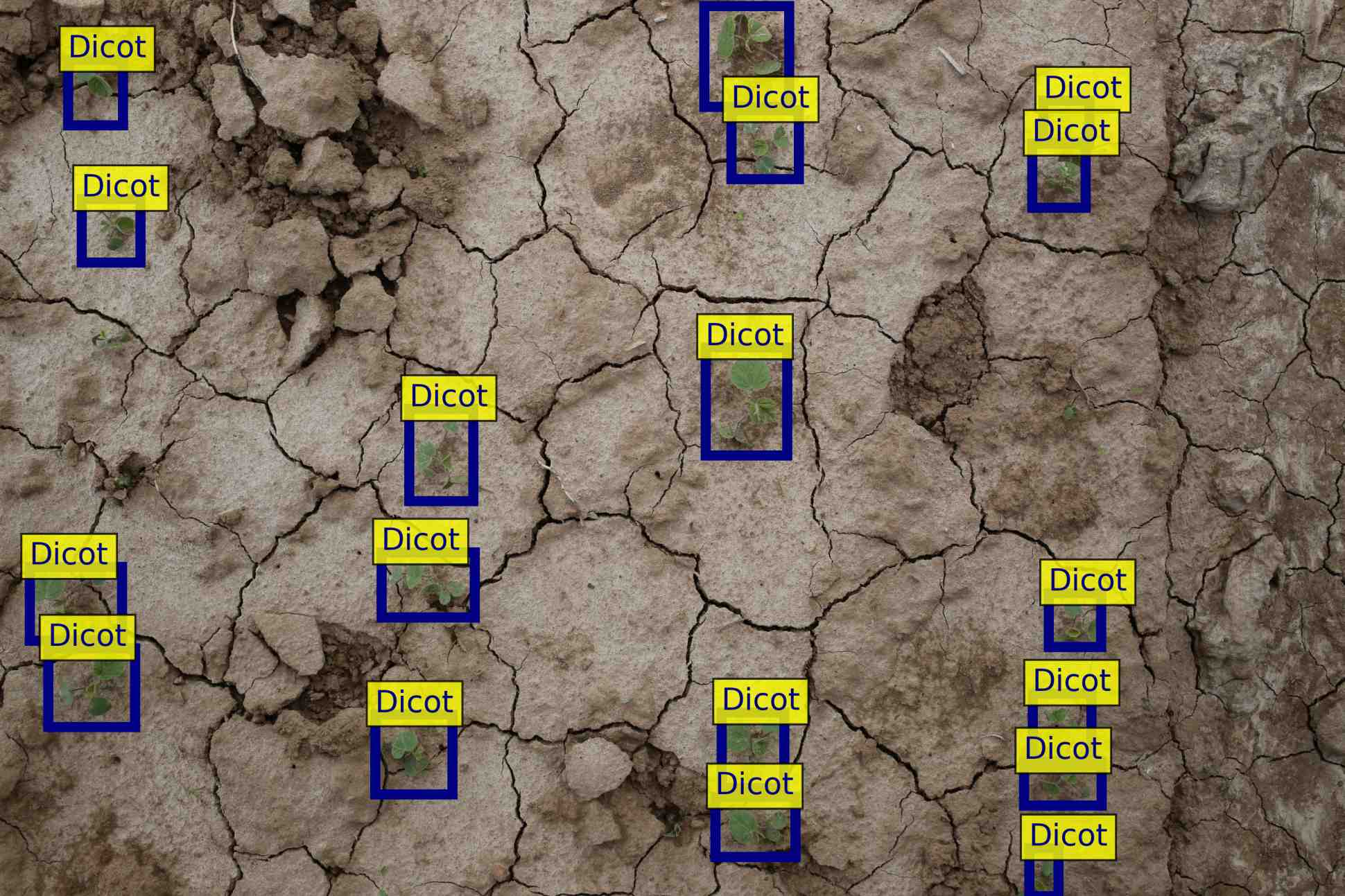}
        \caption{Dataset 2}
    \end{subfigure}

    \begin{subfigure}{0.3\textwidth}
        \centering
        \includegraphics[width=\textwidth]{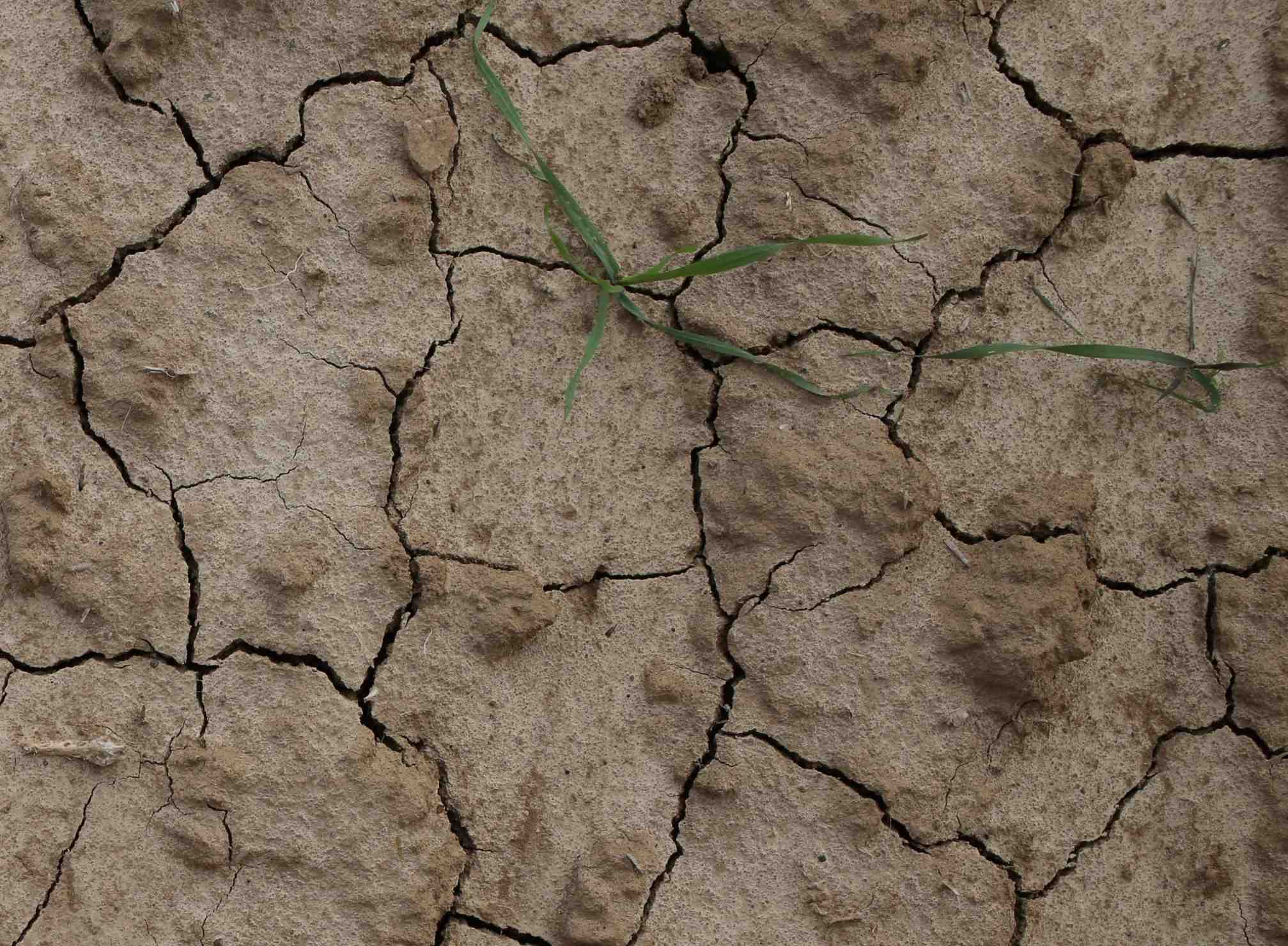}
        \caption{\textit{Elymus repens} (L.) Gould}
    \end{subfigure}
    \begin{subfigure}{0.3\textwidth}
        \centering
        \includegraphics[width=\textwidth]{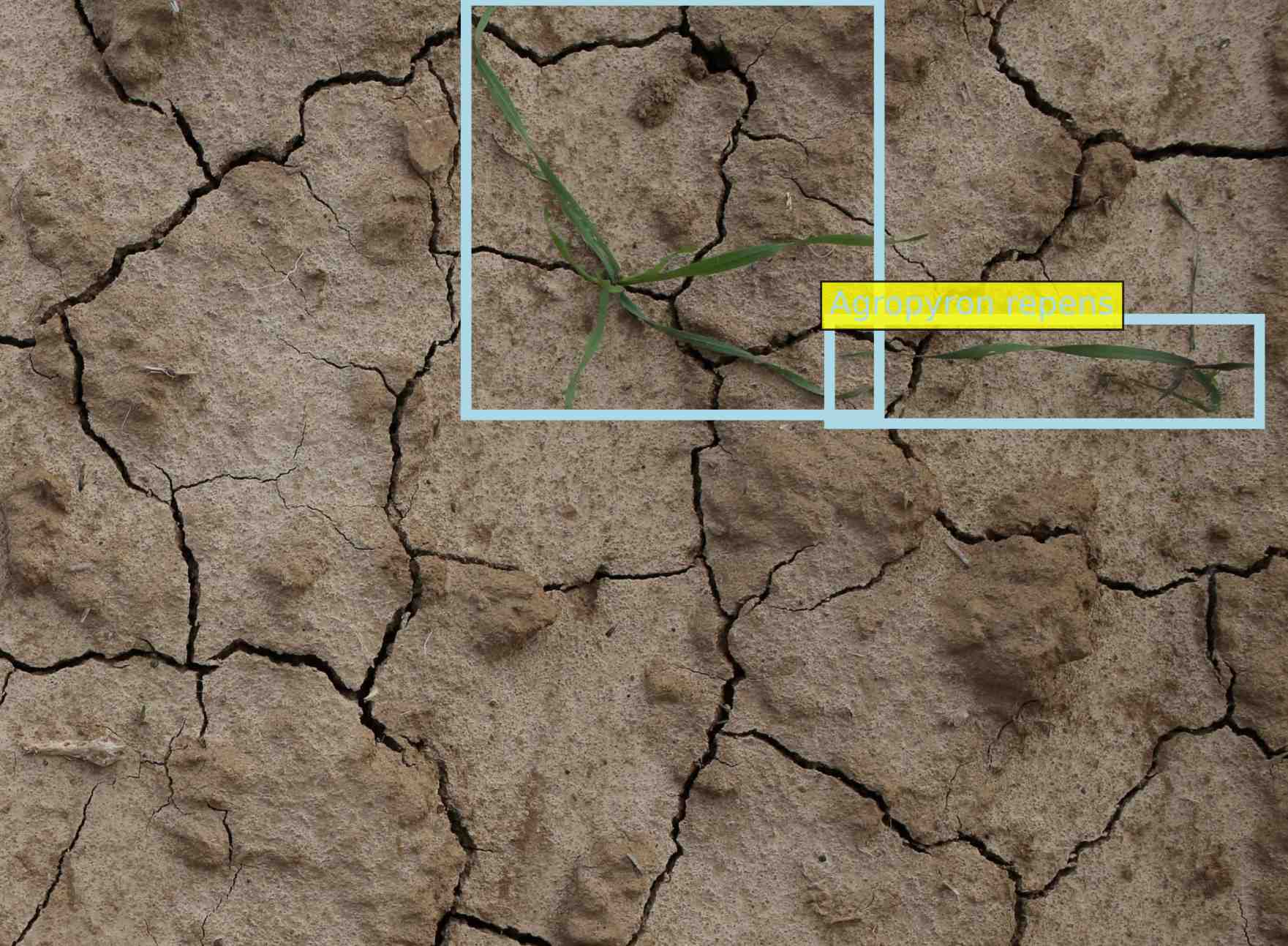}
        \caption{Dataset 1}
    \end{subfigure}
    \begin{subfigure}{0.3\textwidth}
        \centering
        \includegraphics[width=\textwidth]{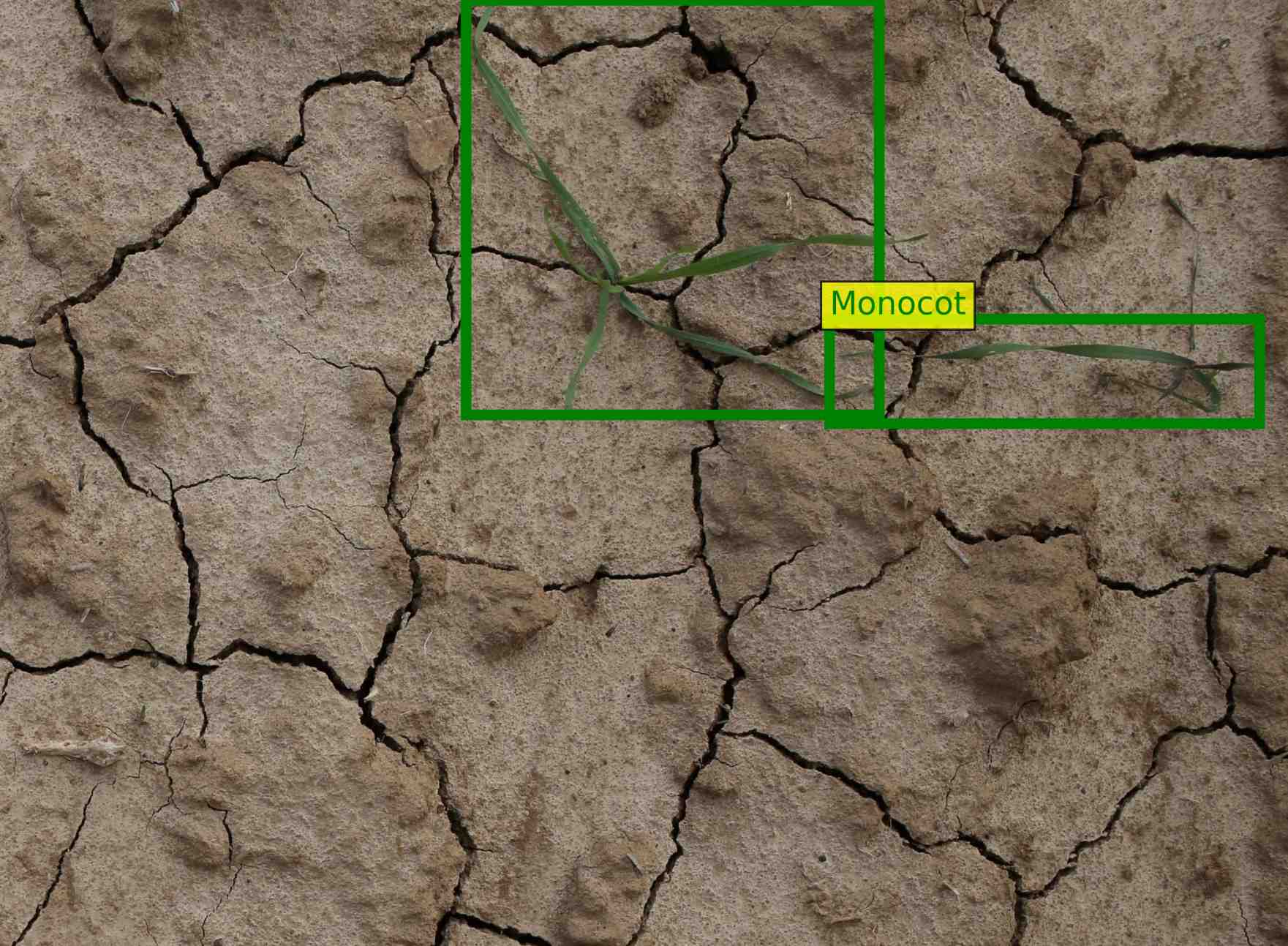}
        \caption{Dataset 2}
    \end{subfigure}

    \begin{subfigure}{0.3\textwidth}
        \centering
        \includegraphics[width=\textwidth]{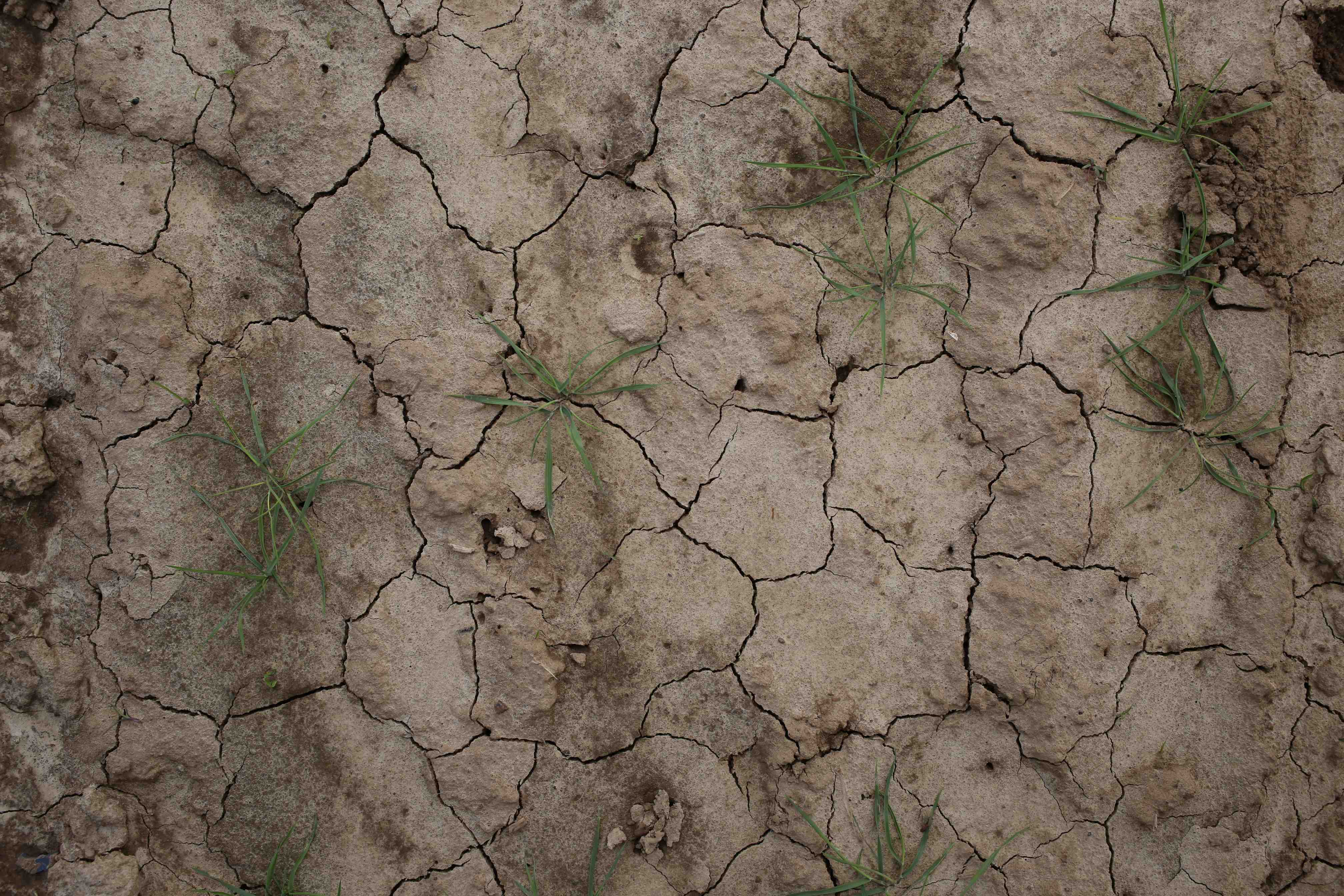}
        \caption{\textit{Alopecurus myosuroides} Huds.}
    \end{subfigure}
    \begin{subfigure}{0.3\textwidth}
        \centering
        \includegraphics[width=\textwidth]{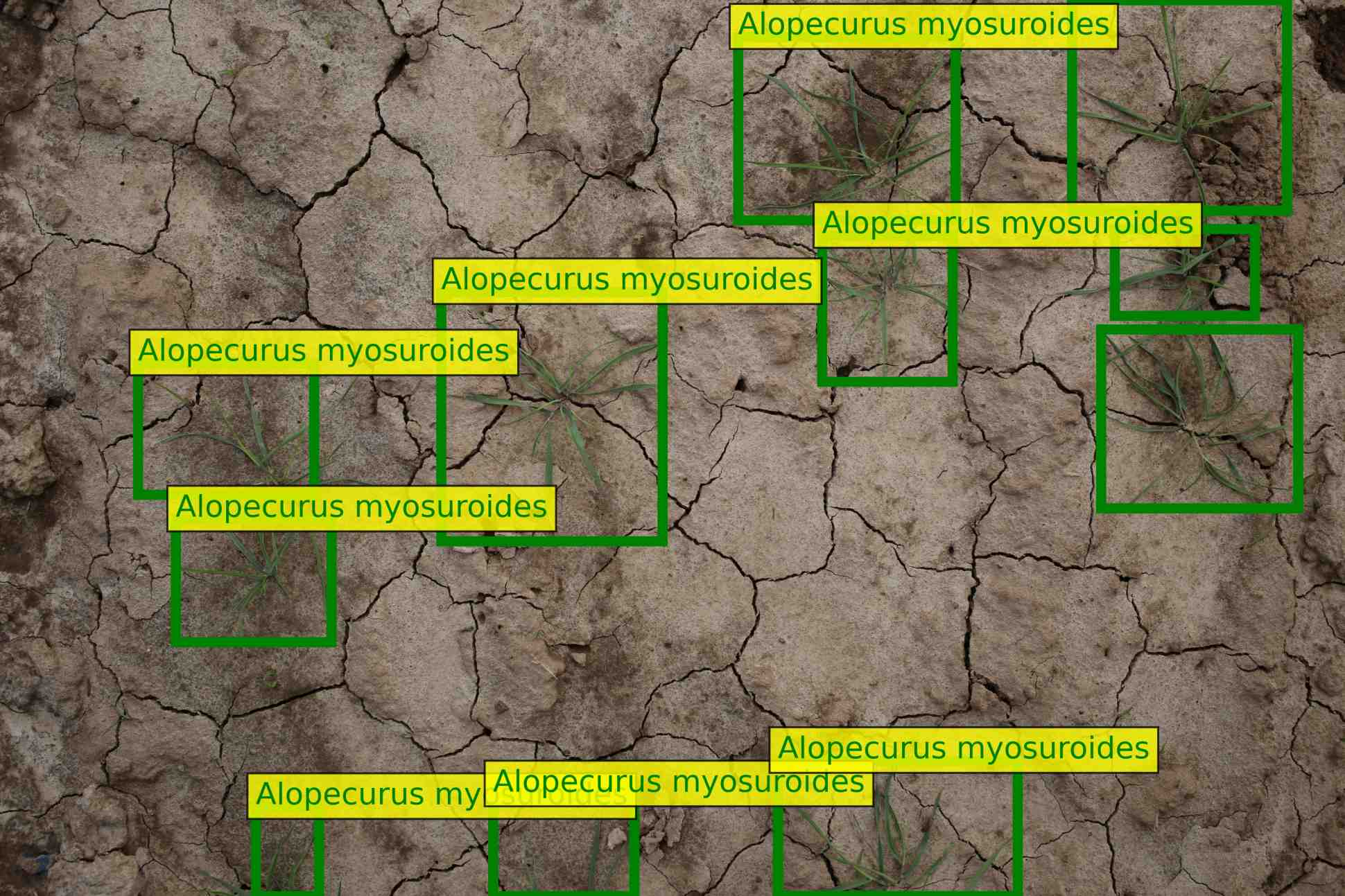}
        \caption{Dataset 1}
    \end{subfigure}
    \begin{subfigure}{0.3\textwidth}
        \centering
        \includegraphics[width=\textwidth]{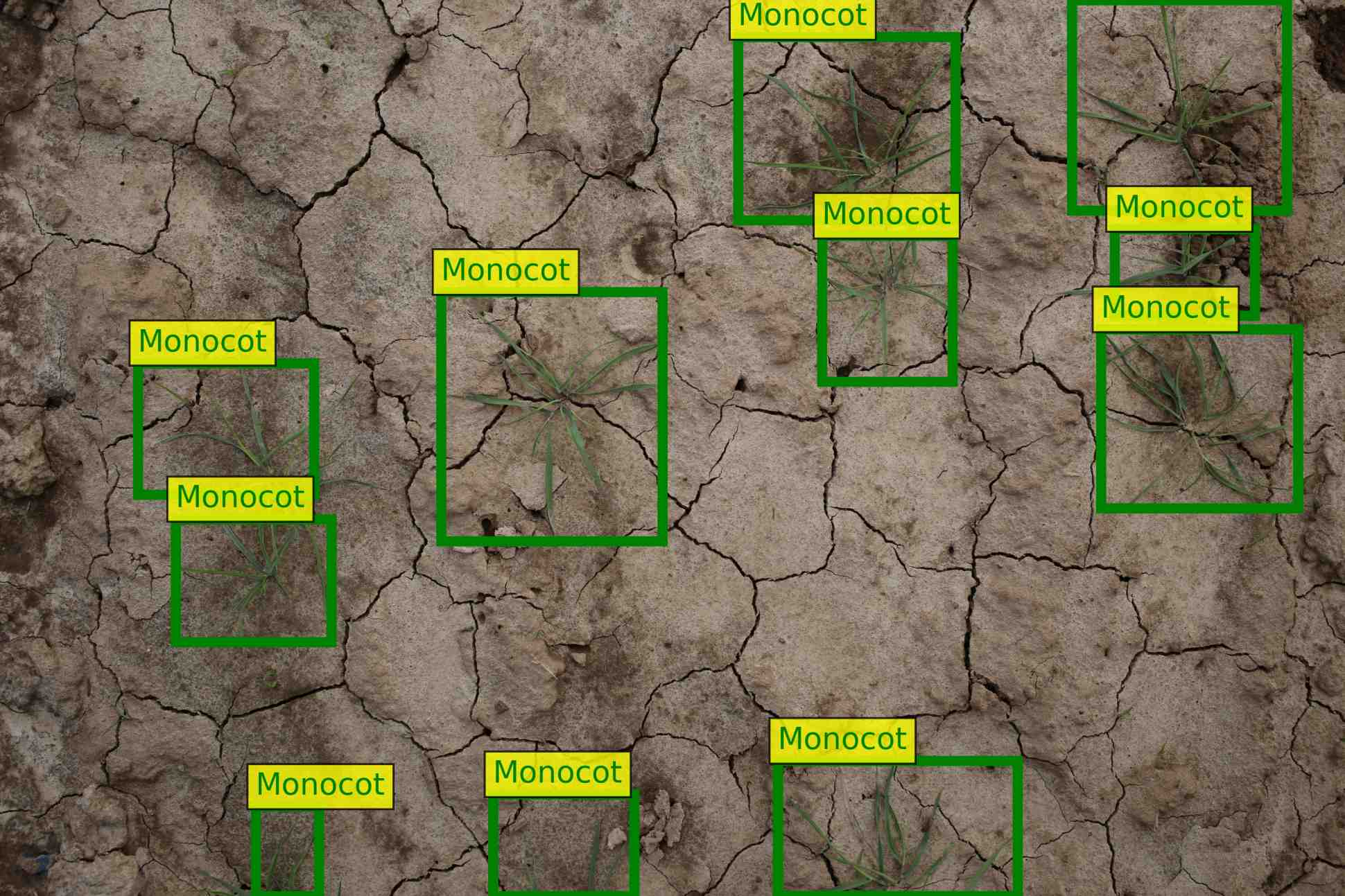}
        \caption{Dataset 2}
    \end{subfigure}

    \begin{subfigure}{0.3\textwidth}
        \centering
        \includegraphics[width=\textwidth]{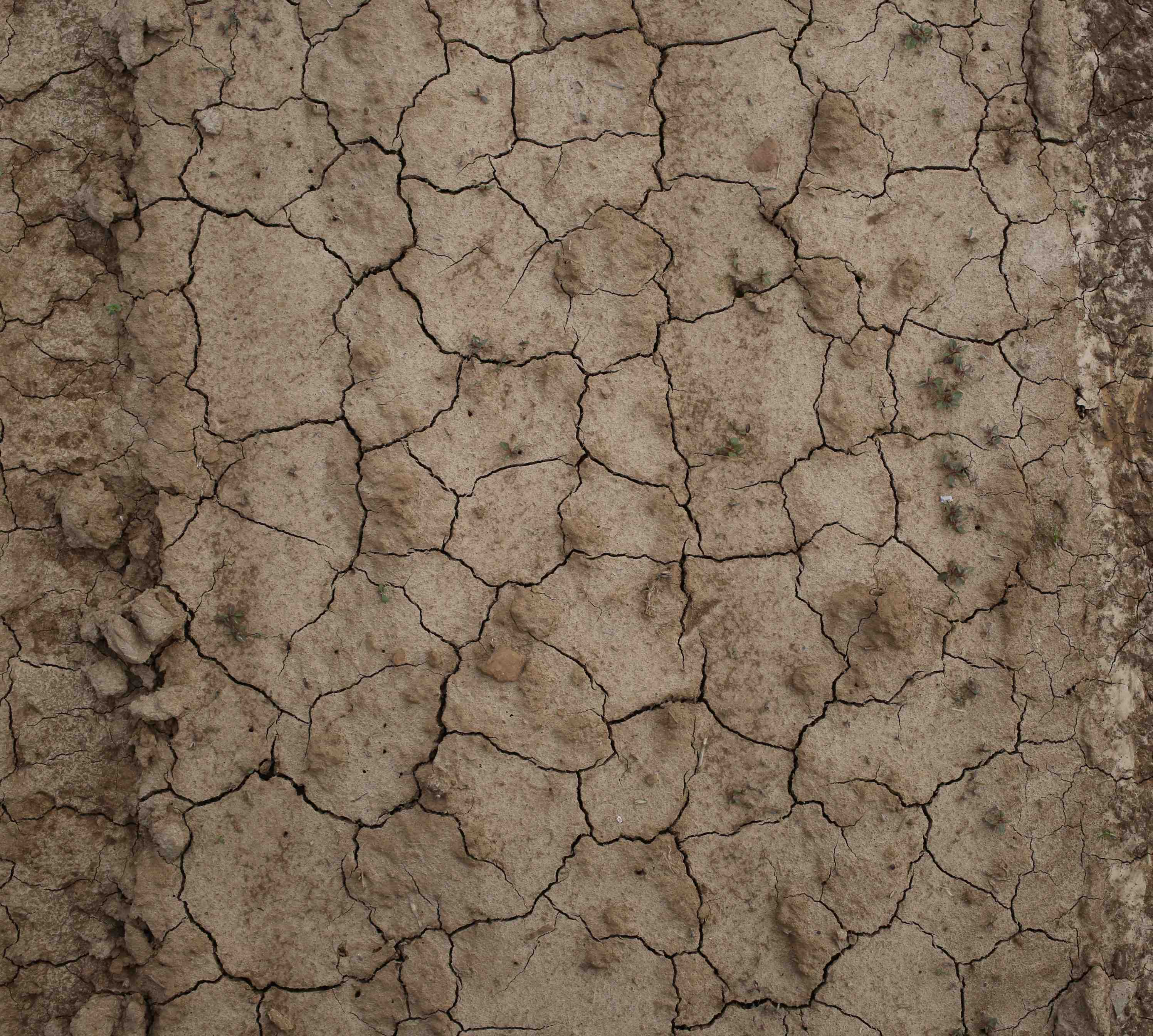}
        \caption{\textit{Amaranthus retroflexus} L.}
    \end{subfigure}
    \begin{subfigure}{0.3\textwidth}
        \centering
        \includegraphics[width=\textwidth]{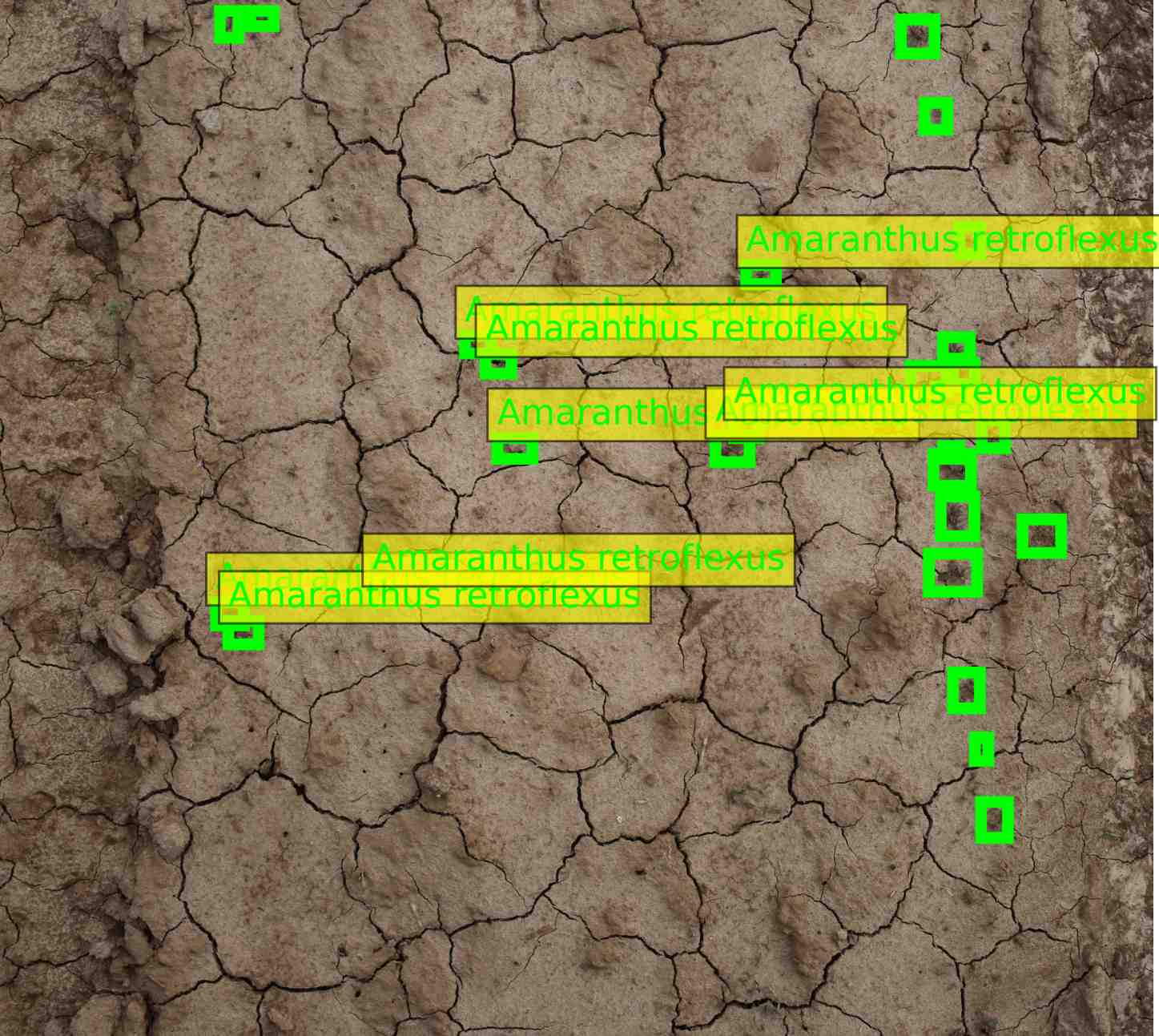}
        \caption{Dataset 1}
    \end{subfigure}
    \begin{subfigure}{0.3\textwidth}
        \centering
        \includegraphics[width=\textwidth]{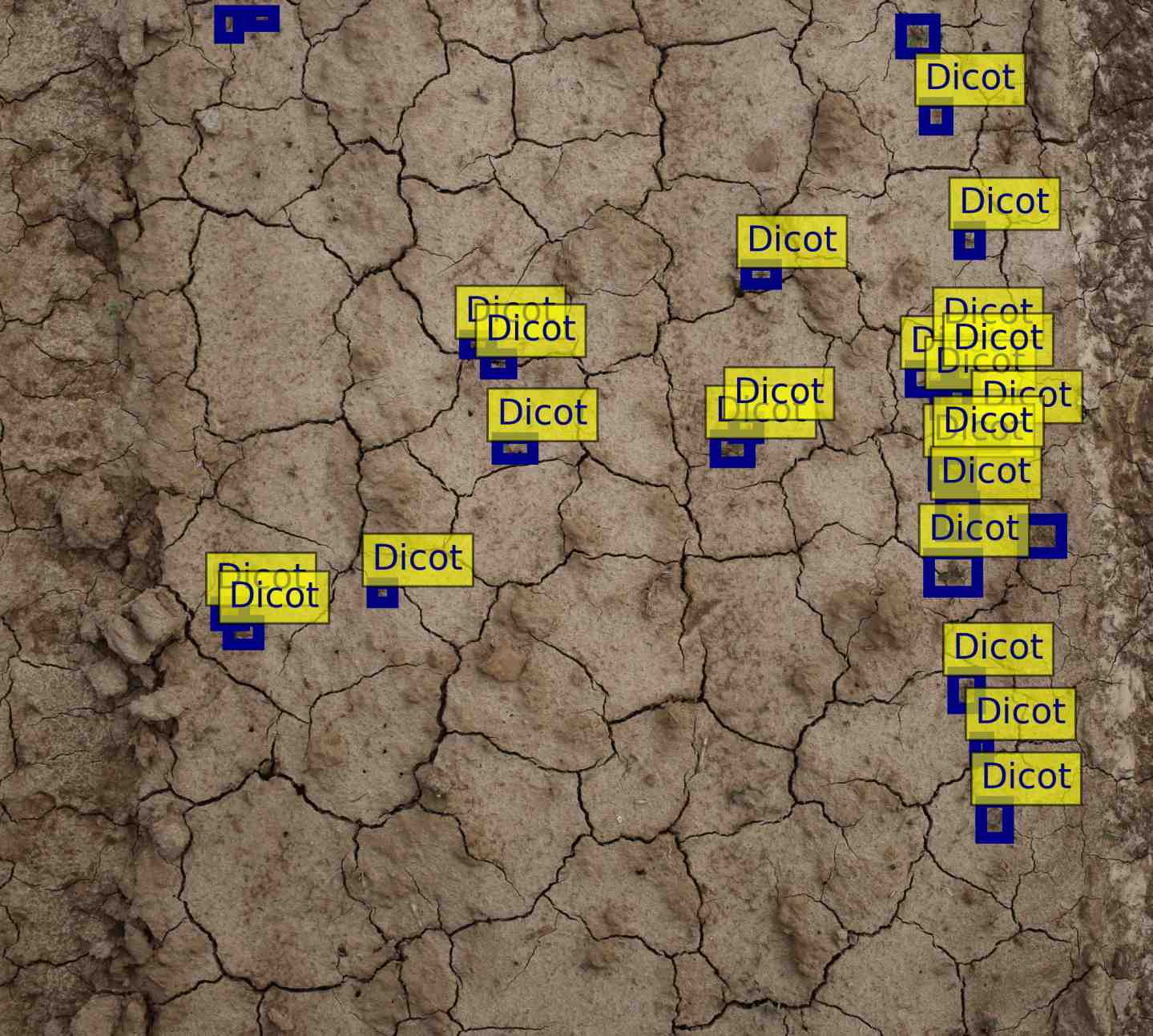}
        \caption{Dataset 2}
    \end{subfigure}

    \begin{subfigure}{0.3\textwidth}
        \centering
        \includegraphics[width=\textwidth]{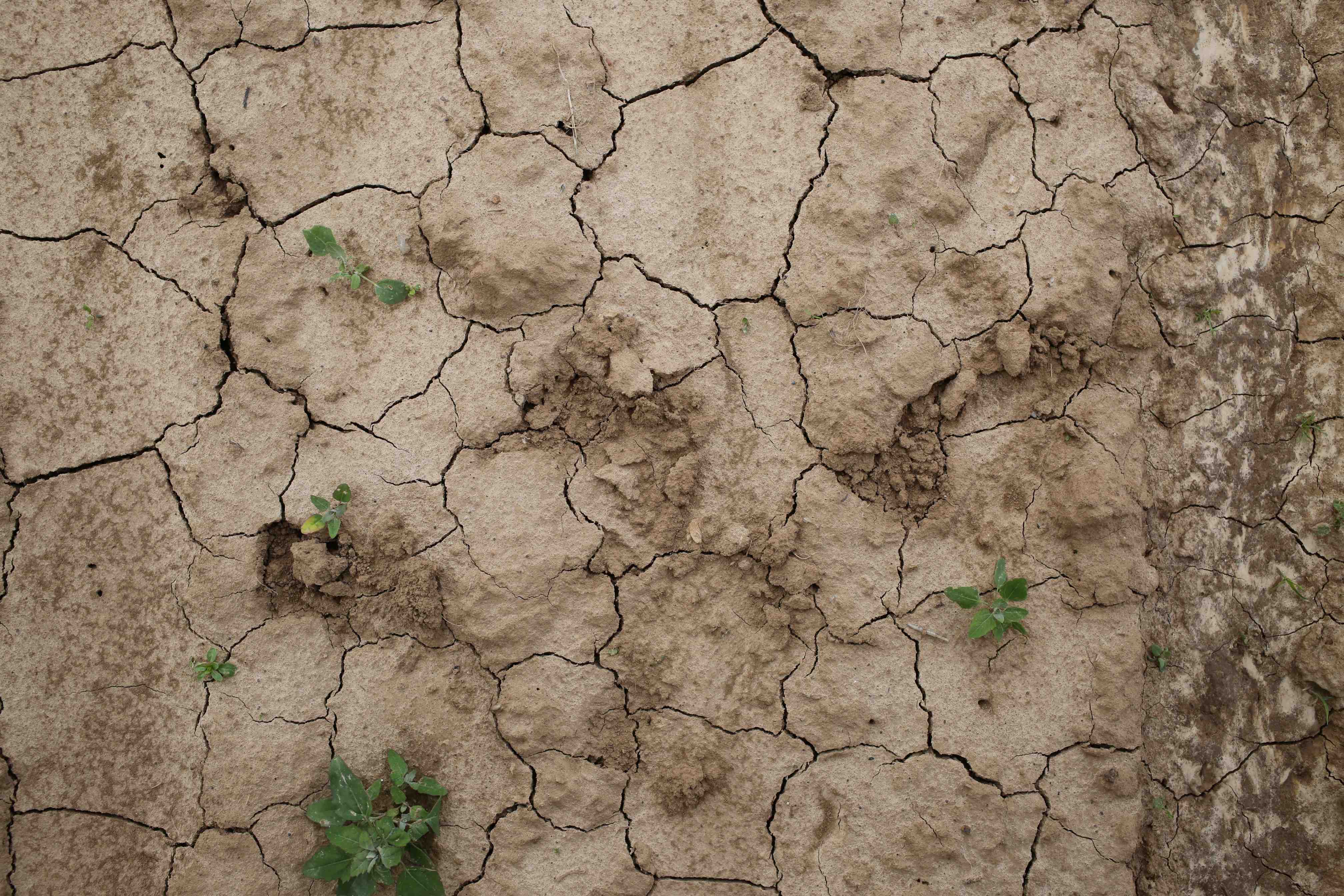}
        \caption{\textit{Chenopodium album} L.}
    \end{subfigure}
    \begin{subfigure}{0.3\textwidth}
        \centering
        \includegraphics[width=\textwidth]{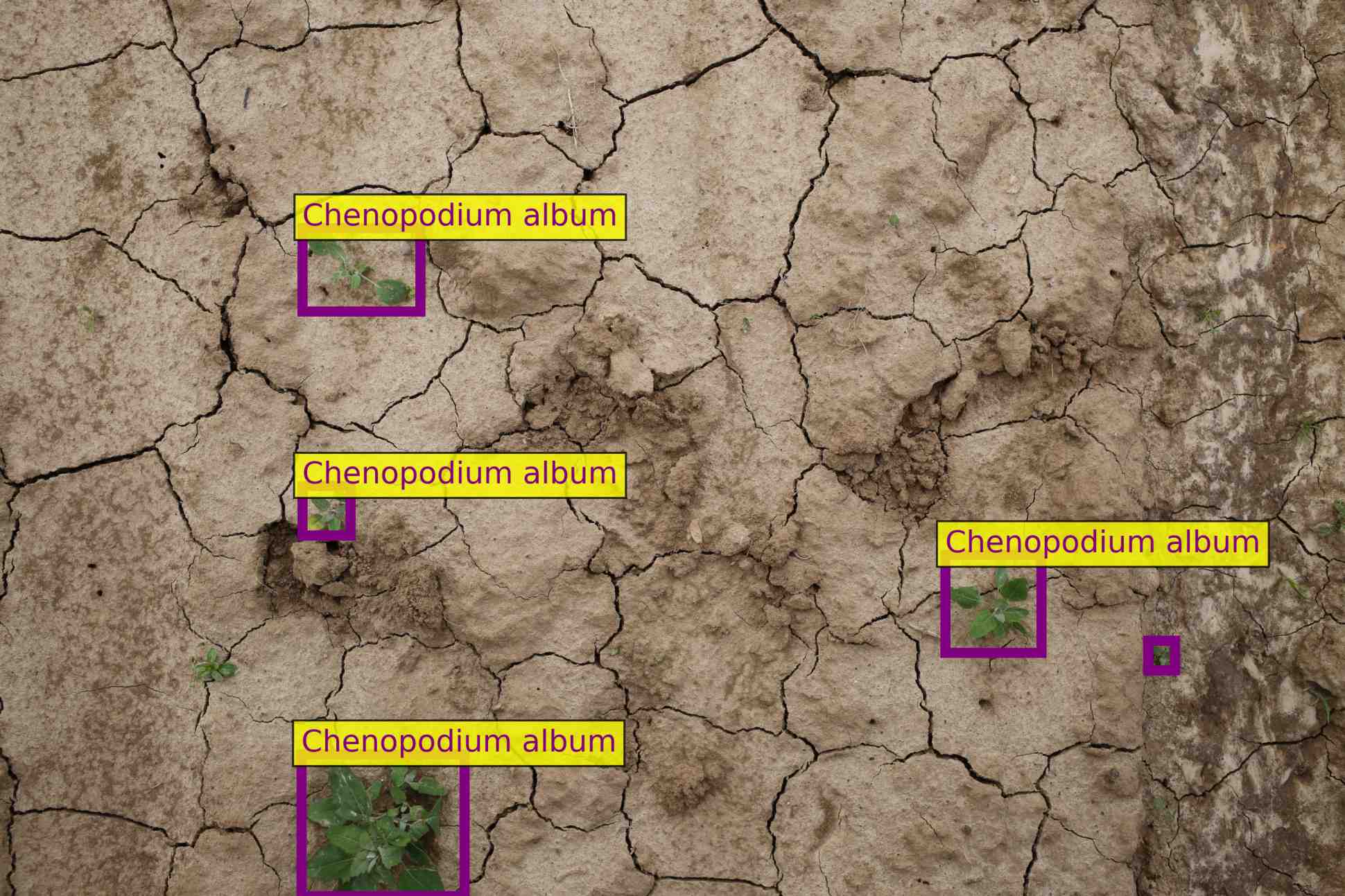}
        \caption{Dataset 1}
    \end{subfigure}
    \begin{subfigure}{0.3\textwidth}
        \centering
        \includegraphics[width=\textwidth]{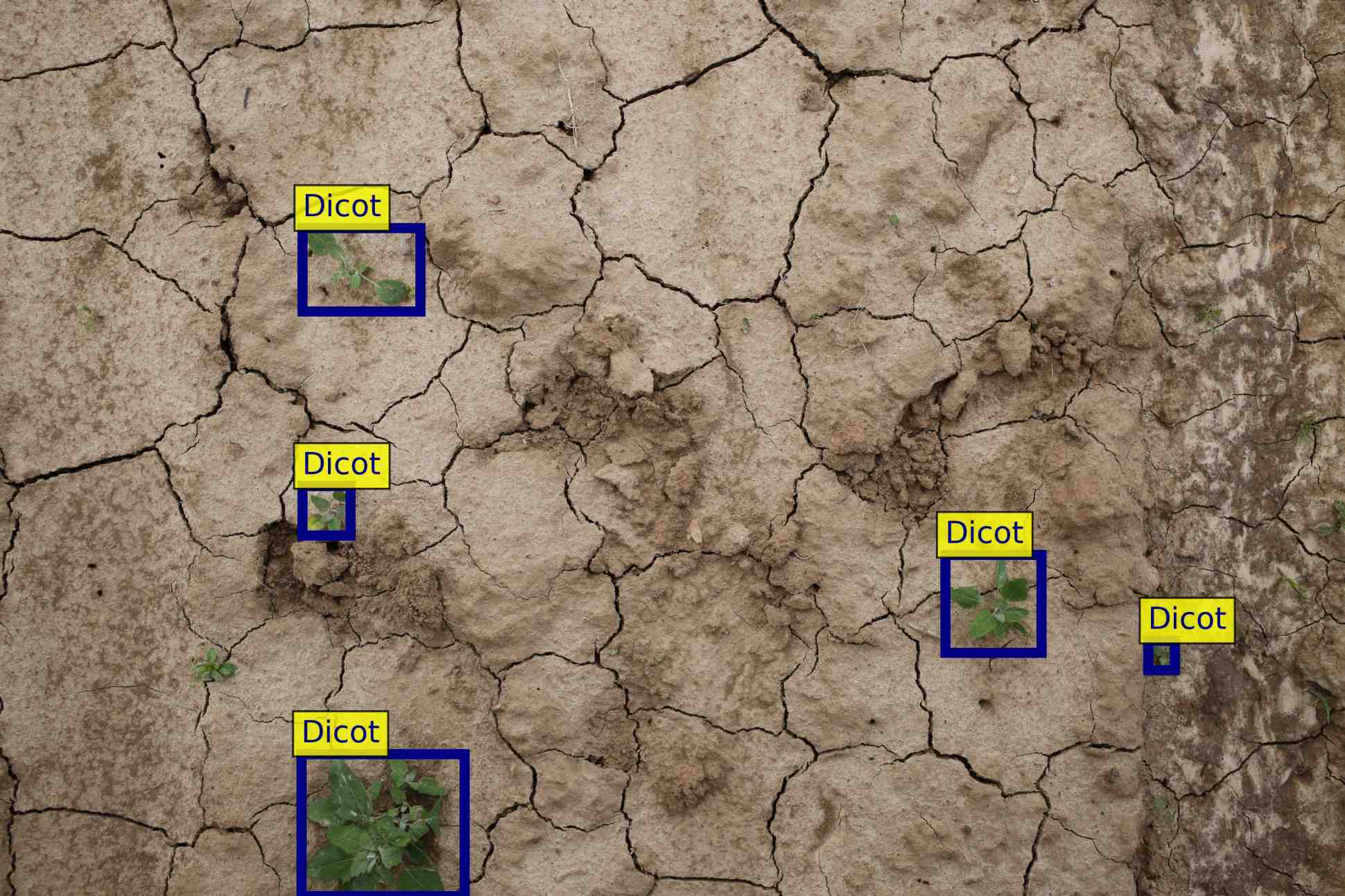}
        \caption{Dataset 2}
    \end{subfigure}

  \caption{Comparison of captured and annotated ground truth images for dataset 1 and dataset 2.}
  \label{fig:comparison_groundtruth_annotated}
\end{figure}

Fig. \ref{fig:comparison_groundtruth_annotated} illustrates the differences in the annotation scheme between dataset 1 and 2 as well as the ground-truth, the rest of the species are shown in the supplementary material.

\begin{figure}[H]
    \centering
    \begin{subfigure}{0.3\textwidth}
        \centering
        \includegraphics[width=\textwidth]{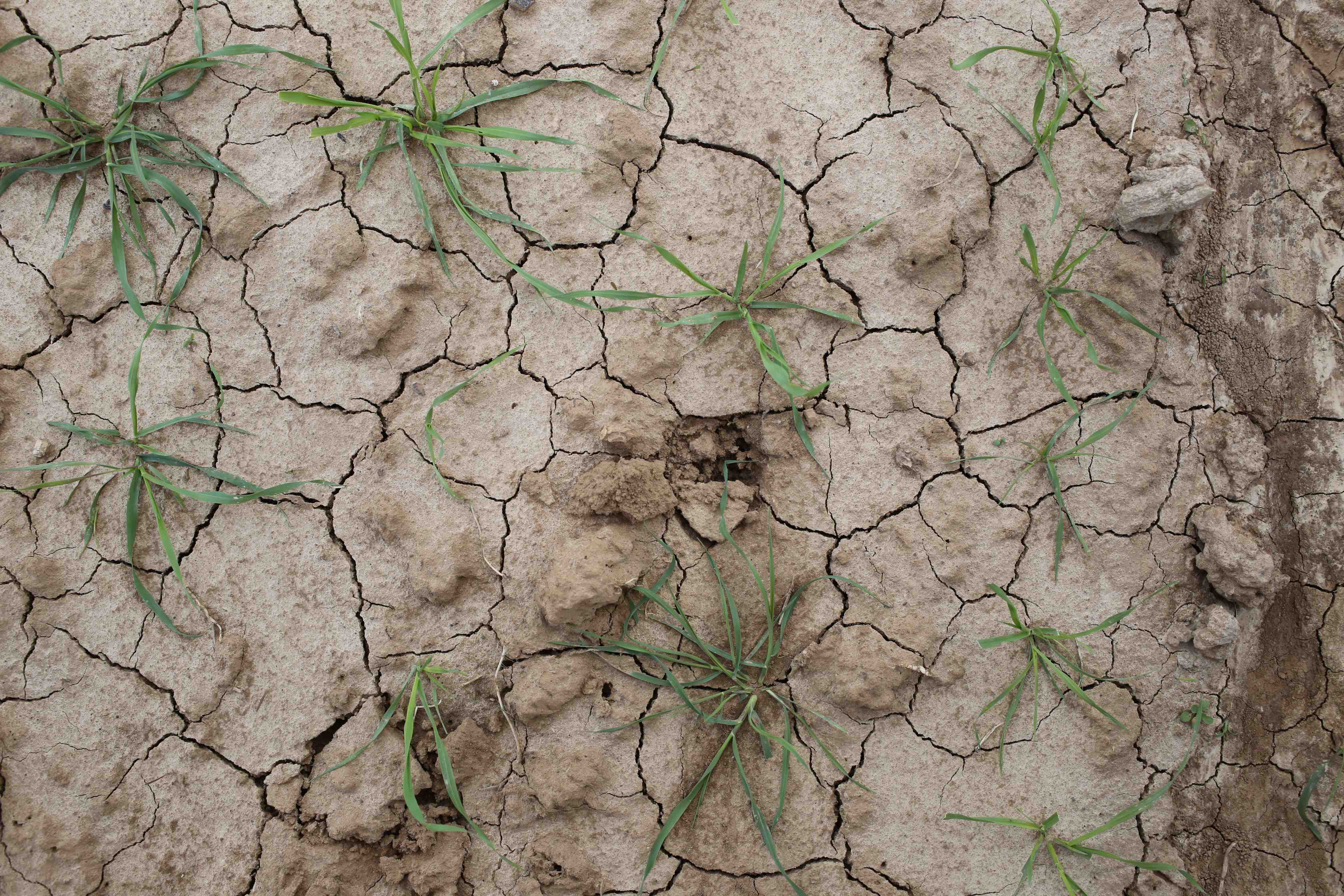}
        \caption{\textit{Avena fatua} L.}
    \end{subfigure}
    \begin{subfigure}{0.3\textwidth}
        \centering
        \includegraphics[width=\textwidth]{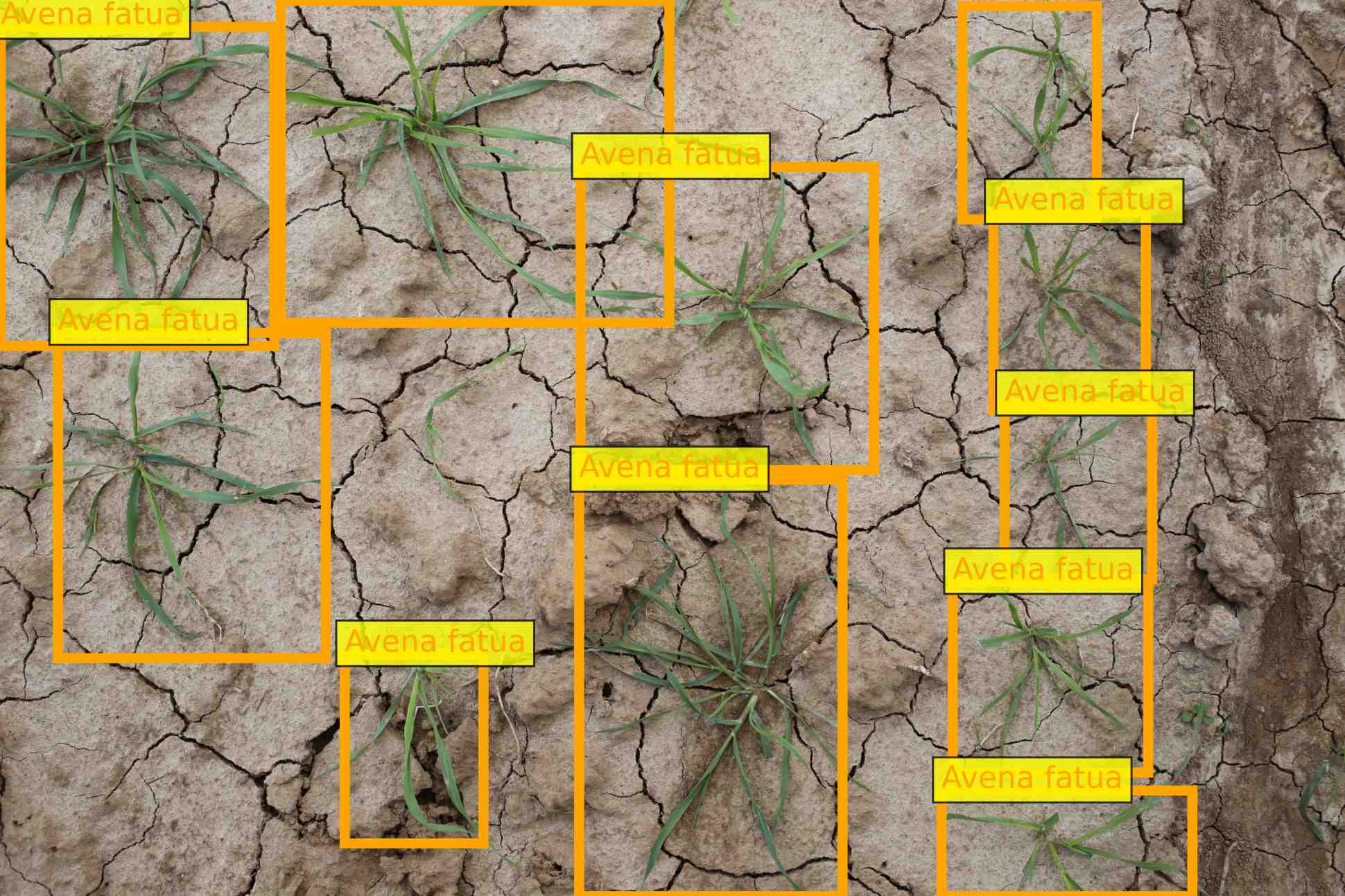}
        \caption{Dataset 1}
    \end{subfigure}
    \begin{subfigure}{0.3\textwidth}
        \centering
        \includegraphics[width=\textwidth]{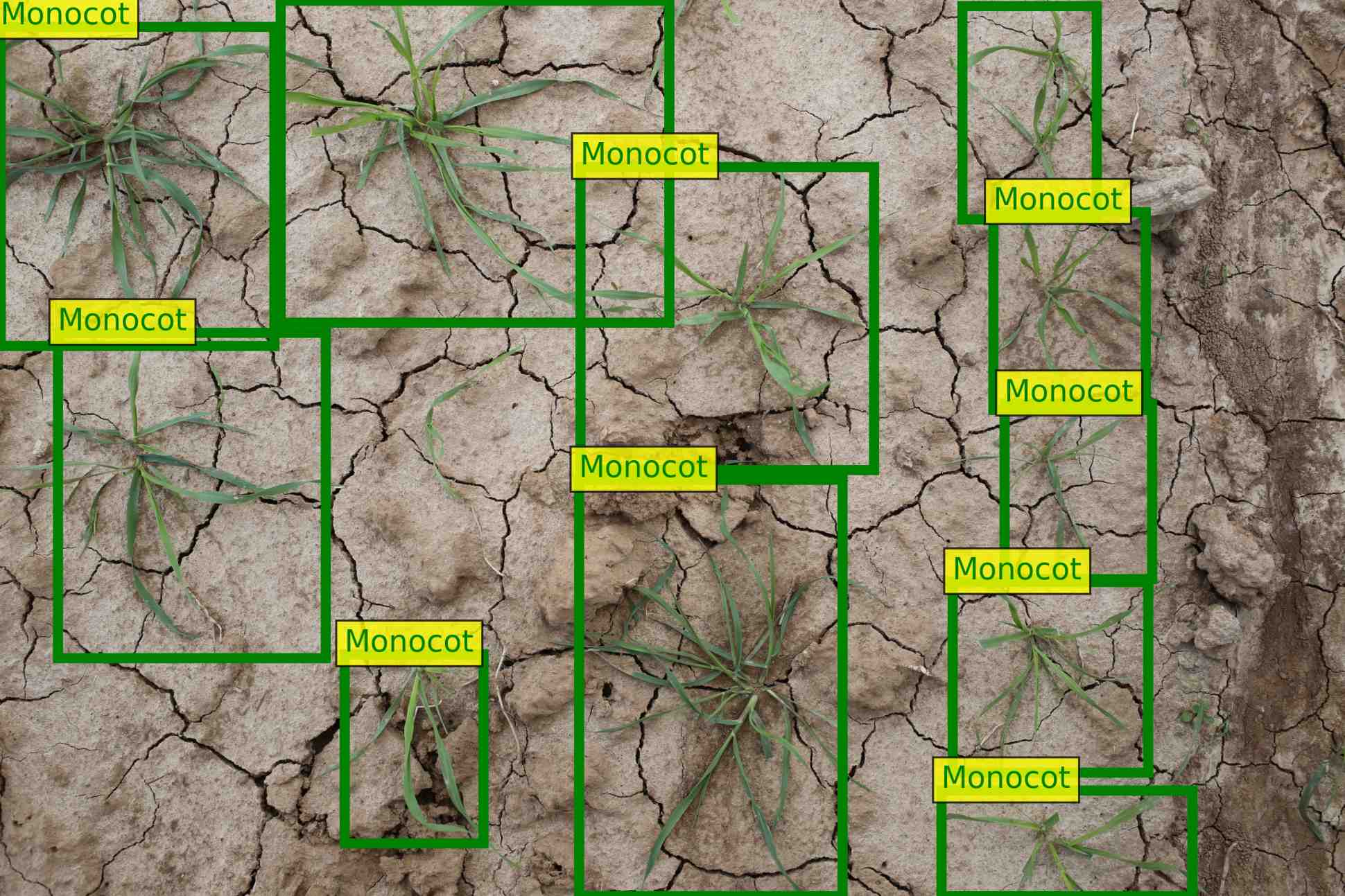}
        \caption{Dataset 2}
    \end{subfigure}

    \begin{subfigure}{0.3\textwidth}
        \centering
        \includegraphics[width=\textwidth]{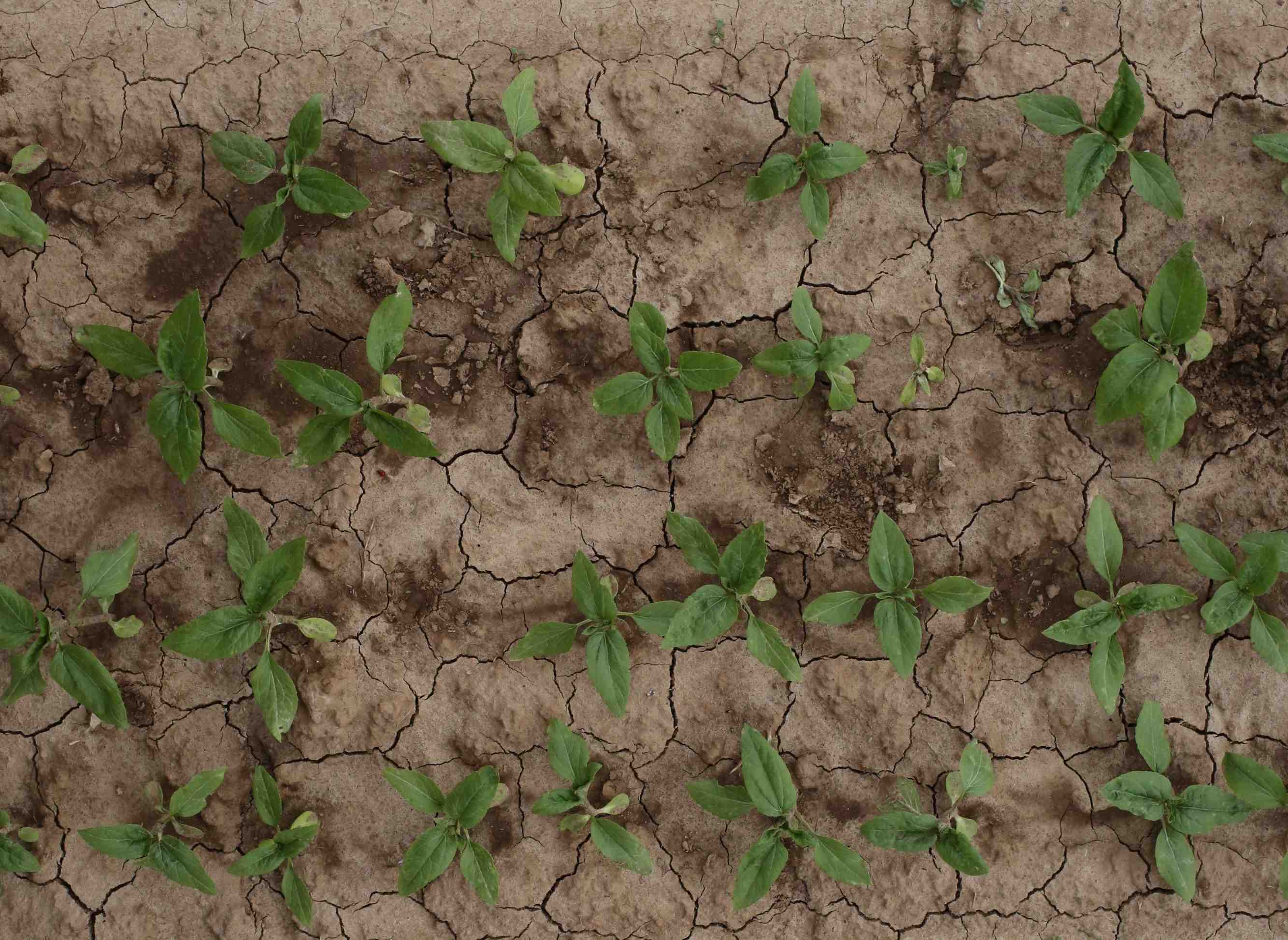}
        \caption{\textit{Helianthus annuus} L.}
    \end{subfigure}
    \begin{subfigure}{0.3\textwidth}
        \centering
        \includegraphics[width=\textwidth]{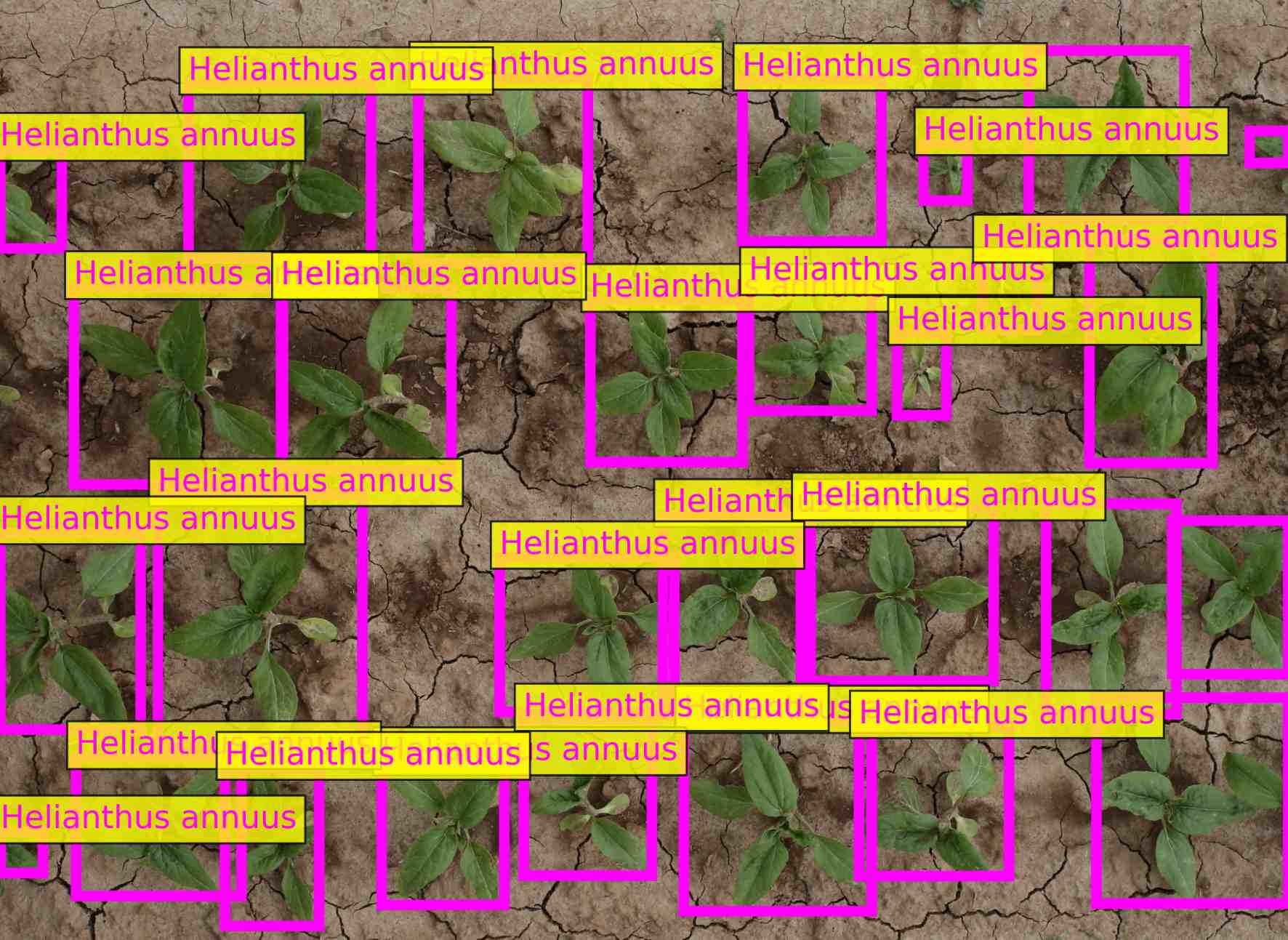}
        \caption{Dataset 1}
    \end{subfigure}
    \begin{subfigure}{0.3\textwidth}
        \centering
        \includegraphics[width=\textwidth]{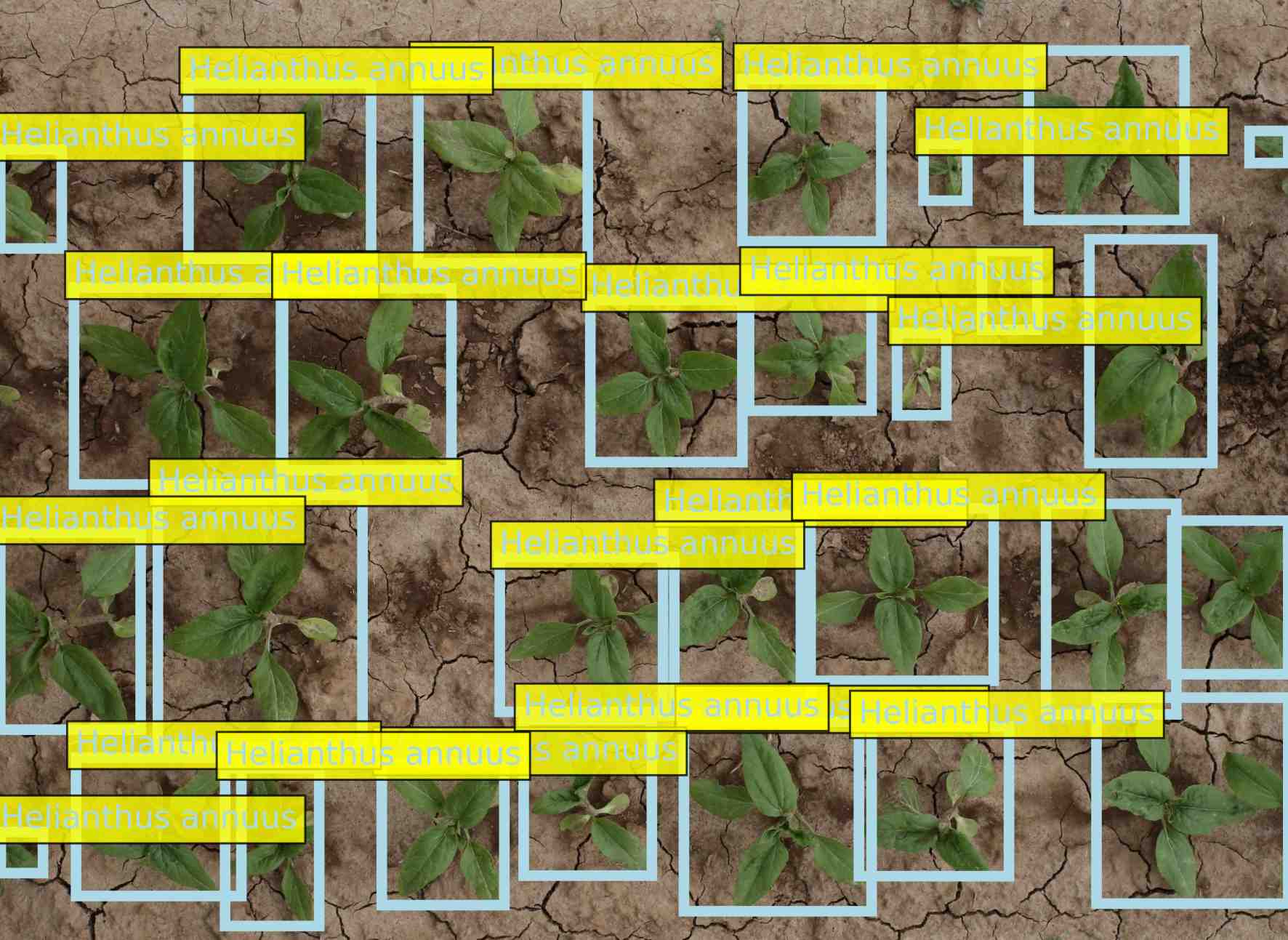}
        \caption{Dataset 2}
    \end{subfigure}

    \begin{subfigure}{0.3\textwidth}
        \centering
        \includegraphics[width=\textwidth]{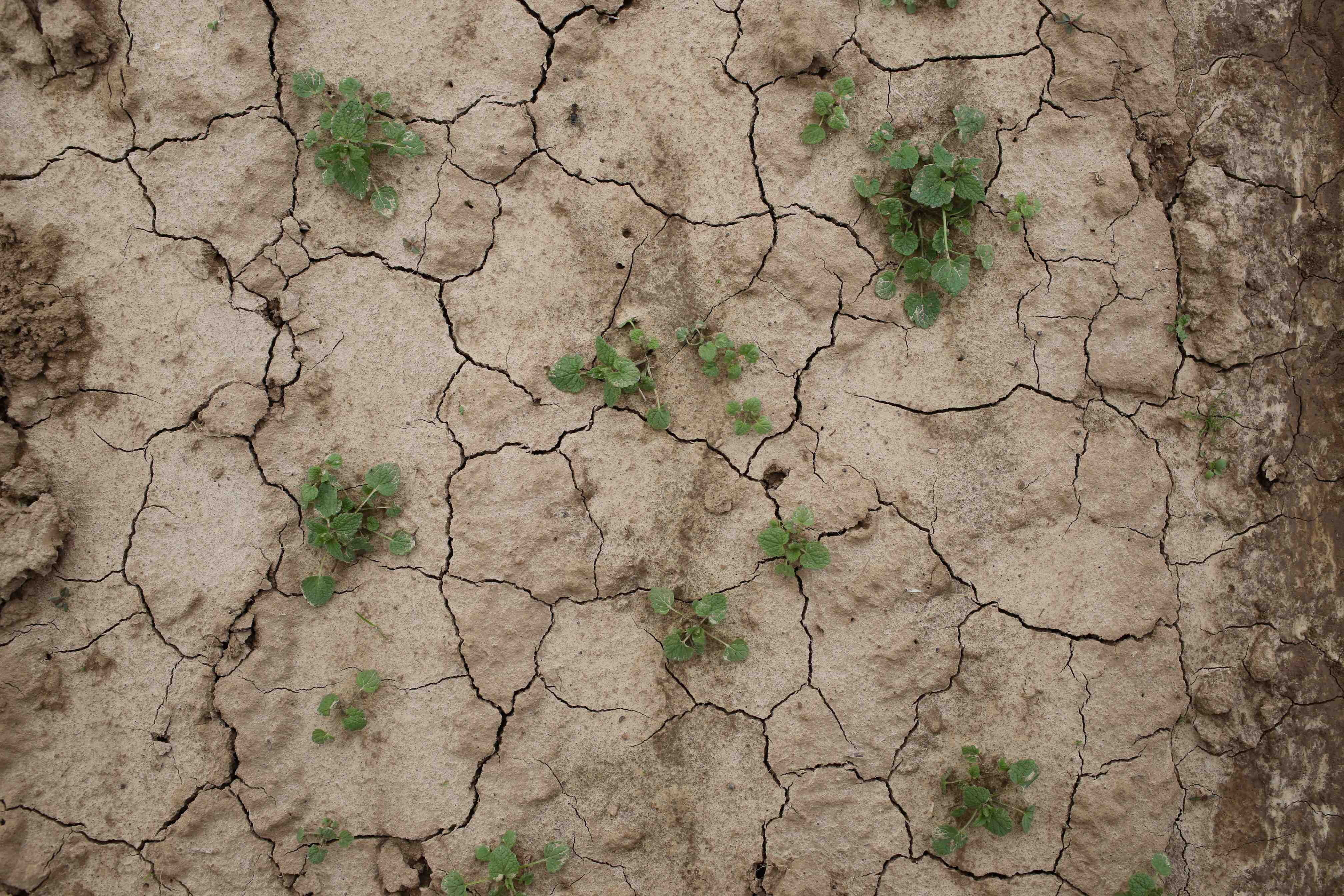}
        \caption{\textit{Lamium purpureum} L.}
    \end{subfigure}
    \begin{subfigure}{0.3\textwidth}
        \centering
        \includegraphics[width=\textwidth]{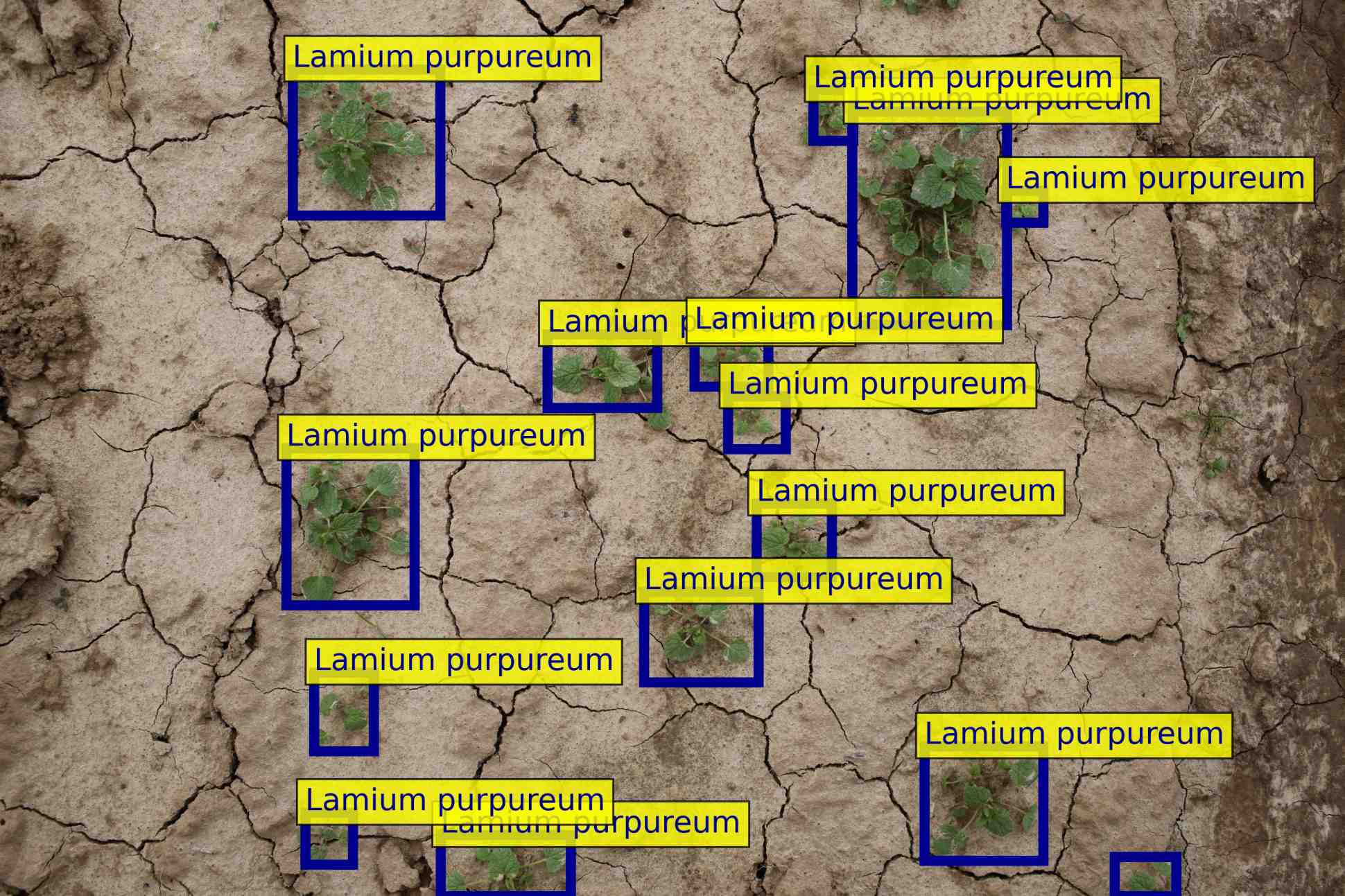}
        \caption{Dataset 1}
    \end{subfigure}
    \begin{subfigure}{0.3\textwidth}
        \centering
        \includegraphics[width=\textwidth]{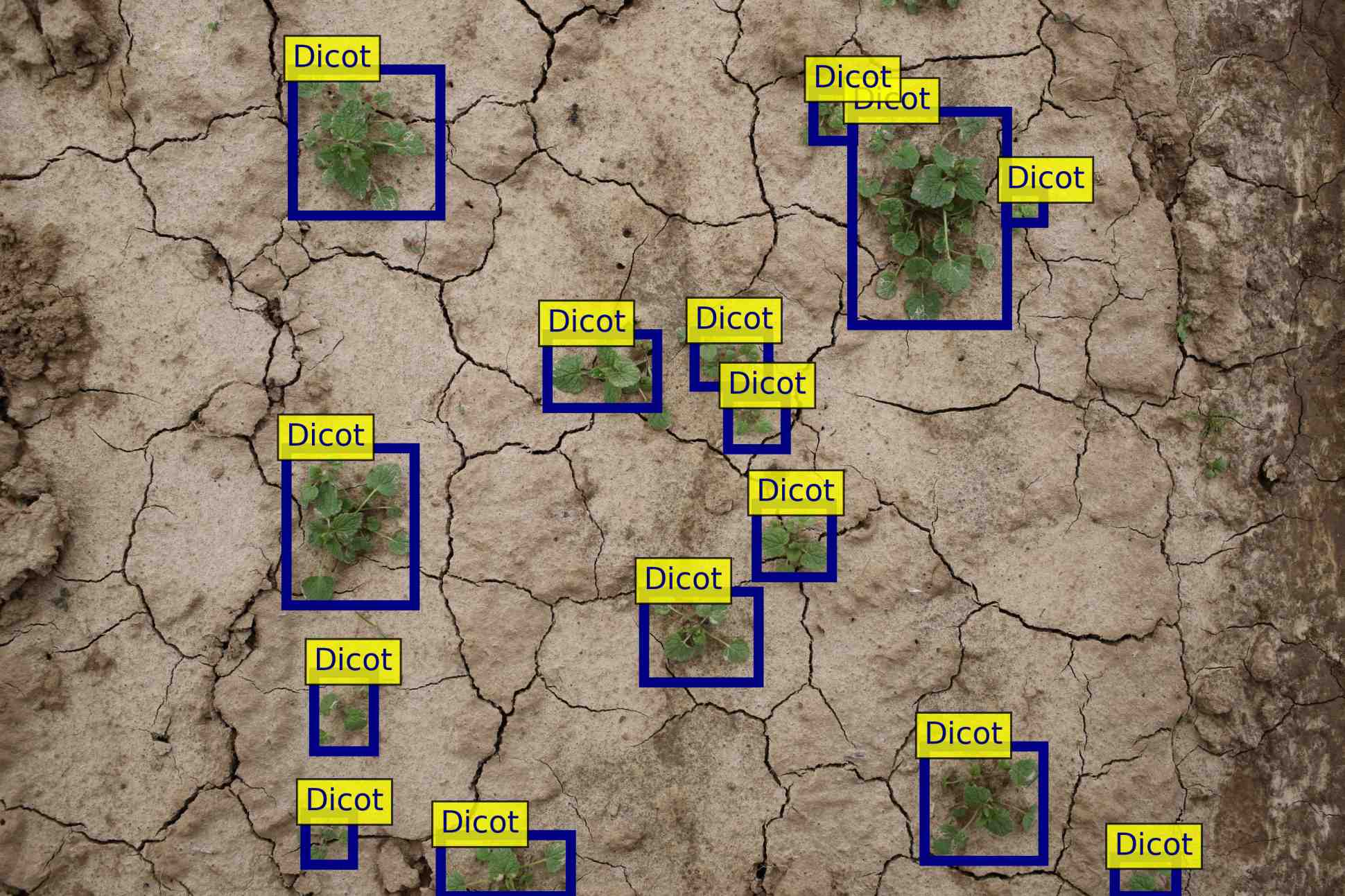}
        \caption{Dataset 2}
    \end{subfigure}

  \caption{Comparison of captured and annotated ground truth images for dataset 1 and dataset 2.}
  \label{fig:comparison_groundtruth_annotated3}
\end{figure}

\subsection{Models tested}
The full range of  YOLOv8, YOLOv9 and YOLOv10 versions available at the time were trained, as well as the RT-DETR-x and RT-DETR-l models. The choice of models was based on the criteria of operational suitability in real-time conditions, with the decision for one-stage detectors being made on this basis. The selection of RT-DETR was made because they represent the current state of the art for real-time Transformers. The characteristics of the different models are described in greater detail below.

\subsubsection{YOLOv8}

In contrast to its predecessors, YOLOv8 employs a  CNN comprising a backbone and a head. The backbone comprises a modified CSPDarknet53 with a C2f module, which facilitates accelerated calculations. The C2f module incorporates a Spatial Pyramid Pooling Fast (SPPF) layer. The head comprises a series of convolutional layers, with subsequent fully connected layers. The decoupled head permits the independent performance of object classification and regression tasks. The YOLOv8 model incorporates a self-attention mechanism within the head component. Furthermore, a feature pyramid network is incorporated, which facilitates the recognition of objects exhibiting multiple scales. The structure of YOLOv8 enables selective focus on different regions of an image \cite{hussain2023YOLO}. 

\subsubsection{YOLOv9}
The YOLOv9 model has been developed for the real-time recognition of objects. The Programmable Gradient Information mechanism (PGI mechanism) employed herein serves to obviate information bottlenecks, thereby resolving the issue of information loss, which is prevalent in DNNs. This mechanism ensures the preservation of crucial data across all available layers. This results in the generation of reliable gradients and enhanced model convergence, which is crucial for recognition tasks in particular. Another noteworthy aspect is the utilization of reversible functions. These are of particular importance for lightweight models, given that considerable data loss often occurs during processing. The reversible functions facilitate the inversion of data without loss and ensure the maintenance of information integrity throughout the entire network depth. Additionally, YOLOv9 employs the Generalised Efficient Layer Aggregation Network (GELAN), which enables flexible integration of diverse calculation blocks through parameter utilisation and calculation efficiency. The GELAN guarantees that YOLOv9 can adapt to a multitude of applications while maintaining its speed and accuracy in all cases \cite{ultralytics, wang2024YOLOv9}. 

\subsubsection{YOLOv10}

YOLOv10 represents a further advancement of the preceding models, particularly in terms of the more efficient architectural design. During the training phase, the non-maximal suppression technique has been removed. YOLOv10 employs a dual assignment strategy. Furthermore, the computational effort has been reduced, which has resulted in an increase in performance and the model accuracy has  been enhanced. This has enabled more efficient processing by YOLOv10, which has allowed the available resources to be utilised more effectively. The optimised architecture of YOLOv10 has been designed to achieve a balance between speed and accuracy, while maintaining a high level of precision. The model is also more efficient due to a reduction in the number of parameters and lower latency without affecting recognition performance \cite{ultralytics, wang2024YOLOv9}.

\subsubsection{Real-time Detection Transformer (RT-DETR)}

The RT-DETR, developed by Baidu, is based on the architectural principles of Transformers. It enhances the process of object recognition and facilitates utilisation in real-time scenarios. The incorporation of attention mechanisms represents a significant advantage of RT-DETR, as it enhances the ability to recognize objects in complex and diverse scenes. The mechanism enables the model to focus exclusively on the pertinent regions of the image, thus enhancing the accuracy of the recognition process. Consequently, the RT-DETR is suitable for high-speed processing with high accuracy. The reduction in latency and computing requirements permits the processing of even large and high-resolution images with optimal efficiency \cite{ultralytics, zhao2024detrs}. In this study, the RT-DETR-l and RT-DETR-x were selected for analysis. The RT-DETR-l employs 42 million parameters, while the RT-DETR-x utilizes 76 million \cite{zhao2024detrs}.

\subsection{Performance evaluation metrics}

In this study, performance of the trained models was assessed by utilizing common evaluation metrics for the object detection task \cite{padilla2020survey}. Next to the standard metrics mAP50 and mAP50-95 metrics were further assessed. 

In this context, it is necessary to utilize the metrics of precision and recall. Precision is defined as the ratio of true positive detections to the total number of positive detections. The term recall is defined as the ratio of true positive detections to the total number of actual positives. The calculation and plotting of precision and recall for varying confidence levels yields a precision-recall curve. The area under the curve (AUC) represents the average precision.
The Intersection over Union (IoU) threshold quantifies the degree of overlap between the predicted bounding box and the ground truth bounding box. Since precision, recall and average precision are all class-specific metrics, the mean average precision (mAP) at IoU-threshold 0.50 (mAP50) represents the mean value across all classes of the average precision (AP) when an IoU threshold value of 0.50 is selected. The mean average precision (mAP) from IoU 0.50 to 0.95 (mAP50-95) represents an extension of the mAP50 score. The AP is calculated for a range of IoU values, from 0.50 to 0.95, with a step size of 0.05 and again averaged over the resulting values. 
In the context of real-time applications, inference time, the time required for the model to localize and predict objects in a single image, is particular relevant. It reflects the speed at which a model processes input and generates an output. Inference time can be influenced by several factors, including the hardware used. This study compares different hardware architectures to determine their suitability for real-time applications, using the inference time of each architecture as key metric in the evaluation.

\subsection{Experimental setting}

\begin{figure}[H]
  \centering
  \includegraphics[width=1\textwidth]{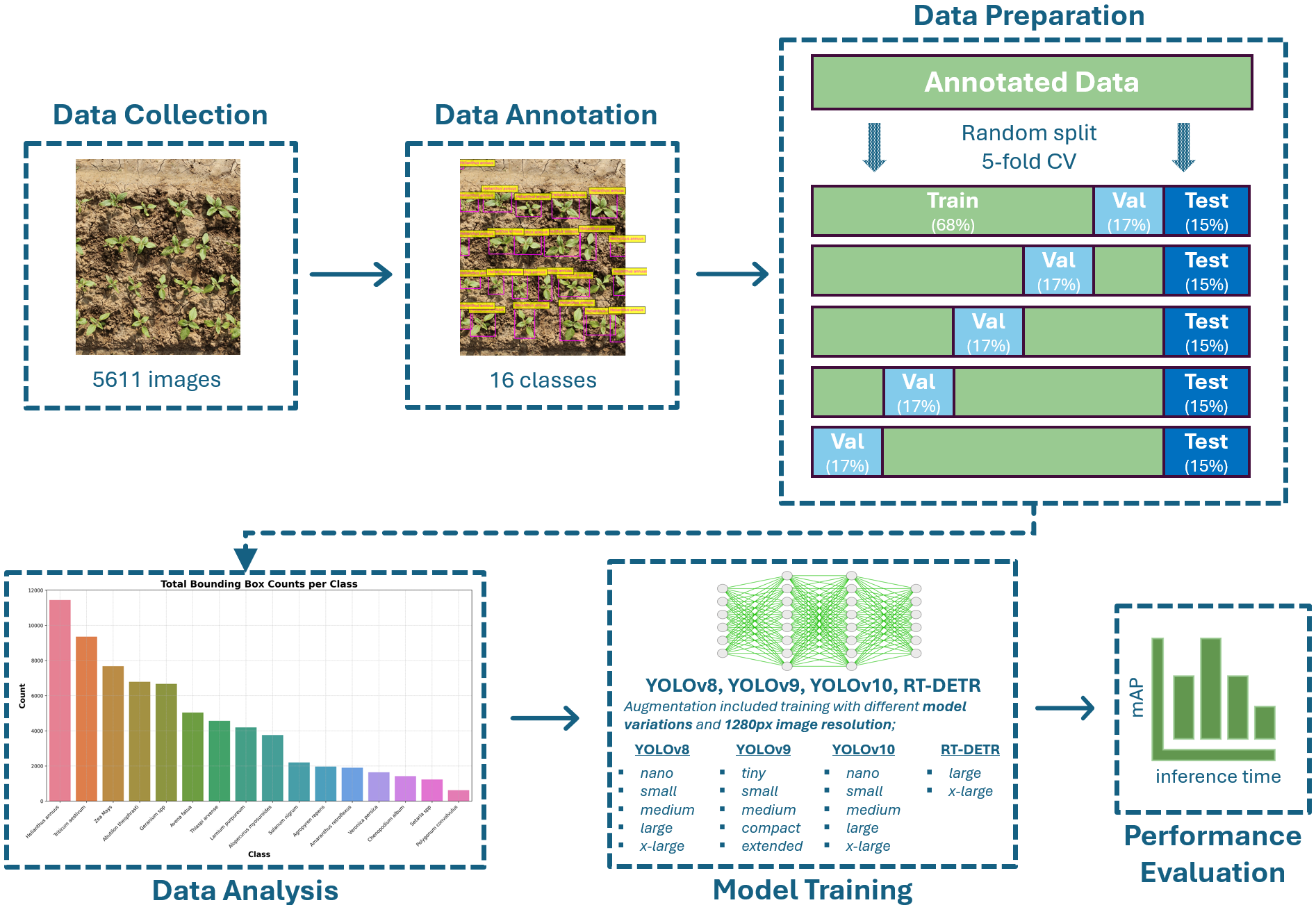}
  \caption{Machine Learning workflow applied in this study.}
  \label{fig:pipeline}
\end{figure}

The experimental workflow for this study can be summarized into six primary stages: Data Collection, Data Annotation, Data Preparation, Data Analysis, Model Training, and Performance Evaluation, as illustrated in Fig. \ref{fig:pipeline}. The first three stages are summarized in the previous sections.

The annotated data was analyzed to investigate the bounding box counts per class, ensuring a clear understanding of class distribution within the dataset. This step is crucial for identifying any class imbalances, which could potentially influence model performance and guide adjustments in the training process if necessary. It should be noted that although the number of images per species is balanced, the number of plants per image is not. Given that the images reflect a realistic field situation and that the number of plants per image can vary, it follows that the number of bounding boxes and consequently the number of instances per image also vary, which results in an imbalance. However, the Ultralytics framework mitigates the impact of class imbalance by employing focal loss \footnote{Ultralytics focal loss:\url{https://docs.ultralytics.com/reference/utils/loss/}, last accessed: 12-Sept-2024}, which reduces the influence of dominant classes. Furthermore, it applies a more extensive data augmentation \footnote{Ultralytics augmentation:\\ \url{https://docs.ultralytics.com/modes/train/\#augmentation-settings-and-hyperparameters}, last accessed: 12-Sept-2024} to underrepresented classes and less augmentation to overrepresented ones, thereby improving model robustness and overall performance.

The DNNs were trained on two different devices for separate datasets, i.e., dataset 1 and dataset 2. The workstation used for dataset 1 was equipped with a GeForce RTX 4090 GPU and an Intel(R) Core(TM) i9-13900K CPU @ 3.0 GHz, while the GPU server used for dataset 2 was equipped with an NVIDIA A100-SXM4-80GB GPU and 10 AMD EPYC 75F3 32-Core Processors.

The YOLO and RT-DETR models utilized were sourced from ultralytics framework \cite{ultralytics} with activated online augmentation during the training process as defined in \cite{ultralytics}\footnotemark[4] and using the COCO \cite{COCO_dataset}\footnote{Coco Dataset: \url{https://cocodataset.org/\#home}, last accessed: 12-Sept-2024}
pretrained weights for fine-tuning. Training for YOLO models and RT-DETR was performed with PyTorch 2.3.1 and CUDA 12.1 \cite{pytorch}, with a consistent image resolution of 1280 pixels. The training process comprised 150 epochs, with an early stopping patience of 50. A learning rate of 0.01 was used for YOLO and 0.001 for RT-DETR, along with a cosine learning rate decay. The optimizer used was ADAMw, which is recognized for its decoupled weight decay regularization \cite{ADAMw}. The selected hyperparameters were identified with the images of the validation set, in which a selected range of potential combinations has been tested. The parameters within this particular combination were found to yield the most favorable results in the prelimanary studies. Table  \ref{tab:hyperparameters} summarizes the identified hyperparameter configurations. The trained models were individually evaluated, maintaining a consistent image resolution of 1280 pixels. Efficiency evaluations and deployments (metrics vs. inference times) were performed on two different NVIDIA GPU models (RTX 3090 and RTX 4090), as well as two distinct CPU models: the Intel Core i9-14900K (32-core) and AMDRyzen 9 5950X (16-core). To optimize real-time deployment efficiency, trained model files were converted to TensorRT\footnote{TensorRT: \url{https://developer.nvidia.com/tensorrt}, last accessed: 15-Sept-2024} for GPU inferences, OpenVINO\footnote{Openvino: \url{https://docs.openvino.ai/2024/index.html}, last accessed: 15-Sept-2024} for the Intel CPU, and ONNX\footnote{ONNX: \url{https://onnx.ai/}, last accessed: 15-Sept-2024} for the AMD CPU.

\begin{table}[H]
  \caption{Hyperparameters used for this study}
  \label{tab:hyperparameters}
  \centering
  \begin{tabular}{@{}ll@{}}
    \toprule
    \textbf{Hyperparameter} & \textbf{Value}\\
    \midrule
    Epoch & 150\\
    Early stopping with patience & 50\\
    Learning rate & 0.01 for YOLOs \& 0.001 for RT-DETR\\
    Batch size & 4 for Dataset 1 \& 8 for Dataset 2 \\
    Optimizer & ADAMw (Decoupled Weight Decay Regularization)\\
    Learning rate scheduler & Cosine\\
    \bottomrule
  \end{tabular}
\end{table}

\section{Results}

\subsection{Average performance comparison of YOLOs and RT-DETR}

\begin{figure}[H]
  \centering
  \includegraphics[width=1\textwidth]{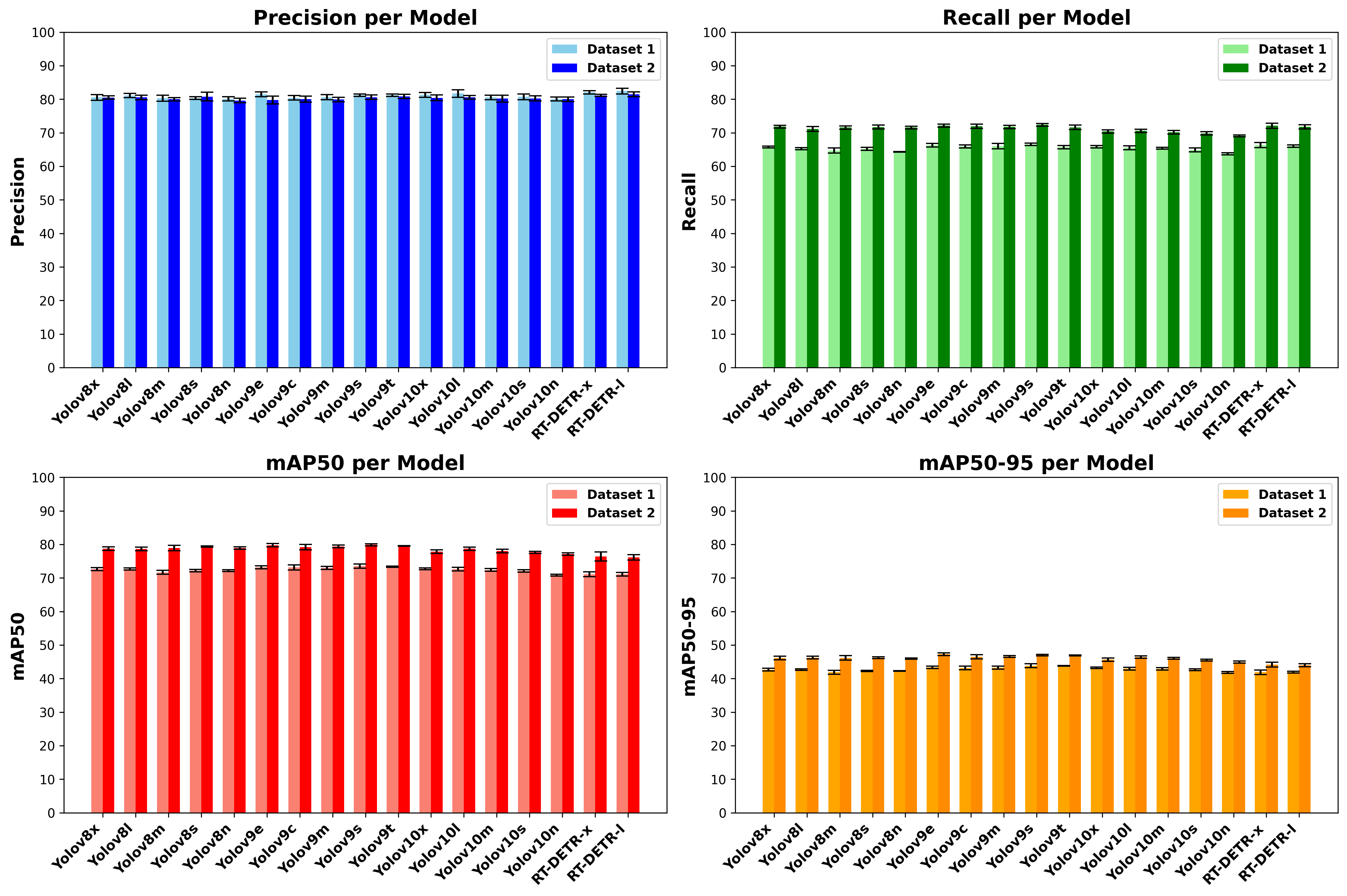}
  \caption{Comparison of performance metrics across datasets with 5-fold cross validation, showing Precision, Recall, mAP50, and mAP50-95 for datasets 1 and 2.}
  \label{fig:performance_metrics}
\end{figure}

The results of the weed detection models are presented in Fig. \ref{fig:performance_metrics}, Table \ref{tab:performance_YOLO_dataset1} and \ref{tab:performance_YOLO_dataset2}, which provide a comprehensive overview of the performance across different metrics and the two datasets. 

In this subsection, the terms precision and recall are computed as the average across all classes.  The precision scores across the models were consistently high for both datasets reaching values from 80.08\% (YOLOv10n) to 82.44\% (RT-DETR-l) for dataset 1. In particular, the RT-DETR models (RT-DETR-x and RT-DETR-l) obtained the highest precision scores, indicating their strong ability to minimize false positives. Similarly, in dataset 2, precision scores were similar, ranging from 79.62\% (YOLOv8n) to 81.46\% (RT-DETR-l). 
Recall scores showed some variability, reflecting the models' ability to detect weed and crop species across the datasets. In dataset 1, recall ranged from 63.76\% (YOLOv10n) to 66.58\% (YOLOv9s). YOLOv9s and YOLOv9e models consistently achieved higher recall, indicating their effectiveness in identifying a larger proportion of weed instances, even at the risk of increasing false positives. In dataset 2, recall scores improved significantly, with values ranging from 69.10\% (YOLOv10n) to 72.36\% (YOLOv9s). These results indicate that the YOLOv9s model is particularly effective in capturing most weed instances either being encoded species-wise or categorized in the broader classes monocot and dicot, although it may come at the cost of slightly lower precision compared to the RT-DETR models.

The mAP50 metric, which measures the overall detection accuracy at a 50\% IoU threshold, showed high performance across all models. For dataset 1, mAP50 scores ranged from 70.82\% (YOLOv10n) to 73.52\% (YOLOv9s), with YOLOv9s standing out as the superior model. In dataset 2, the mAP50 scores were generally higher, ranging from 76.14\% (RT-DETR-l) to 79.86\% (YOLOv9s). The consistent performance of YOLOv9s across both datasets highlights its robustness and effectiveness in detecting weeds, particularly at a lower IoU threshold. The increased mAP50 scores in dataset 2 for nearly all models indicate a stronger overall detection capability in this dataset.

The mAP50-95 metric, which averages the detection accuracy over a range of IoU thresholds, revealed more pronounced differences among the models. In dataset 1, mAP50-95 scores ranged from 41.84\% (YOLOv10n) to 43.82\% (YOLOv9s). The YOLOv9 series, particularly YOLOv9s and YOLOv9e, showed superior performance, indicating their ability to maintain accuracy across varying detection strictness levels. In dataset 2, mAP50-95 scores were higher, ranging from 43.98\% (RT-DETR-l) to 47.26\% (YOLOv9e). The YOLOv9 models again demonstrated their strength, with YOLOv9e slightly outperforming YOLOv9s in this more stringent evaluation metric. The elevated mAP50-95 scores across all models in dataset 2 indicates that the models demonstrated enhanced generalisation capabilites and maintain detection accuracy across a broader range of IoU thresholds within this dataset.

\begin{table}[H]
\caption{Performance comparison of YOLO and RT-DETR models on dataset 1, showing Precision, Recall, mAP50, and mAP50-95 (in percentages), with standard deviations (+/- SD) indicated. Bold values represent the highest values.}
\label{tab:performance_YOLO_dataset1}
\resizebox{\textwidth}{!}{%
\begin{tabular}{c|c|c|c|c}
\toprule
\textbf{Model} & \textbf{Precision} & \textbf{Recall} & \textbf{mAP50} & \textbf{mAP50-95} \\
\midrule
YOLOv8x & 80.54 ± 0.84 & 65.76 ± 0.27 & 72.64 ± 0.44 & 42.64 ± 0.40 \\ 
YOLOv8l & 81.12 ± 0.64 & 65.30 ± 0.32 & 72.66 ± 0.30 & 42.70 ± 0.21 \\ 
YOLOv8m & 80.34 ± 0.89 & 64.70 ± 0.76 & 71.68 ± 0.58 & 41.90 ± 0.58 \\ 
YOLOv8s & 80.34 ± 0.42 & 65.22 ± 0.48 & 72.18 ± 0.35 & 42.26 ± 0.21 \\ 
YOLOv8n & 80.10 ± 0.64 & 64.34 ± 0.09 & 72.20 ± 0.29 & 42.30 ± 0.10 \\ 
YOLOv9e & 81.48 ± 0.72 & 66.28 ± 0.56 & 73.18 ± 0.43 & 43.32 ± 0.36 \\ 
YOLOv9c & 80.44 ± 0.67 & 65.92 ± 0.47 & 73.16 ± 0.77 & 43.16 ± 0.52 \\ 
YOLOv9m & 80.62 ± 0.75 & 66.02 ± 0.83 & 72.98 ± 0.44 & 43.30 ± 0.45 \\ 
YOLOv9s & 81.18 ± 0.35 & \textbf{66.58 ± 0.35} & \textbf{73.52 ± 0.63} & \textbf{43.82 ± 0.58} \\ 
YOLOv9t & 81.22 ± 0.35 & 65.72 ± 0.50 & 73.34 ± 0.18 & 43.80 ± 0.07 \\ 
YOLOv10x & 81.32 ± 0.70 & 65.82 ± 0.37 & 72.70 ± 0.25 & 43.20 ± 0.25 \\ 
YOLOv10l & 81.70 ± 1.12 & 65.54 ± 0.55 & 72.64 ± 0.56 & 42.98 ± 0.41 \\ 
YOLOv10m & 80.54 ± 0.64 & 65.36 ± 0.27 & 72.38 ± 0.40 & 42.88 ± 0.34 \\ 
YOLOv10s & 80.74 ± 0.86 & 64.88 ± 0.60 & 72.06 ± 0.36 & 42.64 ± 0.29 \\ 
YOLOv10n & 80.08 ± 0.58 & 63.76 ± 0.32 & 70.82 ± 0.26 & 41.84 ± 0.29 \\ 
RT-DETR-x & 82.06 ± 0.49 & 66.34 ± 0.80 & 71.08 ± 0.73 & 41.88 ± 0.65 \\ 
RT-DETR-l & \textbf{82.44 ± 0.84} & 66.02 ± 0.33 & 71.10 ± 0.51 & 41.88 ± 0.27 \\ 
\bottomrule
\end{tabular}
}
\end{table}

\begin{table}[H]
\caption{Performance comparison of YOLO and RT-DETR models on dataset 2, showing Precision, Recall, mAP50, and mAP50-95 (in percentages), with standard deviations (+/- SD) indicated. Bold values represent the highest values.}
\label{tab:performance_YOLO_dataset2}
\resizebox{\textwidth}{!}{%
\begin{tabular}{c|c|c|c|c}
\toprule
\textbf{Model} & \textbf{Precision} & \textbf{Recall} & \textbf{mAP50} & \textbf{mAP50-95} \\
\midrule
YOLOv8x & 80.50 ± 0.49 & 71.84 ± 0.42 & 78.74 ± 0.50 & 46.14 ± 0.55 \\ 
YOLOv8l & 80.54 ± 0.68 & 71.14 ± 0.72 & 78.68 ± 0.53 & 46.26 ± 0.44 \\ 
YOLOv8m & 80.02 ± 0.51 & 71.56 ± 0.50 & 78.92 ± 0.82 & 46.16 ± 0.65 \\ 
YOLOv8s & 80.78 ± 1.32 & 71.72 ± 0.59 & 79.36 ± 0.19 & 46.20 ± 0.25 \\ 
YOLOv8n & 79.62 ± 0.70 & 71.58 ± 0.39 & 78.92 ± 0.33 & 45.94 ± 0.21 \\ 
YOLOv9e & 79.78 ± 1.14 & 72.14 ± 0.44 & 79.76 ± 0.52 & \textbf{47.26 ± 0.38} \\ 
YOLOv9c & 80.04 ± 0.91 & 71.96 ± 0.60 & 79.18 ± 0.82 & 46.52 ± 0.64 \\ 
YOLOv9m & 79.92 ± 0.69 & 71.74 ± 0.46 & 79.40 ± 0.40 & 46.58 ± 0.29 \\ 
YOLOv9s & 80.62 ± 0.64 & \textbf{72.36 ± 0.43} & \textbf{79.86 ± 0.30} & 47.00 ± 0.24 \\ 
YOLOv9t & 80.82 ± 0.62 & 71.62 ± 0.65 & 79.56 ± 0.09 & 46.90 ± 0.10 \\ 
YOLOv10x & 80.44 ± 0.82 & 70.40 ± 0.51 & 77.82 ± 0.54 & 45.64 ± 0.53 \\ 
YOLOv10l & 80.56 ± 0.52 & 70.58 ± 0.47 & 78.72 ± 0.50 & 46.40 ± 0.35 \\ 
YOLOv10m & 80.20 ± 1.03 & 70.18 ± 0.54 & 78.04 ± 0.53 & 46.04 ± 0.23 \\ 
YOLOv10s & 80.24 ± 0.75 & 69.82 ± 0.51 & 77.62 ± 0.30 & 45.52 ± 0.25 \\ 
YOLOv10n & 80.00 ± 0.69 & 69.10 ± 0.29 & 77.14 ± 0.36 & 44.92 ± 0.28 \\ 
RT-DETR-x & 81.16 ± 0.33 & 72.10 ± 0.78 & 76.40 ± 1.37 & 44.14 ± 0.73 \\ 
RT-DETR-l & \textbf{81.46 ± 0.72} & 71.78 ± 0.59 & 76.14 ± 0.84 & 43.98 ± 0.48 \\ 
\bottomrule
\end{tabular}
}
\end{table}

\subsection{Class-wise Results}

In this section,  a detailed analysis of the class-wise performance of four weed detection models YOLOv8l, YOLOv9c, YOLOv10l and RT-DETR-l in dataset 1 is presented. These models were chosen for their comparable size, and the results highlight their precision, recall, mAP50, and mAP50-95 metrics across various weed species. The corresponding results for each model can be found in Tables \ref{tab:performance_YOLOv8l_class_wise_dataset1}, \ref{tab:performance_YOLOv9c_class_wise_dataset1}, \ref{tab:performance_YOLOv10l_class_wise_dataset1}, and \ref{tab:performance_rt_detr_l_class_wise_dataset1}. The results for other models are available in the Supplementary Material.

\subsubsection{Dataset 1}

The YOLOv8l model demonstrated strong average performance in all weed species, with an average precision of 81.12\% and an average recall of 65.30\%. The model excelled particularly in detecting the monocot \textit{A. myosuroides}, achieving a precision of 91.64\% and a recall of 81.28\%. The highest AP50 of 88.26\% was also observed for\textit{A. myosuroides}. However, the model showed relatively lower performance in detecting the monocot \textit{Setaria spp.}, with a precision of 76.64\% and a recall of 41.14\%, resulting in a AP50-95 of 29.22\%. This indicates a challenge in accurately detecting instances of \textit{Setaria spp.} under more stringent IoU thresholds.

The YOLOv9c model exhibited a slightly higher overall precision compared to YOLOv8l, with an average precision of 80.44\% and an average recall of 65.92\%. Similarly to YOLOv8l, YOLOv9c also performed exceptionally well on \textit{A. myosuroides}, achieving a precision of 91.54\% and a recall of 82.70\%. Additionally, YOLOv9c showed notable improvement in detecting the dicot crop \textit{H. annuus}, with a recall of 75.84\% and a AP50 of 84.60\%. However, \textit{Setaria spp.} remained challenging, with AP50-95 falling to 30.62\%, indicating a need for better handling of this class across different IoU thresholds.

The YOLOv10l model achieved an average precision of 81.70\% and an average recall of 65.54\% on average across all classes. Its performance on \textit{A. myosuroides} showed again a robust level of performance, with precision reaching 90.44\% and recall at 82.58\%. YOLOv10l showed improved performance in detecting the dicot \textit{C. album} with a precision of 81.66\% and a AP50-95 of 41.00\%, surpassing the previous models. However, the dicot \textit{F. convolvulus} was more difficult to recognize, with a lower precision of 77.32\% and AP50-95 of 44.34\%, suggesting room for improvement in handling this class.

The RT-DETR-l model demonstrated the highest average precision at 82.44\%, paired with an average recall of 66.02\%, making it the best performer among the four models. The model excelled in detecting the dicot \textit{A. retroflexus} with a precision of 88.00\% and a AP50 of 68.92\%, highlighting its robustness in handling this species. The RT-DETR-l model also achieved the highest AP50-95 for \textit{H. annuus} at 57.06\%, indicating strong performance under stricter detection criteria. However, similar to the other models, \textit{Setaria spp.} remained a challenge, with a lower AP50-95 of 28.30\%.

\begin{table}[H]
\caption{Performance of YOLOv8l across images and instances per species, with Precision, Recall, AP50, and AP50-95 provided in percentages, with standard deviations (+/- SD) indicated. Bold values represent the highest values.}
\label{tab:performance_YOLOv8l_class_wise_dataset1}
\resizebox{\textwidth}{!}{%
\begin{tabular}{c|c|c|c|c|c|c}
\toprule
\textbf{Class} & \textbf{Images} & \textbf{Instances} & \textbf{Precision} & \textbf{Recall} & \textbf{AP50} & \textbf{AP50-95} \\
\midrule
\textit{Abutilon theophrasti} Medik. & 48 & 1166 & 76.14 ± 1.69 & 76.22 ± 1.82 & 80.18 ± 0.75 & 45.00 ± 0.77 \\ 
\textit{Elymus repens} (L.) Gould & 53 & 309 & 72.28 ± 2.49 & 72.88 ± 1.62 & 75.08 ± 1.84 & 35.54 ± 0.38 \\ 
\textit{Alopecurus myosuroides} Huds. & 77 & 635 & \textbf{91.64 ± 1.06} & 81.28 ± 0.81 & \textbf{88.26 ± 0.86} & 48.84 ± 1.33 \\ 
\textit{Amaranthus retroflexus} L. & 94 & 418 & 87.20 ± 0.87 & 58.72 ± 0.71 & 68.94 ± 1.06 & 51.54 ± 1.23 \\ 
\textit{Avena fatua} L. & 77 & 1131 & 81.56 ± 2.18 & 64.64 ± 1.09 & 74.52 ± 1.09 & 40.68 ± 1.15 \\ 
\textit{Chenopodium album} L. & 101 & 329 & 76.44 ± 3.05 & 55.10 ± 1.55 & 65.04 ± 0.95 & 39.40 ± 0.46 \\ 
\textit{Geranium spp.} & 53 & 1347 & 81.18 ± 0.78 & 56.54 ± 2.63 & 65.62 ± 2.91 & 30.86 ± 1.22 \\ 
\textit{Helianthus annuus} L. & 56 & 1873 & 80.94 ± 2.02 & 75.36 ± 0.57 & 83.70 ± 1.23 & \textbf{60.86 ± 1.06} \\ 
\textit{Lamium purpureum} L. & 214 & 1069 & 89.96 ± 1.55 & 57.72 ± 1.42 & 69.08 ± 1.21 & 40.08 ± 0.58 \\ 
\textit{Fallopia convolvulus} (L.) Á.  Löve & 78 & 115 & 72.12 ± 4.46 & 58.62 ± 1.82 & 62.48 ± 1.40 & 44.64 ± 1.03 \\ 
\textit{Setaria spp.} & 134 & 275 & 76.64 ± 4.70 & 41.14 ± 2.81 & 49.16 ± 2.10 & 29.22 ± 1.63 \\ 
\textit{Solanum nigrum} L. & 57 & 331 & 79.26 ± 3.17 & \textbf{82.84 ± 1.40} & 86.12 ± 0.84 & 52.40 ± 1.19 \\ 
\textit{Thlaspi arvense} L. & 273 & 1208 & 88.52 ± 1.71 & 51.46 ± 1.46 & 64.96 ± 0.55 & 42.16 ± 0.62 \\ 
\textit{Triticum aestivum} L. & 58 & 1772 & 74.04 ± 1.92 & 66.86 ± 1.44 & 71.78 ± 1.94 & 31.76 ± 1.36 \\ 
\textit{Veronica persica} Poir. & 108 & 322 & 86.24 ± 0.65 & 67.72 ± 1.32 & 72.70 ± 0.98 & 39.22 ± 1.23 \\ 
\textit{Zea mays} L. & 65 & 1327 & 83.74 ± 1.24 & 77.64 ± 1.62 & 84.74 ± 0.91 & 51.44 ± 0.43 \\

\bottomrule
\end{tabular}

}
\end{table}

\begin{table}[H]
\caption{Performance of YOLOv9c across images and instances per species, with Precision, Recall, AP50, and AP50-95 provided in percentages,  with standard deviations (+/- SD) indicated. Bold values represent the highest values..}
\label{tab:performance_YOLOv9c_class_wise_dataset1}
\resizebox{\textwidth}{!}{%
\begin{tabular}{c|c|c|c|c|c|c}
\toprule
\textbf{Class} & \textbf{Images} & \textbf{Instances} & \textbf{Precision} & \textbf{Recall} & \textbf{AP50} & \textbf{AP50-95} \\
\midrule
\textit{Abutilon theophrasti} Medik. & 48 & 1166 & 74.96 ± 1.47 & 76.02 ± 0.40 & 79.30 ± 1.42 & 44.38 ± 1.21 \\ 
\textit{Elymus repens} (L.) Gould & 53 & 309 & 69.70 ± 2.14 & 72.48 ± 4.04 & 74.42 ± 2.38 & 35.52 ± 1.81 \\ 
\textit{Alopecurus myosuroides} Huds. & 77 & 635 & \textbf{91.54 ± 0.94} & \textbf{82.70 ± 0.94} & \textbf{88.64 ± 0.69} & 49.36 ± 1.45 \\ 
\textit{Amaranthus retroflexus} L. & 94 & 418 & 84.44 ± 1.23 & 60.46 ± 1.90 & 69.58 ± 1.30 & 51.74 ± 0.84 \\ 
\textit{Avena fatua} L. & 77 & 1131 & 76.58 ± 2.49 & 65.44 ± 1.43 & 73.46 ± 1.21 & 40.42 ± 0.82 \\ 
\textit{Chenopodium album} L. & 101 & 329 & 77.00 ± 3.42 & 58.62 ± 2.50 & 67.68 ± 1.47 & 41.08 ± 1.08 \\ 
\textit{Geranium spp.} & 53 & 1347 & 81.30 ± 2.07 & 59.46 ± 1.65 & 68.70 ± 1.92 & 32.60 ± 1.04 \\ 
\textit{Helianthus annuus} L. & 56 & 1873 & 80.54 ± 1.91 & 75.84 ± 0.84 & 84.60 ± 0.91 & \textbf{61.78 ± 0.45} \\ 
\textit{Lamium purpureum} L. & 214 & 1069 & 89.26 ± 1.70 & 58.24 ± 1.93 & 70.50 ± 1.18 & 41.36 ± 0.77 \\ 
\textit{Fallopia convolvulus} (L.) Á.  Löve & 78 & 115 & 74.30 ± 5.58 & 57.56 ± 3.17 & 61.44 ± 2.05 & 42.94 ± 1.91 \\ 
\textit{Setaria spp.} & 134 & 275 & 76.82 ± 3.49 & 42.18 ± 2.29 & 50.18 ± 2.37 & 30.62 ± 0.93 \\ 
\textit{Solanum nigrum} L. & 57 & 331 & 78.40 ± 4.61 & 81.96 ± 1.19 & 85.50 ± 0.99 & 51.24 ± 0.59 \\ 
\textit{Thlaspi arvense} L. & 273 & 1208 & 88.64 ± 1.11 & 51.96 ± 1.21 & 66.76 ± 1.37 & 43.12 ± 0.64 \\ 
\textit{Triticum aestivum} L. & 58 & 1772 & 74.60 ± 2.90 & 67.08 ± 1.30 & 73.32 ± 2.04 & 32.72 ± 1.41 \\ 
\textit{Veronica persica} Poir. & 108 & 322 & 85.06 ± 2.94 & 68.04 ± 0.63 & 72.24 ± 1.59 & 40.10 ± 1.08 \\ 
\textit{Zea mays} L. & 65 & 1327 & 83.90 ± 1.27 & 77.02 ± 1.22 & 84.58 ± 0.72 & 51.60 ± 1.22 \\
\bottomrule
\end{tabular}

}
\end{table}

\begin{table}[H]
\caption{Performance of YOLOv10l across images and instances per species, with Precision, Recall, AP50, and AP50-95 provided in percentages, with standard deviations (+/- SD) indicated. Bold values represent the highest values.}
\label{tab:performance_YOLOv10l_class_wise_dataset1}
\resizebox{\textwidth}{!}{%
\begin{tabular}{c|c|c|c|c|c|c}
\toprule
\textbf{Class} & \textbf{Images} & \textbf{Instances} & \textbf{Precision} & \textbf{Recall} & \textbf{AP50} & \textbf{AP50-95} \\
\midrule
\textit{Abutilon theophrasti} Medik. & 48 & 1166 & 76.42 ± 1.66 & 75.62 ± 1.33 & 80.26 ± 0.61 & 45.90 ± 0.43 \\ 
\textit{Elymus repens} (L.) Gould & 53 & 309 & 70.98 ± 2.21 & 72.78 ± 2.41 & 74.92 ± 2.27 & 35.76 ± 1.31 \\ 
\textit{Alopecurus myosuroides} Huds. & 77 & 635 & \textbf{90.44 ± 1.32} & 82.58 ± 1.38 & 87.96 ± 1.02 & 49.22 ± 0.91 \\ 
\textit{Amaranthus retroflexus} L. & 94 & 418 & 85.98 ± 0.84 & 58.26 ± 1.75 & 68.96 ± 1.16 & 51.36 ± 0.85 \\ 
\textit{Avena fatua} L. & 77 & 1131 & 79.14 ± 2.25 & 65.74 ± 1.90 & 73.02 ± 1.16 & 40.14 ± 0.62 \\ 
\textit{Chenopodium album} L. & 101 & 329 & 81.66 ± 3.49 & 56.32 ± 1.07 & 67.44 ± 2.00 & 41.00 ± 1.41 \\ 
\textit{Geranium spp.} & 53 & 1347 & 79.36 ± 1.91 & 57.16 ± 2.23 & 66.20 ± 2.56 & 31.08 ± 0.75 \\ 
\textit{Helianthus annuus} L. & 56 & 1873 & 80.46 ± 1.57 & 73.90 ± 1.08 & 81.94 ± 0.71 & 60.00 ± 0.54 \\ 
\textit{Lamium purpureum} L. & 214 & 1069 & 90.20 ± 1.00 & 57.34 ± 0.60 & 67.72 ± 0.78 & 39.68 ± 0.74 \\ 
\textit{Fallopia convolvulus} (L.) Á.  Löve & 78 & 115 & 77.32 ± 5.85 & 57.66 ± 1.99 & 62.24 ± 2.55 & 44.34 ± 1.73 \\ 
\textit{Setaria spp.} & 134 & 275 & 77.86 ± 2.71 & 43.50 ± 1.71 & 49.28 ± 1.01 & 30.24 ± 1.27 \\ 
\textit{Solanum nigrum} L. & 57 & 331 & 80.88 ± 2.34 & \textbf{84.78 ± 0.97} & 89.18 ± 0.68 & \textbf{54.24 ± 1.51} \\ 
\textit{Thlaspi arvense} L. & 273 & 1208 & 89.62 ± 1.39 & 51.52 ± 0.79 & 64.08 ± 0.57 & 41.74 ± 0.51 \\ 
\textit{Triticum aestivum} L. & 58 & 1772 & 76.94 ± 1.35 & 66.26 ± 1.13 & 72.48 ± 1.16 & 32.20 ± 0.97 \\ 
\textit{Veronica persica} Poir. & 108 & 322 & 85.38 ± 2.65 & 67.34 ± 1.13 & 72.32 ± 1.56 & 39.02 ± 1.45 \\ 
\textit{Zea mays} L. & 65 & 1327 & 84.68 ± 0.88 & 77.60 ± 1.55 & \textbf{84.64 ± 0.65} & 51.84 ± 0.50 \\
\bottomrule
\end{tabular}

}
\end{table}

\begin{table}[H]
\caption{Performance of RT-DETR-l across images and instances per species, with Precision, Recall, AP50, and AP50-95 provided in percentages, with standard deviations (+/- SD) indicated. Bold values represent the highest values.}
\label{tab:performance_rt_detr_l_class_wise_dataset1}
\resizebox{\textwidth}{!}{%
\begin{tabular}{c|c|c|c|c|c|c}
\toprule
\textbf{Class} & \textbf{Images} & \textbf{Instances} & \textbf{Precision} & \textbf{Recall} & \textbf{AP50} & \textbf{AP50-95} \\
\midrule
\textit{Abutilon theophrasti} Medik. & 48 & 1166 & 76.98 ± 0.77 & 76.72 ± 0.71 & 78.08 ± 0.88 & 45.56 ± 0.98 \\ 
\textit{Elymus repens} (L.) Go & 53 & 309 & 77.64 ± 3.17 & 69.90 ± 2.54 & 75.06 ± 2.14 & 34.18 ± 1.57 \\ 
\textit{Alopecurus myosuroides} Huds. & 77 & 635 & \textbf{91.22 ± 0.90} & 81.88 ± 0.50 & \textbf{86.38 ± 0.59} & 49.16 ± 0.82 \\ 
\textit{Amaranthus retroflexus} L. & 94 & 418 & 88.00 ± 1.38 & 57.84 ± 1.80 & 68.92 ± 1.51 & 49.60 ± 0.97 \\ 
\textit{Avena fatua} L. & 77 & 1131 & 78.66 ± 1.59 & 66.90 ± 1.93 & 72.34 ± 1.11 & 39.88 ± 0.70 \\ 
\textit{Chenopodium album} L. & 101 & 329 & 82.66 ± 3.41 & 56.12 ± 2.37 & 67.18 ± 1.85 & 40.54 ± 0.90 \\ 
\textit{Geranium spp.} & 53 & 1347 & 81.12 ± 1.25 & 59.42 ± 1.57 & 66.46 ± 1.31 & 30.36 ± 0.81 \\ 
\textit{Helianthus annuus} L. & 56 & 1873 & 82.20 ± 1.51 & 74.96 ± 1.22 & 79.18 ± 1.91 & \textbf{57.06 ± 1.26} \\ 
\textit{Lamium purpureum} L. & 214 & 1069 & 89.76 ± 1.11 & 58.76 ± 1.53 & 64.60 ± 1.09 & 37.16 ± 0.90 \\ 
\textit{Fallopia convolvulus} (L.) Á.  Löve & 78 & 115 & 73.90 ± 7.05 & 62.18 ± 0.88 & 63.02 ± 0.93 & 48.22 ± 1.34 \\ 
\textit{Setaria spp.} & 134 & 275 & 79.30 ± 5.37 & 43.90 ± 0.86 & 47.20 ± 0.29 & 28.30 ± 1.10 \\ 
\textit{Solanum nigrum} L. & 57 & 331 & 78.14 ± 3.86 & \textbf{83.82 ± 2.28} & 86.16 ± 0.75 & 52.42 ± 0.97 \\ 
\textit{Thlaspi arvense} L. & 273 & 1208 & 88.86 ± 1.26 & 49.72 ± 1.18 & 59.32 ± 0.61 & 38.14 ± 0.57 \\ 
\textit{Triticum aestivum} L. & 58 & 1772 & 76.94 ± 1.45 & 68.84 ± 0.90 & 69.84 ± 2.13 & 30.28 ± 1.03 \\ 
\textit{Veronica persica} Poir. & 108 & 322 & 88.88 ± 2.31 & 68.10 ± 1.67 & 71.62 ± 0.83 & 39.80 ± 0.61 \\ 
\textit{Zea mays} L. & 65 & 1327 & 84.54 ± 1.50 & 77.38 ± 0.71 & 81.98 ± 0.44 & 49.62 ± 0.59 \\
\bottomrule
\end{tabular}

}
\end{table}

\subsubsection{Dataset 2}

In this section, a detailed analysis of the class-wise performance of the YOLOv8l, YOLOv9c, YOLOv10l, and RT-DETR-l models in dataset 2 is presented. For this dataset, weed species were grouped into two broad categories: Monocot and Dicot, while the crop classes (\textit{H. annuus}, \textit{T. aestivum} and \textit{Z. mays}) remained unchanged from dataset 1. The corresponding results for each model can be found in Tables \ref{tab:performance_yolov8_dataset2}, \ref{tab:performance_yolov9_dataset2}, \ref{tab:performance_yolov10_dataset2} and \ref{tab:performance_rtdetrl_dataset2}. The results for the remaining models are available in the Supplementary Material.

The YOLOv8l model displayed solid performance across the grouped weed classes and the crops, with a average precision of 80.54\% and an average recall of 71.14\%. The model performed particularly well on the Dicot class, achieving a high precision of 85.28\% and an AP50 of 76.90\%. Among the crops, the dicot \textit{H. annuus} was detected with a precision of 79.74\% and a AP50-95 of 60.96\%, the highest among the crops. However, detection of the monocot crop \textit{T. aestivum} proved challenging, with a lower AP50-95 of 32.02\%, indicating difficulties in handling this crop at more stringent IoU thresholds.

The YOLOv9c model demonstrated comparable performance to YOLOv8l, with a slightly lower average precision of 80.04\% and an average recall of 71.96\%. It showed strong results for the Dicot class with a precision of 84.76\% and an AP50 of 77.32\%. \textit{H. annuus} was detected with similar effectiveness, achieving a precision of 78.72\% and the highest AP50-95 among crops at 61.40\%. While YOLOv9c managed a better performance on \textit{T. aestivum} with an AP50-95 of 32.58\%, Monocot weeds showed relatively lower performance, with an AP50-95 of 42.32\%, suggesting some challenges in detecting this group under stricter IoU criteria.

The YOLOv10l model performed consistently well, with an average precision of 80.56\% and an average recall of 70.58\%. It particularly excelled in detecting the Dicot class with a precision of 85.10\% and a AP50 of 77.68\%. For \textit{H. annuus}, YOLOv10l achieved a strong AP50-95 of 60.42\%. The performance on \textit{Z. mays} was also commendable, with a precision of 82.72\% and a AP50-95 of 52.04\%, indicating the model's reliability in detecting this crop under more challenging conditions. However, as with other models,\textit{ T. aestivum} remained more difficult to detect accurately, with a AP50-95 of 31.74\%.

The RT-DETR-l model exhibited the highest average precision among the four models, reaching 81.46\%, with an average recall of 71.78\%. The model performed exceptionally well on \textit{H. annuus}, achieving a precision of 80.94\% and a AP50-95 of 56.56\%, the highest in all models. The detection of \textit{Z. mays} also showed strong results, with a precision of 85.06\% and AP50-95 of 50.20\%. However, the performance in \textit{T. aestivum} was again the lowest, with a AP50-95 of 29.42\%, indicating that even the most precise model struggled with this particular crop.

\begin{table}[H]
\caption{Performance of YOLOv8l across images and instances per species, with Precision, Recall, AP50, and AP50-95 provided in percentages, with standard deviations (+/- SD) indicated. Bold values represent the highest values.}
\label{tab:performance_yolov8_dataset2}
\resizebox{\textwidth}{!}{%
\begin{tabular}{c|c|c|c|c|c|c}
\toprule
\textbf{Class} & \textbf{Images} & \textbf{Instances} & \textbf{Precision} & \textbf{Recall} & \textbf{AP50} & \textbf{AP50-95} \\
\midrule
Dicot & 590 & 6305 & \textbf{85.28 ± 0.70} & 65.54 ± 0.91 & 76.90 ± 0.67 & 44.62 ± 0.32 \\ 
\textit{Helianthus annuus} L. & 56 & 1873 & 79.74 ± 0.70 & 76.06 ± 1.29 & 83.52 ± 1.21 & \textbf{60.96 ± 1.04} \\ 
Monocot & 312 & 2350 & 80.94 ± 0.96 & 70.32 ± 1.28 & 77.04 ± 1.02 & 42.22 ± 0.85 \\ 
\textit{Triticum aestivum} L. & 58 & 1772 & 73.80 ± 1.55 & 67.28 ± 0.59 & 71.64 ± 1.09 & 32.02 ± 0.40 \\ 
\textit{Zea mays} L. & 65 & 1327 & 82.90 ± 1.16 & \textbf{76.48 ± 1.22} & \textbf{84.36 ± 0.76} & 51.44 ± 1.14 \\
\bottomrule
\end{tabular}

}
\end{table}

\begin{table}[H]
\caption{Performance of YOLOv9c across images and instances per species, with Precision, Recall, AP50, and AP50-95 provided in percentages, with standard deviations (+/- SD) indicated. Bold values represent the highest values.}
\label{tab:performance_yolov9_dataset2}
\resizebox{\textwidth}{!}{%
\begin{tabular}{c|c|c|c|c|c|c}
\toprule
\textbf{Class} & \textbf{Images} & \textbf{Instances} & \textbf{Precision} & \textbf{Recall} & \textbf{AP50} & \textbf{AP50-95} \\
\midrule
Dicot & 590 & 6305 & \textbf{84.76 ± 0.59} & 66.40 ± 0.88 & 77.32 ± 0.53 & 44.66 ± 0.42 \\ 
\textit{Helianthus annuus} L. & 56 & 1873 & 78.72 ± 1.93 & 76.04 ± 1.22 & 84.26 ± 1.21 & \textbf{61.40 ± 0.99} \\ 
Monocot & 312 & 2350 & 80.64 ± 1.11 & 70.18 ± 0.26 & 77.04 ± 0.64 & 42.32 ± 0.43 \\ 
\textit{Triticum aestivum} L. & 58 & 1772 & 73.20 ± 2.03 & 69.00 ± 1.47 & 72.62 ± 1.96 & 32.58 ± 1.04 \\ 
\textit{Zea mays} L. & 65 & 1327 & 82.98 ± 1.79 & \textbf{78.22 ± 0.98} & \textbf{84.50 ± 0.86} & 51.58 ± 0.83 \\
\bottomrule
\end{tabular}

}
\end{table}

\begin{table}[H]
\caption{Performance of YOLOv10l across images and instances per species, with Precision, Recall, AP50, and AP50-95 provided in percentages, with standard deviations (+/- SD) indicated. Bold values represent the highest values.}
\label{tab:performance_yolov10_dataset2}
\resizebox{\textwidth}{!}{%
\begin{tabular}{c|c|c|c|c|c|c}
\toprule
\textbf{Class} & \textbf{Images} & \textbf{Instances} & \textbf{Precision} & \textbf{Recall} & \textbf{mAP50} & \textbf{mAP50-95} \\
\midrule
Dicot & 590 & 6305 & \textbf{85.10 ± 0.74} & 65.60 ± 1.27 & 77.68 ± 0.99 & 45.24 ± 0.68 \\ 
\textit{Helianthus annuus} L. & 56 & 1873 & 79.04 ± 1.75 & 74.18 ± 0.89 & 82.22 ± 1.00 & \textbf{60.42 ± 1.08} \\ 
Monocot & 312 & 2350 & 80.84 ± 1.53 & 69.82 ± 0.79 & 77.08 ± 0.55 & 42.50 ± 0.44 \\ 
\textit{Triticum aestivum} L. & 58 & 1772 & 75.02 ± 1.12 & 66.62 ± 1.05 & 72.16 ± 0.29 & 31.74 ± 0.30 \\ 
\textit{Zea mays} L. & 65 & 1327 & 82.72 ± 1.00 & \textbf{76.78 ± 0.57} & \textbf{84.38 ± 0.72} & 52.04 ± 0.53 \\
\bottomrule
\end{tabular}

}
\end{table}

\begin{table}[H]
\caption{Performance of RT-DETR-l across images and instances per species, with Precision, Recall, AP50, and AP50-95 provided in percentages, with standard deviations (+/- SD) indicated. Bold values represent the highest values.}
\label{tab:performance_rtdetrl_dataset2}
\resizebox{\textwidth}{!}{%
\begin{tabular}{c|c|c|c|c|c|c}
\toprule
\textbf{Class} & \textbf{Images} & \textbf{Instances} & \textbf{Precision} & \textbf{Recall} & \textbf{AP50} & \textbf{AP50-95} \\
\midrule
Dicot & 590 & 6305 & 84.14 ± 0.95 & 65.88 ± 0.50 & 75.56 ± 0.64 & 43.04 ± 0.05 \\ 
\textit{Helianthus annuus} L & 56 & 1873 & 80.94 ± 0.40 & 75.02 ± 0.91 & 78.52 ± 0.87 & \textbf{56.56 ± 0.80} \\ 
Monocot & 312 & 2350 & 81.42 ± 1.85 & 70.72 ± 0.93 & 75.22 ± 1.05 & 40.66 ± 0.84 \\ 
\textit{Triticum aestivum} L. & 58 & 1772 & 75.74 ± 1.34 & 68.94 ± 1.29 & 68.80 ± 1.98 & 29.42 ± 1.07 \\ 
\textit{Zea mays} L. & 65 & 1327 & \textbf{85.06 ± 1.36} & \textbf{78.30 ± 1.21} & \textbf{82.58 ± 1.03} & 50.20 ± 0.50 \\
\bottomrule
\end{tabular}

}
\end{table}

\subsubsection{Inference times}

The results, summarized in Table \ref{table:inference_time1} and Table \ref{table:inference_time2}, reflect the performance in the two datasets.

Table \ref{table:inference_time1} presents the inference times recorded for each model when evaluated on the dataset 1. Among the YOLOv8 models, YOLOv8x, being the most complex, exhibited the longest inference times, with 51.06 ± 1.37 ms on the NVIDIA GeForce RTX 3090 GPU and 908.74 ± 0.72 ms on the Intel Core i9-14900K CPU. Conversely, YOLOv8n, the smallest and least complex model in the YOLOv8 family, demonstrated much faster inference times, beginning with 8.46 ± 0.32 ms on the RTX 3090 GPU and 38.80 ± 0.29 ms on the Intel CPU. The YOLOv9 models followed a similar trend, with the YOLOv9e model, as a more complex variant, recording 56.76 ± 0.73 ms on the RTX 3090 GPU and 921.20 ± 1.81 ms on the Intel CPU, whereas the YOLOv9t model, a simpler variant, registered 15.06 ± 0.21 ms on the RTX 3090 GPU and 43.58 ± 0.21 ms on the Intel CPU.  When considering the YOLOv10 models, a notable improvement in inference times is observed compared to earlier versions. The YOLOv10x model, for example, achieved 42.78 ± 1.69 ms on the RTX 3090 GPU and 957.62 ± 0.87 ms on the Intel Core i9 CPU.

Table \ref{table:inference_time2} presents the inference times for the same models, this time evaluated on dataset 2. The results show consistency in the trend observed with dataset 1, but with some variations. The YOLOv8 models, for instance, showed slightly higher inference times on dataset 2. YOLOv8x recorded 53.04 ± 0.91 ms on the NVIDIA GeForce RTX 3090 GPU and 909.66 ± 0.92 ms on the Intel Core i9-14900K CPU. Similarly, YOLOv8n, the smallest model, had inference times of 8.96 ± 0.24 ms on the RTX 3090 GPU and 38.62 ± 0.37 ms on the Intel CPU, reflecting a slight increase compared to dataset 1. The YOLOv9 models exhibited a similar pattern. YOLOv9e showed a marginal increase in inference time on dataset 2, with 57.18 ± 0.77 ms on the RTX 3090 GPU and 920.80 ± 1.35 ms on the Intel Core i9 CPU. YOLOv9t, on the other hand, remained relatively consistent, with inference times of 15.04 ± 0.27 ms on the RTX 3090 GPU and 43.26 ± 0.27 ms on the Intel CPU. In the YOLOv10 family, YOLOv10x again demonstrated improved performance, achieving 44.66 ± 2.31 ms on the RTX 3090 GPU and 956.92 ± 0.62 ms on the Intel Core i9 CPU. The RT-DETR-l model also maintained its efficiency, recording 25.94 ± 0.57 ms on the RTX 3090 GPU, further validating its suitability for real-time applications.

\begin{table}[H]
\caption{Inference times (in milliseconds) for various models across different hardware setups for the dataset 1. The results are presented as mean, with standard deviations (+/- SD) indicated. Bold values represent the highest values.}
\label{table:inference_time1}
\resizebox{\textwidth}{!}{%
\begin{tabular}{lcccc}
\toprule
\textbf{Model} & \makecell{\textbf{NVIDIA GeForce} \\ \textbf{RTX 3090 GPU}} & \makecell{\textbf{NVIDIA GeForce} \\ \textbf{RTX 4090 GPU}} & \makecell{\textbf{Intel Core i9-14900K} \\ \textbf{32-Core CPU}} & \makecell{\textbf{AMD Ryzen 9 5950X} \\ \textbf{16-Core CPU}} \\
\midrule
YOLOv8x & 51.06 ± 1.37 ms & 24.12 ± 0.71 ms & 908.74 ± 0.72 ms & 1845.84 ± 8.72 ms \\
YOLOv8l & 33.98 ± 1.35 ms & 17.50 ± 0.24 ms & 585.44 ± 0.74 ms & 1166.34 ± 11.39 ms \\
YOLOv8m & 20.02 ± 0.53 ms & 10.72 ± 0.17 ms & 288.66 ± 0.77 ms & 671.88 ± 3.35 ms \\
YOLOv8s & 10.64 ± 0.37 ms & 7.78 ± 0.29 ms & 113.92 ± 0.47 ms & 273.12 ± 4.42 ms \\
YOLOv8n & \textbf{8.46 ± 0.32 ms} & \textbf{7.64 ± 0.30 ms} & \textbf{38.80 ± 0.29 ms} & \textbf{120.44 ± 2.96 ms} \\
YOLOv9e & 56.76 ± 0.73 ms & 32.30 ± 0.35 ms & 921.20 ± 1.81 ms & 2949.46 ± 7.75 ms \\
YOLOv9c & 27.08 ± 0.43 ms & 15.20 ± 0.36 ms & 390.40 ± 0.88 ms & 1020.52 ± 4.33 ms \\
YOLOv9m & 22.94 ± 0.52 ms & 13.18 ± 0.07 ms & 311.34 ± 0.99 ms & 750.38 ± 5.29 ms \\
YOLOv9s & 15.42 ± 0.19 ms & 12.04 ± 0.34 ms & 111.16 ± 0.63 ms & 343.48 ± 3.53 ms \\
YOLOv9t & 15.06 ± 0.21 ms & 11.26 ± 0.45 ms & 43.58 ± 0.21 ms & 160.26 ± 0.86 ms \\
YOLOv10x & 42.78 ± 1.69 ms & 21.96 ± 0.69 ms & 957.62 ± 0.87 ms & 1586.34 ± 9.50 ms \\
YOLOv10l & 31.92 ± 0.81 ms & 17.82 ± 0.56 ms & 712.40 ± 0.70 ms & 1036.64 ± 2.62 ms \\
YOLOv10m & 20.02 ± 0.53 ms & 11.02 ± 0.44 ms & 315.54 ± 3.72 ms & 657.38 ± 3.68 ms \\
YOLOv10s & 11.46 ± 0.64 ms & 8.58 ± 0.40 ms & 137.78 ± 2.29 ms & 286.00 ± 0.21 ms \\
YOLOv10n & 9.92 ± 0.40 ms & 8.24 ± 0.37 ms & 61.26 ± 1.16 ms & 127.90 ± 0.11 ms \\
RT-DETR-x & 43.84 ± 0.69 ms & 24.82 ± 0.63 ms & 879.38 ± 1.92 ms & 1327.86 ± 16.64 ms \\
RT-DETR-l & 25.78 ± 0.23 ms & 15.76 ± 0.69 ms & 452.32 ± 0.99 ms & 767.64 ± 4.85 ms \\
\bottomrule
\end{tabular}
}
\end{table}

\begin{table}[H]
\caption{Inference times (in milliseconds) for various models across different hardware setups for the dataset 2. The results are presented as mean, with standard deviations (+/- SD) indicated. Bold values represent the highest values.}
\label{table:inference_time2}
\resizebox{\textwidth}{!}{%
\begin{tabular}{lcccc}
\toprule
\textbf{Model} & \makecell{\textbf{NVIDIA GeForce} \\ \textbf{RTX 3090 GPU}} & \makecell{\textbf{NVIDIA GeForce} \\ \textbf{RTX 4090 GPU}} & \makecell{\textbf{Intel Core i9-14900K} \\ \textbf{32-Core CPU}} & \makecell{\textbf{AMD Ryzen 9 5950X} \\ \textbf{16-Core CPU}} \\
\midrule
YOLOv8x & 53.04 ± 0.91 ms & 25.54 ± 0.81 ms & 909.66 ± 0.92 ms & 1473.34 ± 13.79 ms \\
YOLOv8l & 34.10 ± 0.41 ms & 17.96 ± 0.80 ms & 586.28 ± 0.64 ms & 935.32 ± 2.78 ms \\
YOLOv8m & 19.08 ± 0.13 ms & 10.92 ± 0.23 ms & 288.88 ± 0.47 ms & 568.46 ± 1.95 ms \\
YOLOv8s & 10.26 ± 0.42 ms & 8.18 ± 0.44 ms & 113.52 ± 0.51 ms & 265.26 ± 2.74 ms \\
YOLOv8n & \textbf{8.96 ± 0.24 ms} & \textbf{7.58 ± 0.42 ms} & \textbf{38.62 ± 0.37 ms} & \textbf{124.38 ± 0.48 ms} \\
YOLOv9e & 57.18 ± 0.77 ms & 32.30 ± 0.22 ms & 920.80 ± 1.35 ms & 2699.70 ± 57.86 ms \\
YOLOv9c & 27.60 ± 0.33 ms & 15.04 ± 0.41 ms & 390.72 ± 0.50 ms & 904.24 ± 6.24 ms \\
YOLOv9m & 23.24 ± 0.49 ms & 13.12 ± 0.17 ms & 311.88 ± 0.62 ms & 653.78 ± 1.43 ms \\
YOLOv9s & 15.42 ± 0.28 ms & 11.68 ± 0.29 ms & 111.10 ± 0.46 ms & 312.82 ± 0.60 ms \\
YOLOv9t & 15.04 ± 0.27 ms & 11.40 ± 0.14 ms & 43.26 ± 0.27 ms & 155.50 ± 1.53 ms \\
YOLOv10x & 44.66 ± 2.31 ms & 22.30 ± 0.63 ms & 956.92 ± 0.62 ms & 1385.18 ± 3.12 ms \\
YOLOv10l & 32.32 ± 0.90 ms & 17.14 ± 0.34 ms & 711.44 ± 0.73 ms & 912.86 ± 12.10 ms \\
YOLOv10m & 20.18 ± 0.64 ms & 10.94 ± 0.40 ms & 321.28 ± 10.36 ms & 579.52 ± 3.24 ms \\
YOLOv10s & 11.58 ± 0.31 ms & 8.22 ± 0.26 ms & 133 ± 1.46 ms & 274.52 ± 0.90 ms \\
YOLOv10n & 9.80 ± 0.19 ms & 7.90 ± 0.11 ms & 59.88 ± 2.47 ms & 128.40 ± 1.96 ms \\
RT-DETR-x & 44.92 ± 0.96 ms & 25.30 ± 1.03 ms & 878.52 ± 1.03 ms & 1299.52 ± 2.06 ms \\
RT-DETR-l & 25.94 ± 0.57 ms & 15.16 ± 0.28 ms & 450.84 ± 0.69 ms & 755.44 ± 3.02 ms \\
\bottomrule
\end{tabular}
}
\end{table}

\section{Discussion}
The present study assesses the efficacy of deep neural networks (DNNs) for multi-class weed and crop detection in realistic agricultural scenarios. This is achieved by utilizing images from an experimental field, where only the species composition was controlled in the training and validation sets.
A comparison of the two datasets used in this study reveals that dataset 2 generally produces superior results, particularly in recall and mAP50 and mAP50-95 scores, indicating more effective identification and classification of weeds. This improvement may result from a simplified system for distinguishing between monocot and dicot weeds, which improves the accuracy of classification. However, the reduced precision in dataset 2 is linked to an increase in false positives, as the consolidation of species increases the number of images per species but may lead to incorrect classifications. Nevertheless, the recall values suggest higher recognition due to reduced differentiation between species groups, which facilitates recognition and increases the number of identified objects. Consequently, mAP scores reflect this improvement, as improved recall compensates for the reduction in the precision score.
In contrast to dataset 1, which includes more classes, dataset 2 consistently demonstrates higher precision for dicots, suggesting that fewer classes may reduce confusion among similar categories. Grouping dicot and monocot weeds in dataset 2 further enhanced performance, potentially by reducing complexity and emphasizing broader morphological differences. Notably, species like the crop \textit{Z. Maize} in dataset 2 show effective detection, while in dataset 1 \textit{A. myosuroides} achieved high precision and accuracy, indicating appropriate representation.
The YOLOv9 models, especially YOLOv9s and YOLOv9e, exhibit superior performance across both datasets, supporting their practical application in weed detection, confirmed by other studies \cite{Saltik_2024_ECCV}. The RT-DETR-l model demonstrated the highest overall precision, rendering it a valuable tool in scenarios, where minimizing false positive is essential, particularly in the context of herbicide reduction, which not only leads to cost savings but also offers environmental benefits, thereby advancing the objectives set by the EU Green Deal. 

The variability of species plays a role in the accuracy of detection, with some species being more readily recognised than others. This is likely due to a number of factors, including visual features, the representation of species in training data, or the morphological similarities between weeds and between weeds and crop species. In dataset 1, species like the monocot \textit{Setaria spp.} and the dicot \textit{C. album} presented challenges, as indicated by lower precision and mAP50-95 scores. By contrast, species like the dicot crop \textit{H. annuus} consistently achieved high precision and mAP50-95 scores across both datasets. However the monocot crop \textit{T. aestivum} presented certain difficulties, indicating the necessity for specialized training or optimized data representation. While \textit{A. myosuroides} was precisely identified in dataset 1, the broader Monocot category in dataset 2 showed reduced performance at stringent IoU thresholds, likely due to greater variability within these larger weed groups. In terms of practical implementation, an accuracy of approximately 80\% may be deemed adequate for some operations.  However, when considering spot spraying or selective hoeing, it is questionable whether this level of accuracy is sufficient. An accuracy of 80\% indicates that 20\% of the species, whether crops or weeds, will remain unidentified. In the context of spraying, the issue is less severe since herbicides are selective and would not cause harm to the crop if applied. In contrast, in the case of hoeing, failure to recognize, for example, 20\% of the crop plant would result in a 20\% loss of yield, because the crop plants would be removed by the blades of the hoe, which would be an unacceptable outcome for a farmer.

In general, considering the dataset \cite{dyrmann2018estimation} determined that insufficient size of dataset results in a reduction in classification accuracy. The present study utilized 5611 images, averaging 350 images per species, maintaining a constant number of images while varying the number of plants per image, influencing bounding box counts and instances. The mAP50-95 score was not influenced by images per class but by instances per species, as demonstrated by \textit{C. album}'s higher instance count compared to \textit{F. convolvulus}.
Limited examples challenge the DNN's ability to learn plant characteristics, resulting in higher rate of missclassifications. This was confirmed in dataset 2, where \textit{T. aestivum} had the fewest images and instances, achieving lower mAP50-95 scores. The inclusion of additional instances within a class is more beneficial for the detection accuracy than the mere increase in the number of images.
In this study, relatively large image dimensions were employed, which facilitated the reliable identification of smaller weeds. This approach was based on the findings of previous studies indicating that larger images enhance mAP50-95 scores \cite{Saltik_2024_ECCV}. However, results may vary significantly with reduced image size, warranting further investigation.\\
The distribution of different plant species in a given field represents a further crucial factor to be taken into account when a dataset is being compiled. As outlined by \cite{tian2019apple}, distinguishing thin-leaved grass weeds is challenging, but monocots like the crop \textit{Z. mays} can be more readily distinguished from other monocots. As dataset 2 showed, the Monocot group exhibits higher mAP50-95 scores than the monocot crop \textit{T. aestivum}, correlating with a greater number of instances in the images. In this regard \cite{yang2022detection} demonstrated that recognition accuracy does not vary solely between monocots and dicots; dicot weeds with similar morphology can often be misclassified, particularly in dense vegetation scenarios \cite{yang2022detection}.

The representation of species in images affects both recognition accuracy and DNN processing speed. Evaluating YOLOv8, YOLOv9, YOLOv10, and RT-DETR models show a balance between speed, complexity, and accuracy. YOLOv8n, YOLOv9t, and YOLOv10n achieved inference times below 100 ms, with YOLOv8n and YOLOv9t under 50 ms, making them highly accurate and ideal for speed-critical, resource-limited settings. However, inadequate computing power can hinder the correct identification of weed and crop \cite{Coleman}.
Comparative analysis of YOLOv9s and YOLOv8n revealed that YOLOv9s required approximately twice the processing time, regardless of the dataset. Across all models, the GPUs outperformed the CPUs, with inference times for CPU up to 15 times longer. Advancements in YOLOv10 showed faster inference times than previous versions, especially on high-performance GPUs like the NVIDIA GeForce RTX 4090. RT-DETR models exhibited competitive performance, underscoring their suitability for real-time applications requiring low latency. These results highlight the importance of selecting appropriate models based on operational requirements. The combination of high speed and accuracy in smaller YOLO models allows deployment in scenarios without GPU resources. Utilizing lightweight CPUs or Neural Processing Units (NPUs) may also improve performance, as different hardware significantly influences accuracy and computational time. Nevertheless, according to specific needs, less sophisticated hardware may suffice to achieve desired results \cite{herterich4862267accelerating}.\\
The presented DNNs can be used for species detection, significantly supporting the reduction of herbicide usage. Practical applications include sensor-controlled mechanical weeding systems, which can achieve results comparable to those of chemical weeding systems \cite{kunz2018camera}. Mechanical weeding systems are for example the Kult-iVision hoe, developed in collaboration with the University of Hohenheim \cite{gerhards2024comparison}, where hoe blades retract upon detecting crops in the intra-row area. Furthermore, these models could be integrated into real-time spot spraying solutions, like the Smart Sprayer developed by Robert Bosch GmbH \cite{spaeth2024smart}. An additional option for utilizing these models is to employ a UAV to scan the field prior to the creation of a weed map based on DNNs. This approach has been demonstrated to achieve significant cost savings in other studies, with a reduction of herbicide usage of up to 39.2 \%  for patch spraying \cite{castaldi2017assessing} and up to 47\% for spot spraying without compromising weed control efficacy or reducing yield \cite{allmendinger2024agronomic}. These systems offer farmers and contractors the opportunity to optimize agricultural processes, thereby achieving cost savings for example in maize of up to 42 € per hectare \cite{timmermann2003economic} and sustainability through the reduction of herbicide usage. This study demonstrates that species-specific data collection, plant identification, and real-time response mechanisms are feasible with the currently available models and hardware. The next step involves field integration on tractors, whereby the system performance will be assessed under real-world conditions including variable lightning conditions, diverse crop densities and soil conditions, as well as different crops and weeds.

\section{Conclusions}

The results of the study indicate that the latest models, aligned with current technological standards, are capable of simulating a real-time application. The use of RT-DETR-l is recommended for scenarios that require a reduced number of false positives. As expected, smaller YOLO variants, such as YOLOv9t, show a shorter inference time. In summary, dataset 2 improves model efficacy across most metrics, highlighting its suitability for practical use. When comparing class-wise performance between datasets, key differences emerge. The current state of research does not yet allow for reliable application in all agricultural settings or consistent accuracy across all crops and weed species, making system optimization essential to meet EU Green Deal goals. Further research is required, specifically on weed composition within images relatively to field conditions, which may affect DNN performance. Testing in diverse field conditions, including images from different cameras and angles, is needed. 
Future deployment of DNNs could be optimized for multiple locations by using synthetic training images, allowing varied backgrounds and soil conditions without extensive on-site data collection. This approach is exemplified in \cite{modak2024synthesizing}, which combines synthetic and field-based images. Initial tests already demonstrated the capability to reconstruct the agricultural settings.

\section {Conflict of Interest}
The authors declare that they have no conflict of interest.

\section {Data availability}

Data can be provided on request to the corresponding author.

\section{Funding}

This research was funded by EU-EIT FOOD as projects \#20140 and \#20140-21 “DACWEED”: Detection and Actuation system for WEED management. EIT FOOD is the innovation community on Food of the European Institute of Innovation and Technology (EIT), an EU body under Horizon 2020, the EU Framework Program for Research and Innovation.

The authors thank all those who helped in the realization of the dataset. We would like to thank all the technicians at the research station Heidfeldhof for the field preparations made and the technicians Jan Roggenbuck, Alexandra Heyn, Cathrin Brechlin and Dr. Markus Sökefeld from the Department of Weed Science and Philipp Reichel for their aid during the field work.

\section{Declaration of generative AI and AI-assisted technologies in the writing process}

During the preparation of this work the authors used ChatGPT in order to improve language and readability. After using this tool, the authors reviewed and edited the content as needed and take full responsibility for the content of the publication.

\bibliographystyle{splncs04}
\bibliography{ref}

\section{Supplementary Material}
\subsection{Supplementary A: Comparison of captured and annotated ground truth images for dataset 1 and dataset 2.}

\begin{figure}[H]
    \centering
    \begin{subfigure}{0.3\textwidth}
        \centering
        \includegraphics[width=\textwidth]{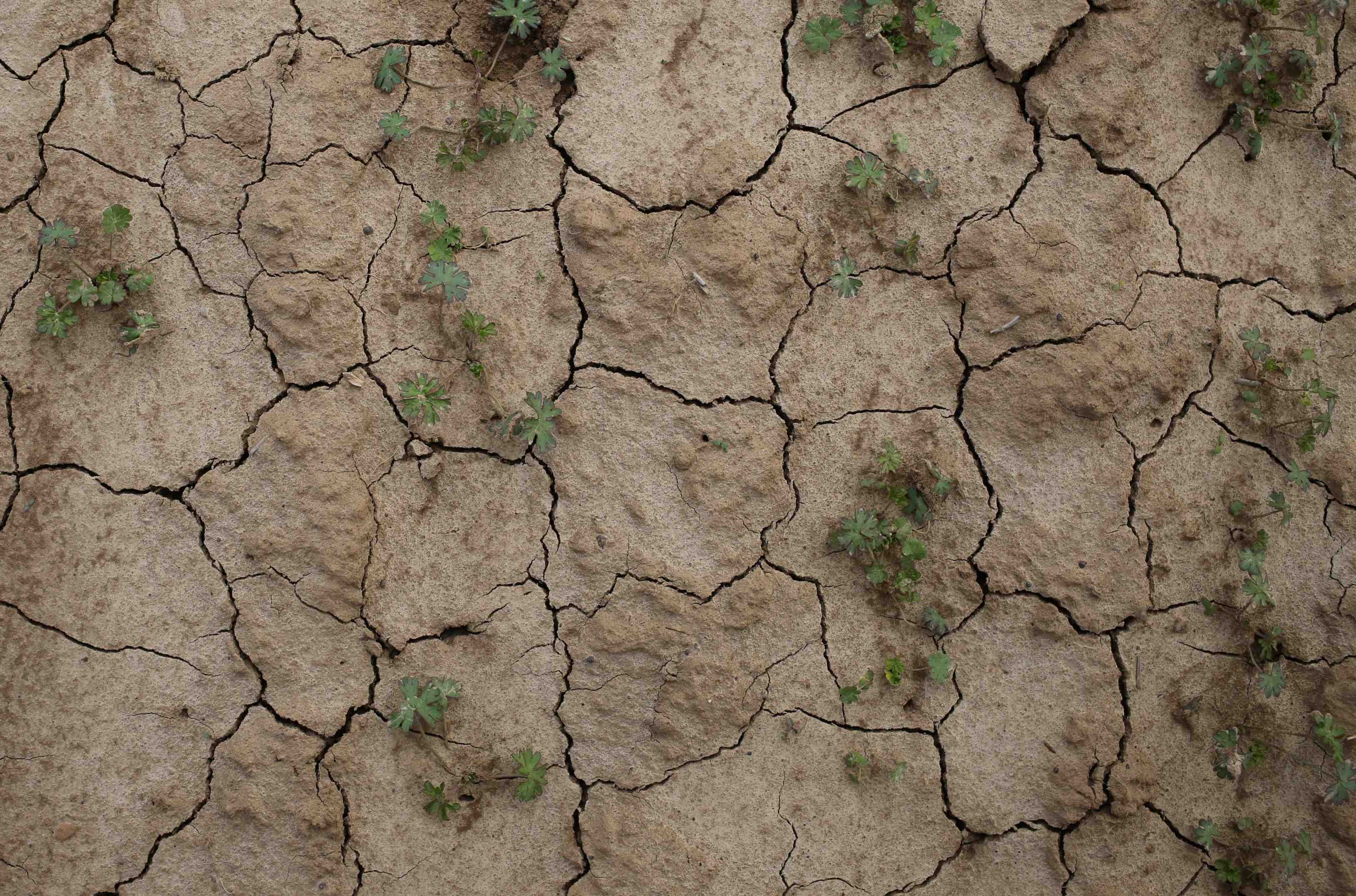}
        \caption{\textit{Geranium spp.}}
    \end{subfigure}
    \begin{subfigure}{0.3\textwidth}
        \centering
        \includegraphics[width=\textwidth]{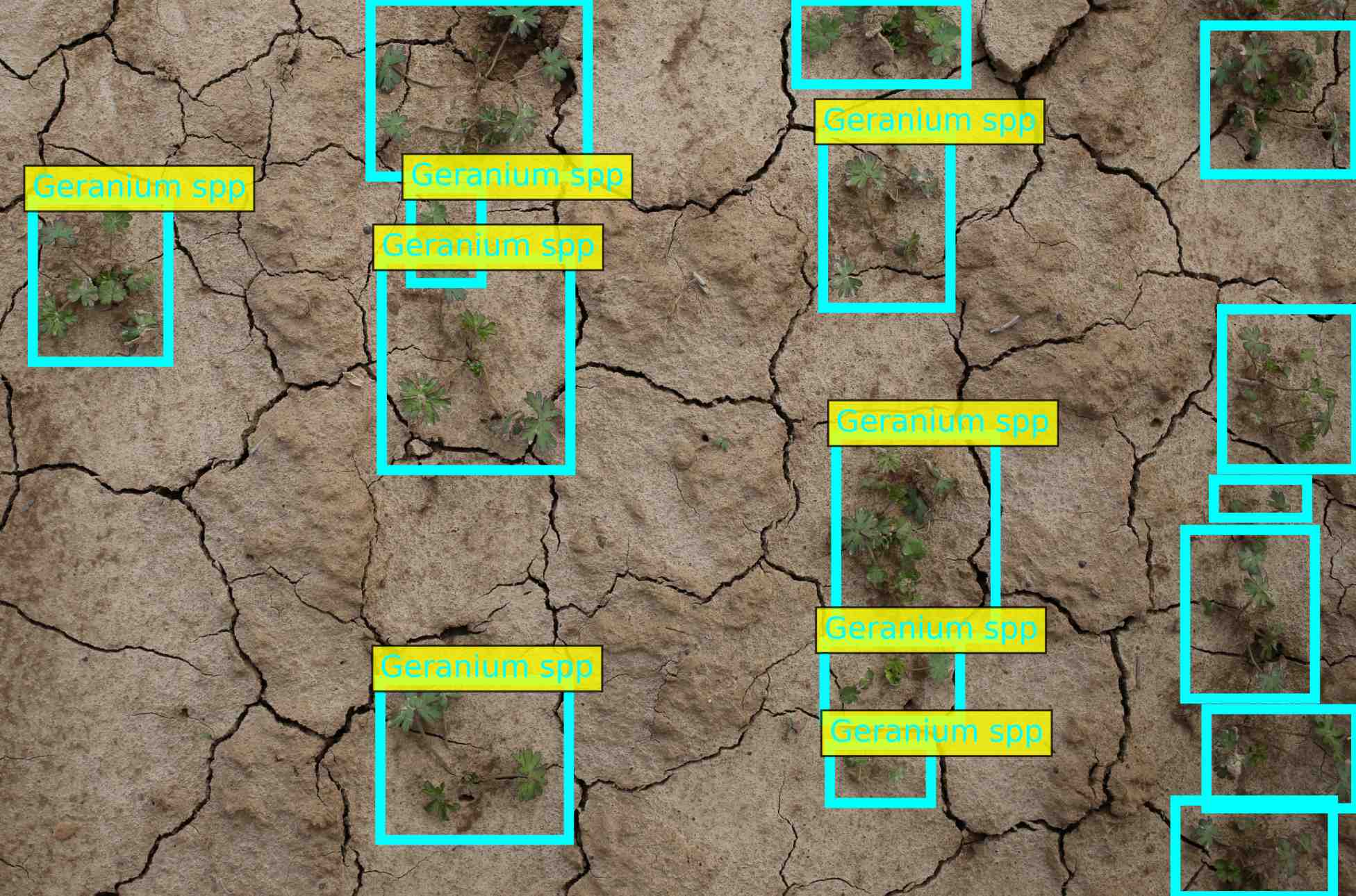}
        \caption{Dataset 1}
    \end{subfigure}
    \begin{subfigure}{0.3\textwidth}
        \centering
        \includegraphics[width=\textwidth]{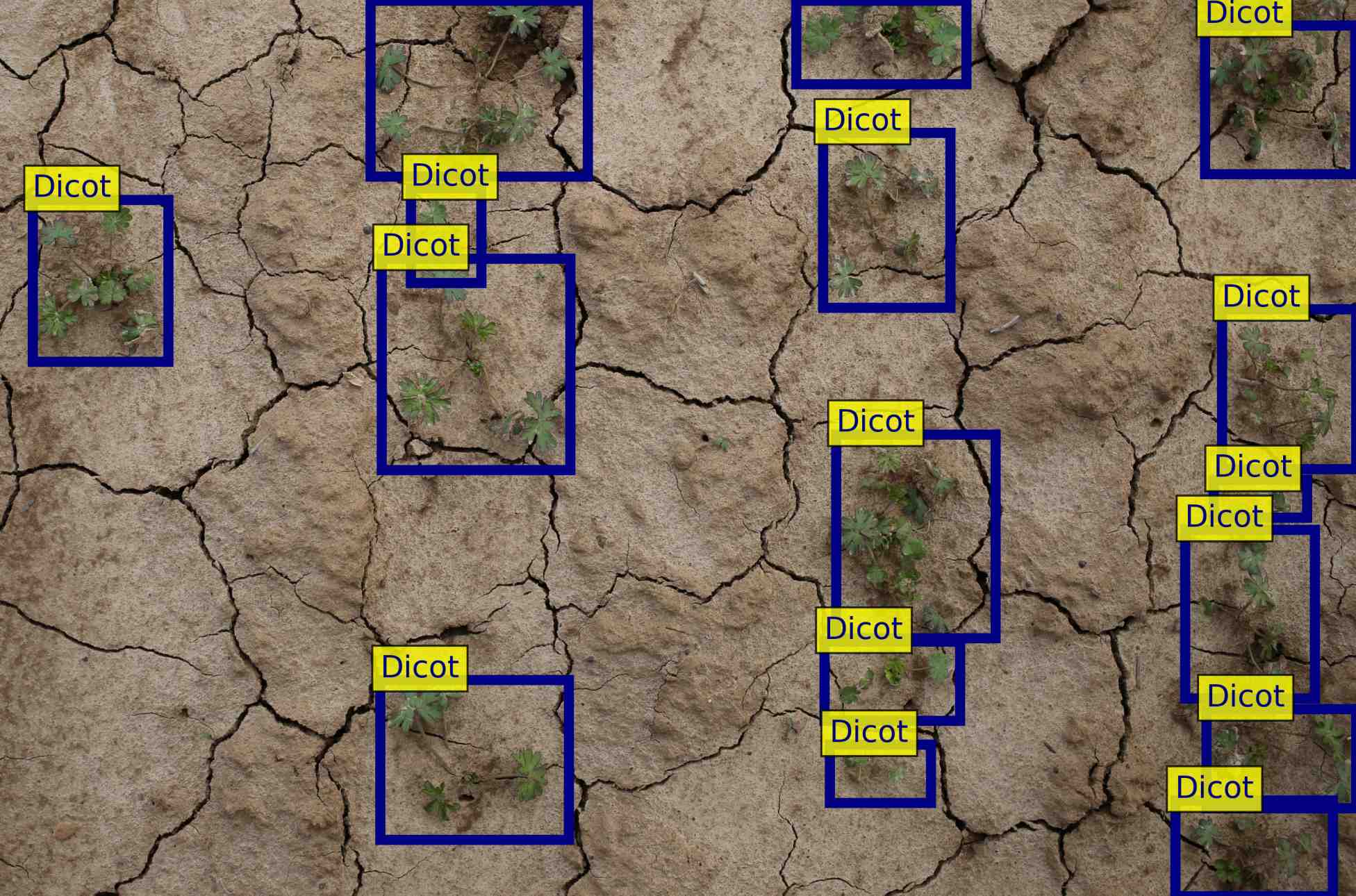}
        \caption{Dataset 2}
    \end{subfigure}

  
    \begin{subfigure}{0.3\textwidth}
        \centering
        \includegraphics[width=\textwidth]{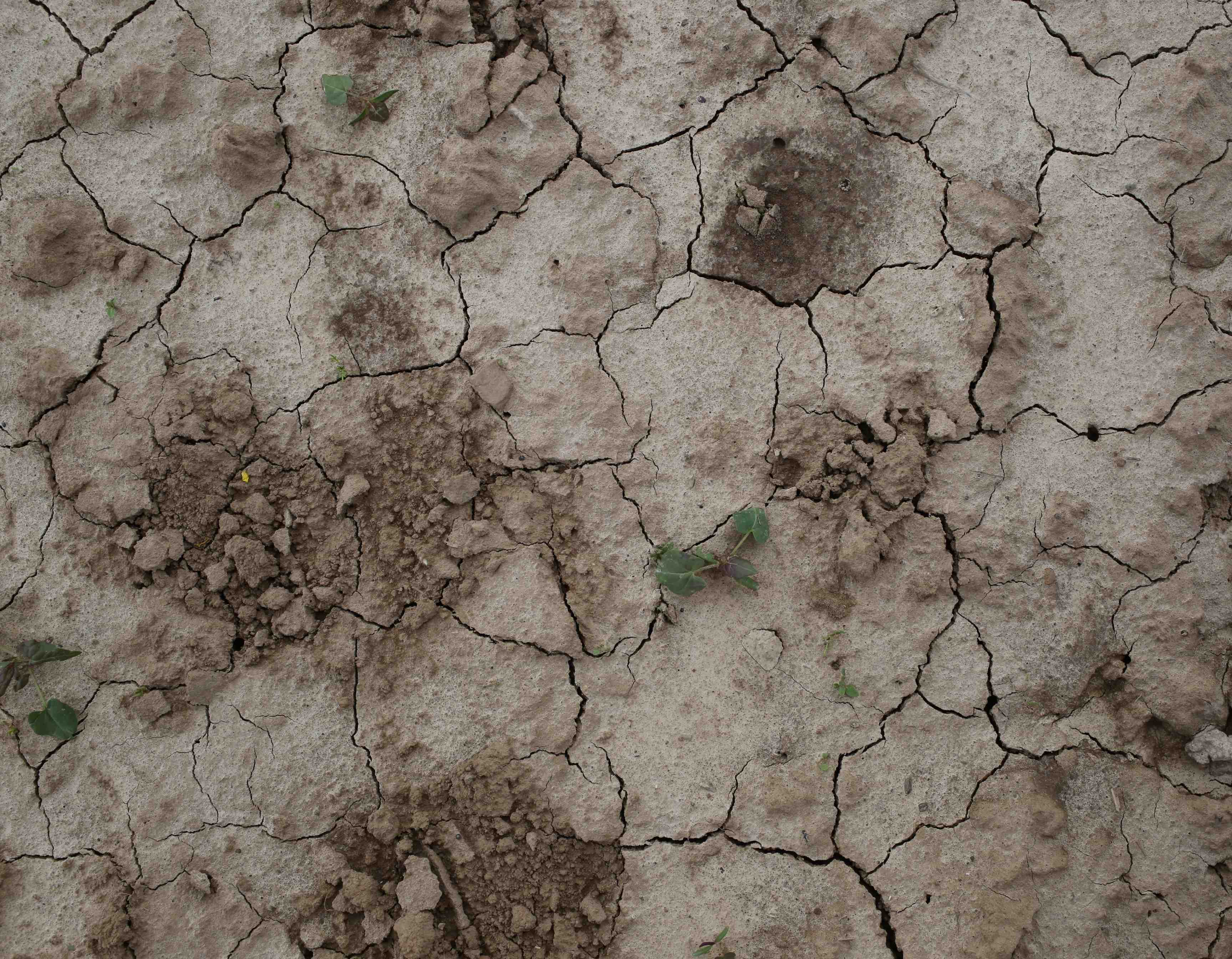}
        \caption{\textit{Fallopia convolvulus} (L.) Á.  Löve}
    \end{subfigure}
    \begin{subfigure}{0.3\textwidth}
        \centering
        \includegraphics[width=\textwidth]{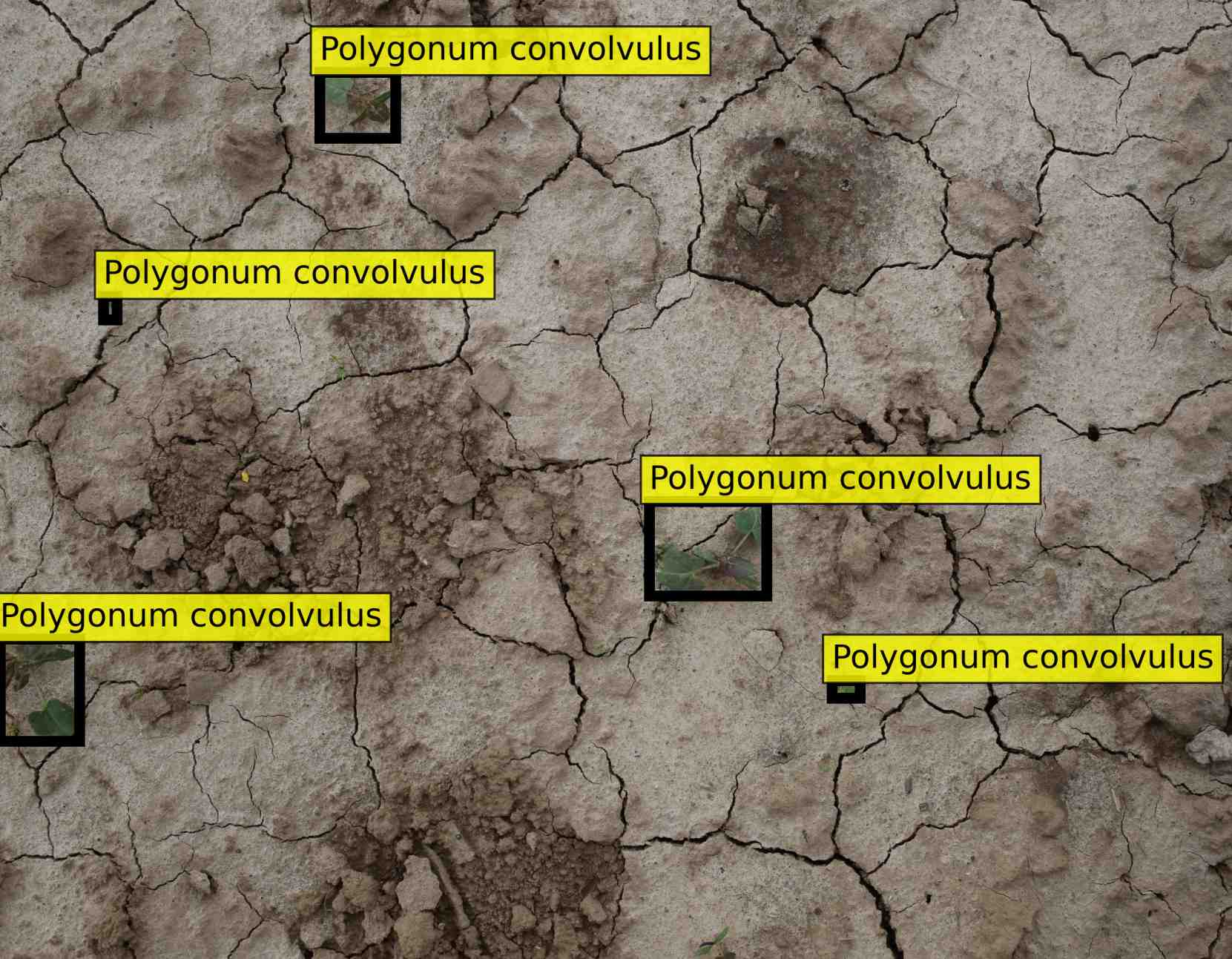}
        \caption{Dataset 1}
    \end{subfigure}
    \begin{subfigure}{0.3\textwidth}
        \centering
        \includegraphics[width=\textwidth]{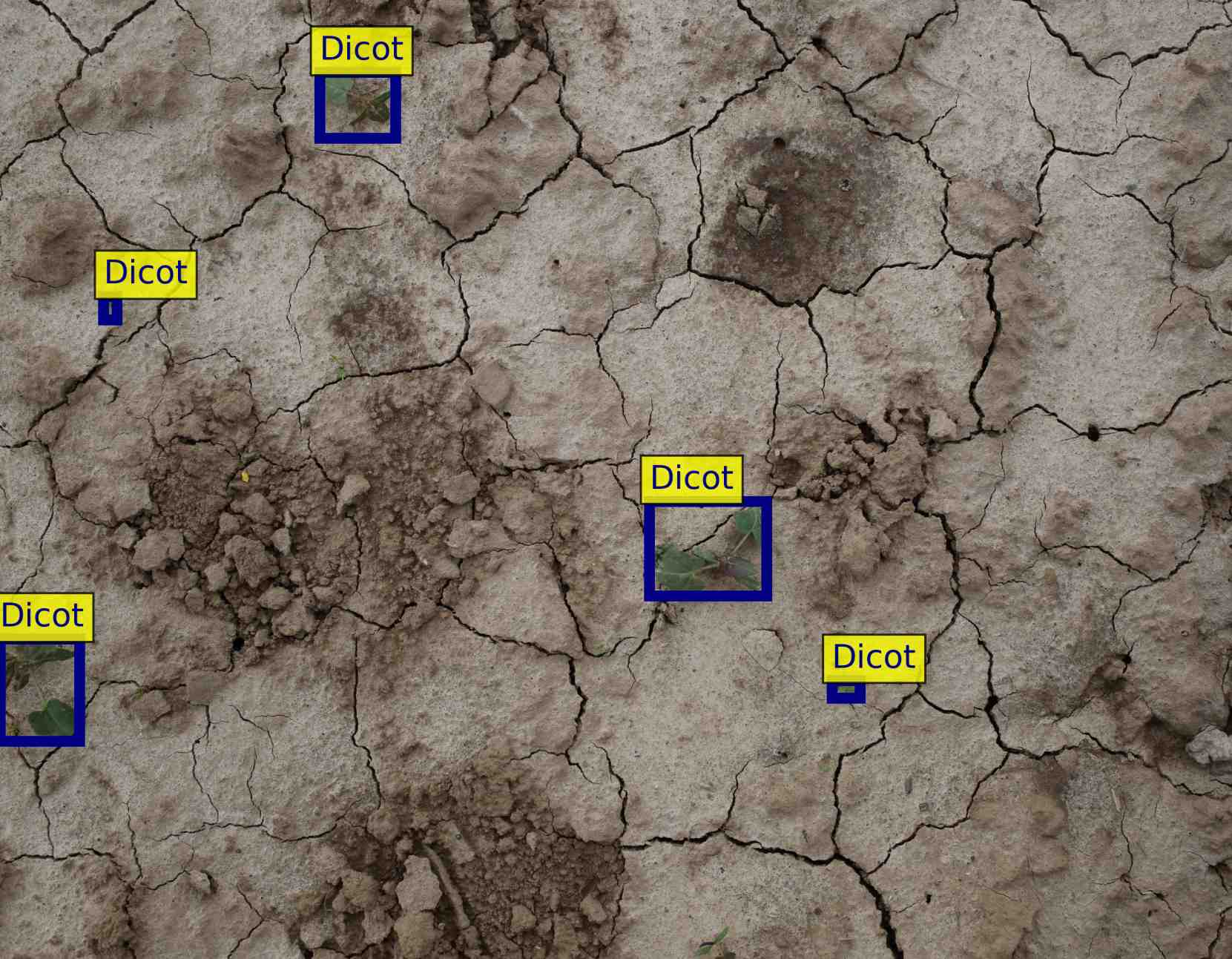}
        \caption{Dataset 2}
    \end{subfigure}

    \begin{subfigure}{0.3\textwidth}
        \centering
        \includegraphics[width=\textwidth]{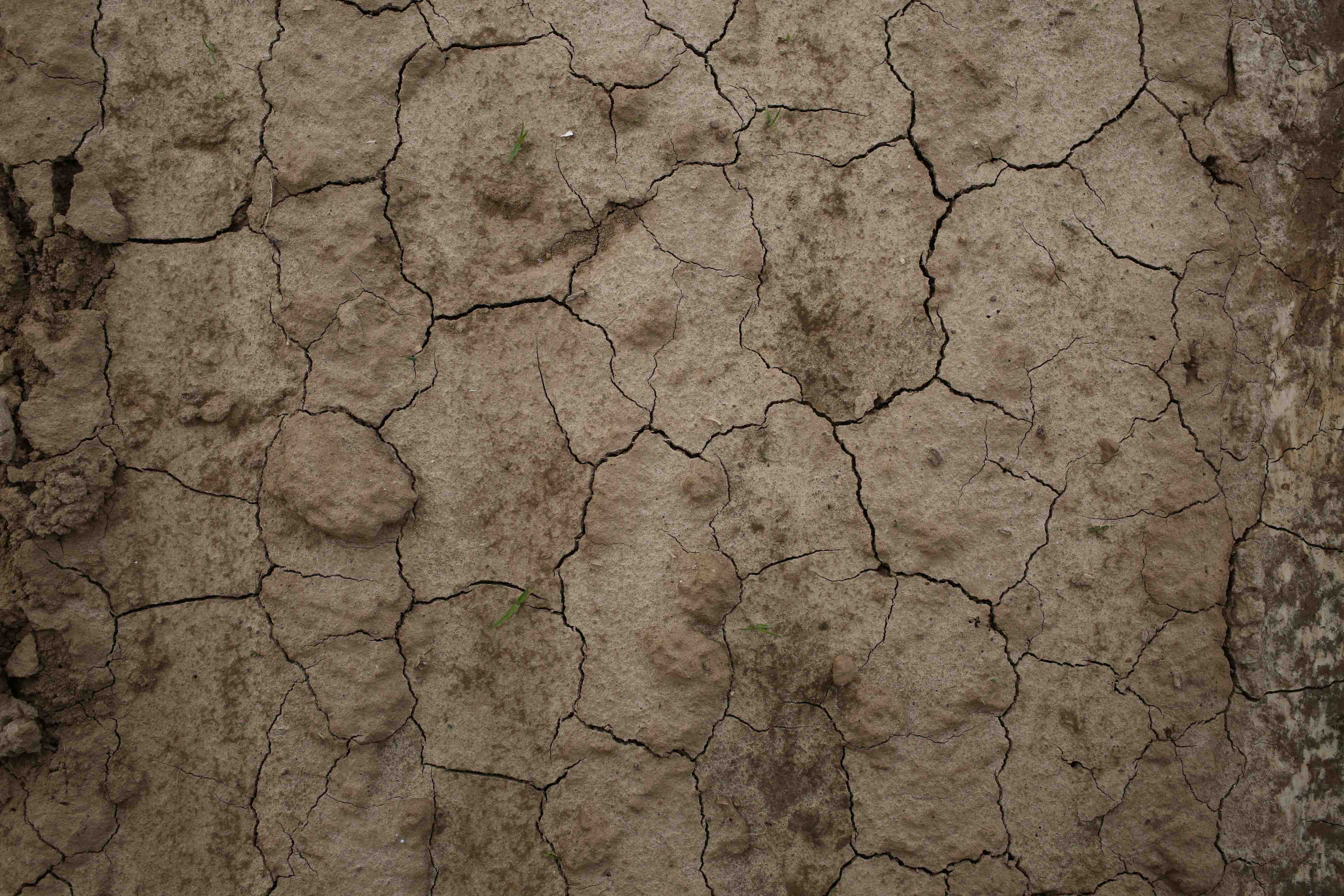}
        \caption{\textit{Setaria spp}}
    \end{subfigure}
    \begin{subfigure}{0.3\textwidth}
        \centering
        \includegraphics[width=\textwidth]{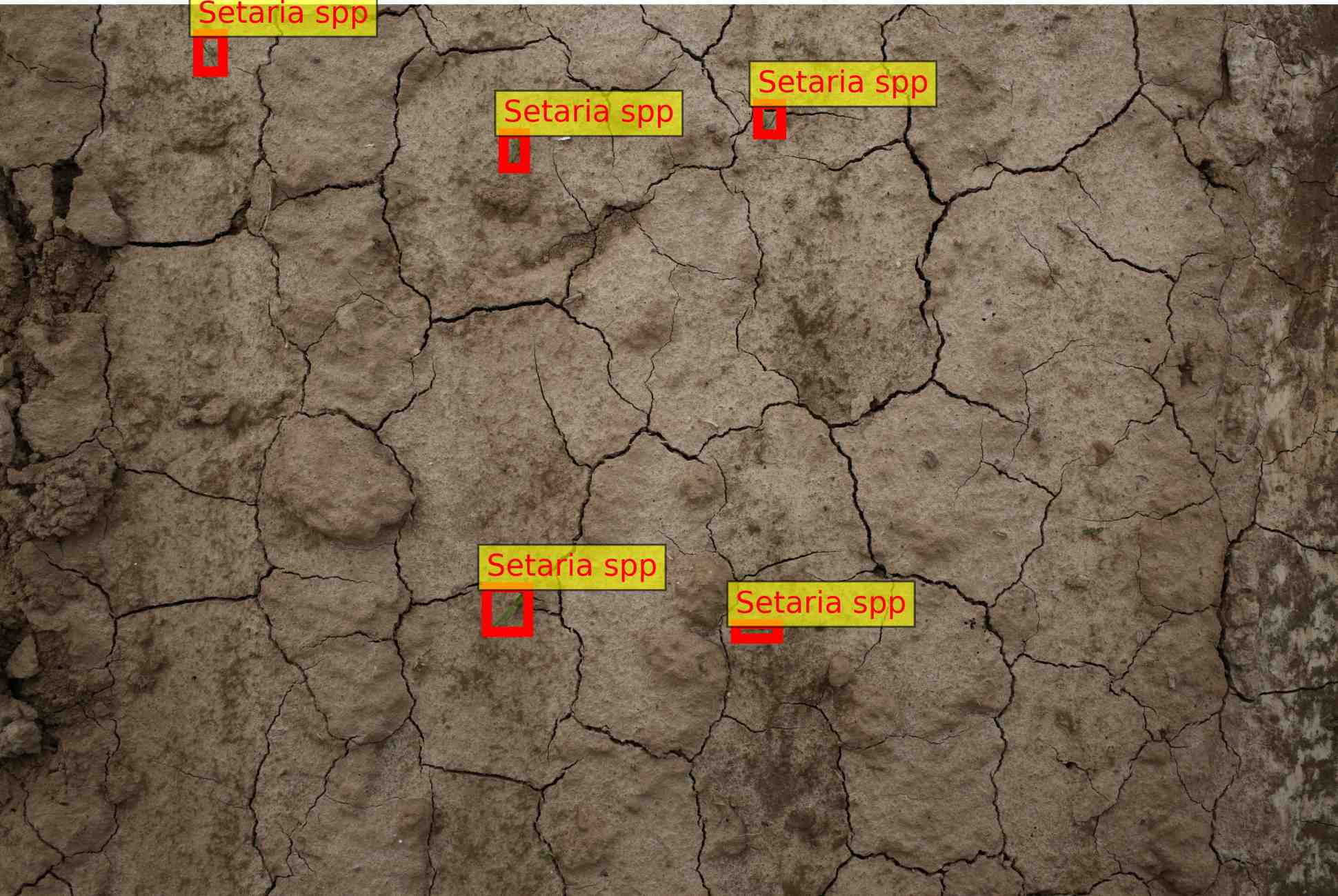}
        \caption{Dataset 1}
    \end{subfigure}
    \begin{subfigure}{0.3\textwidth}
        \centering
        \includegraphics[width=\textwidth]{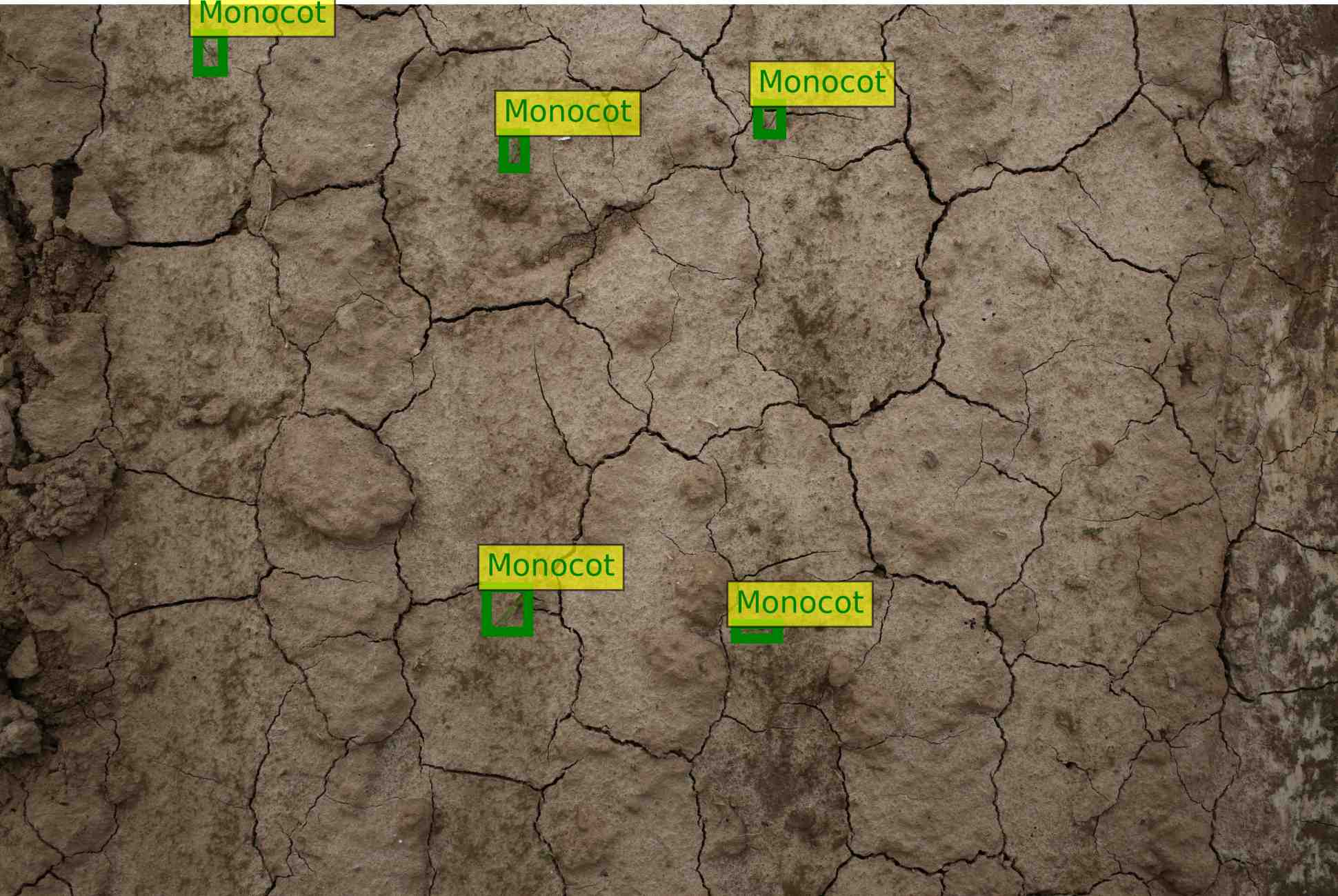}
        \caption{Dataset 2}
    \end{subfigure}

    \begin{subfigure}{0.3\textwidth}
        \centering
        \includegraphics[width=\textwidth]{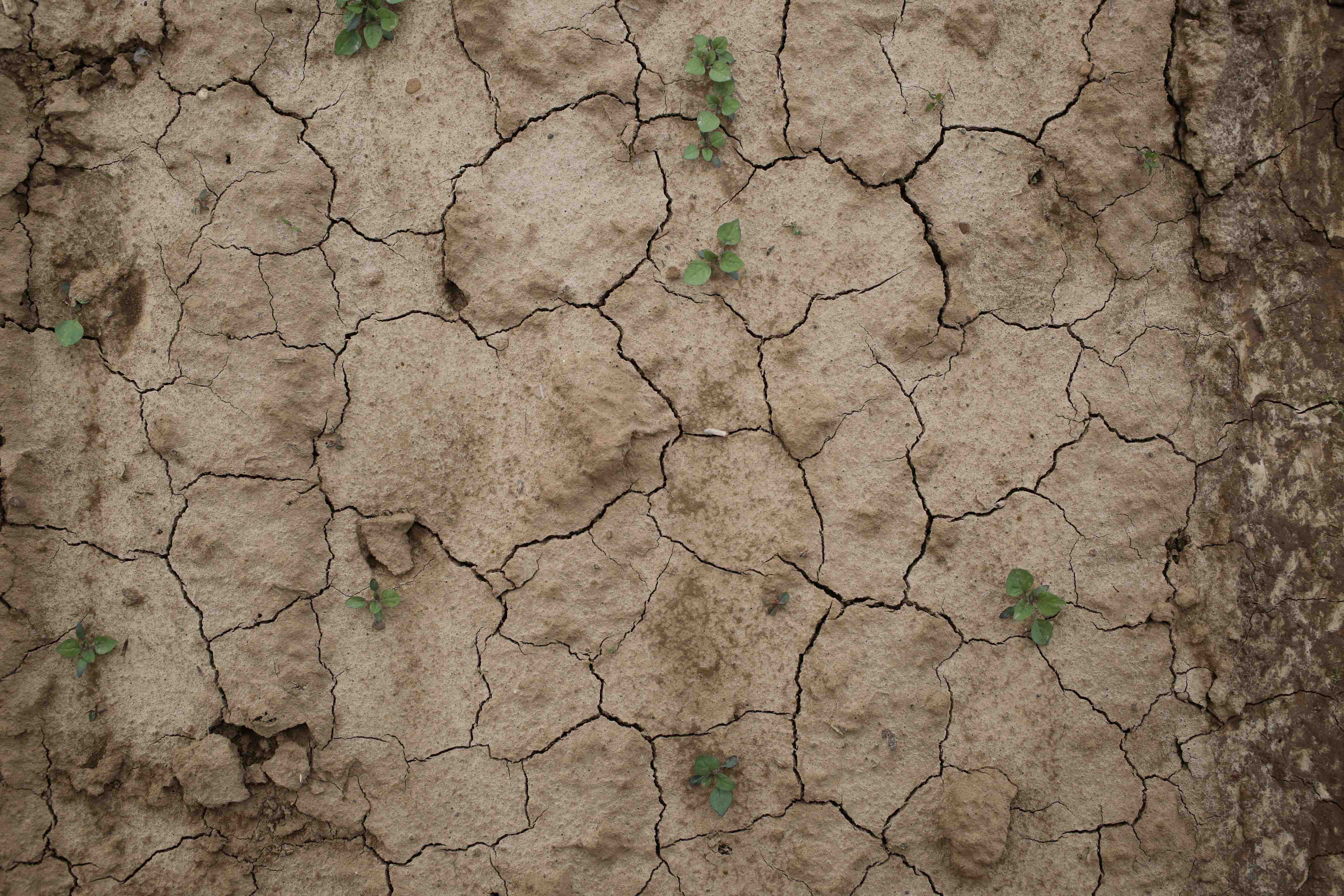}
        \caption{\textit{Solanum nigrum} L.}
    \end{subfigure}
    \begin{subfigure}{0.3\textwidth}
        \centering
        \includegraphics[width=\textwidth]{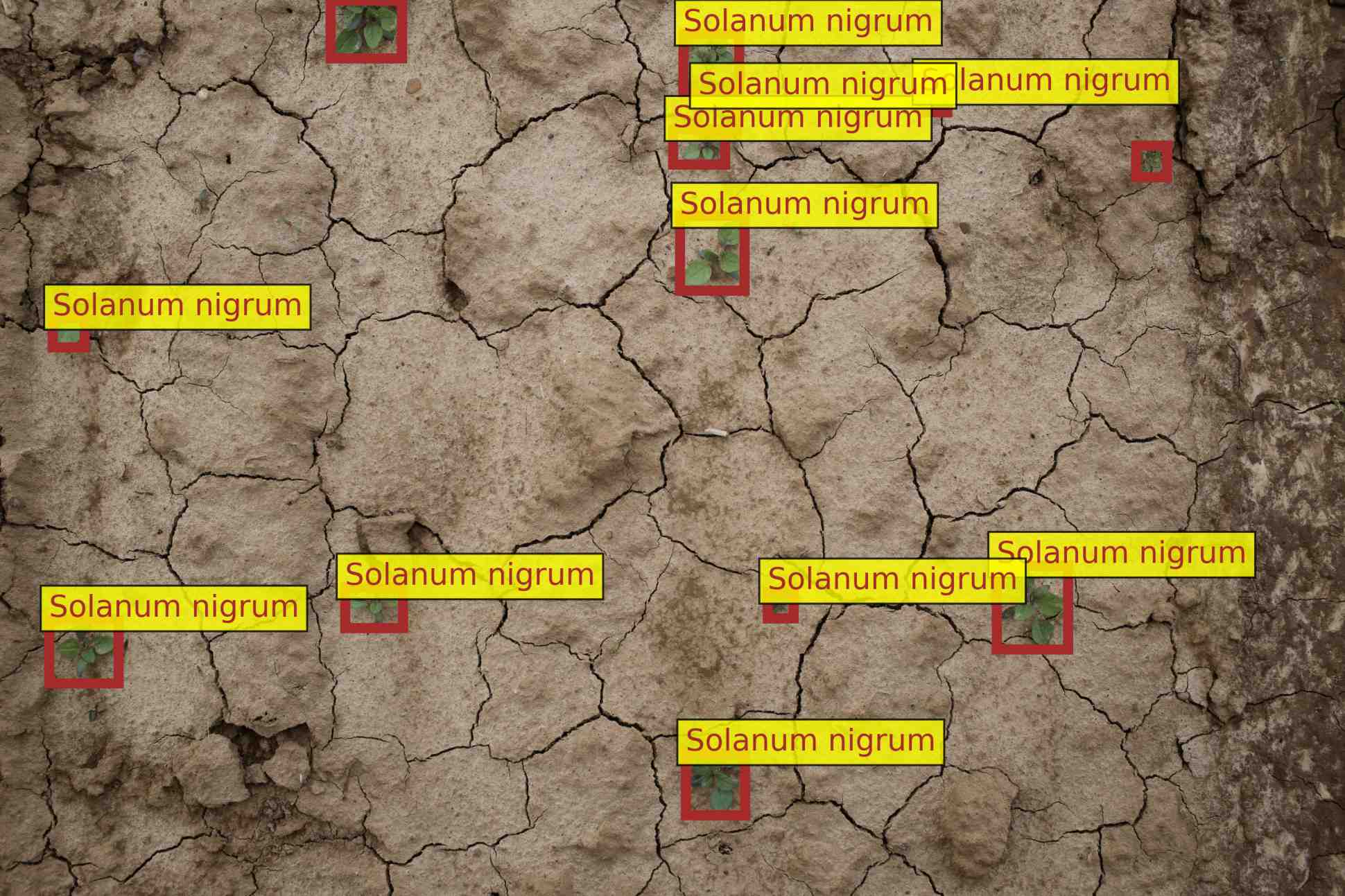}
        \caption{Dataset 1}
    \end{subfigure}
    \begin{subfigure}{0.3\textwidth}
        \centering
        \includegraphics[width=\textwidth]{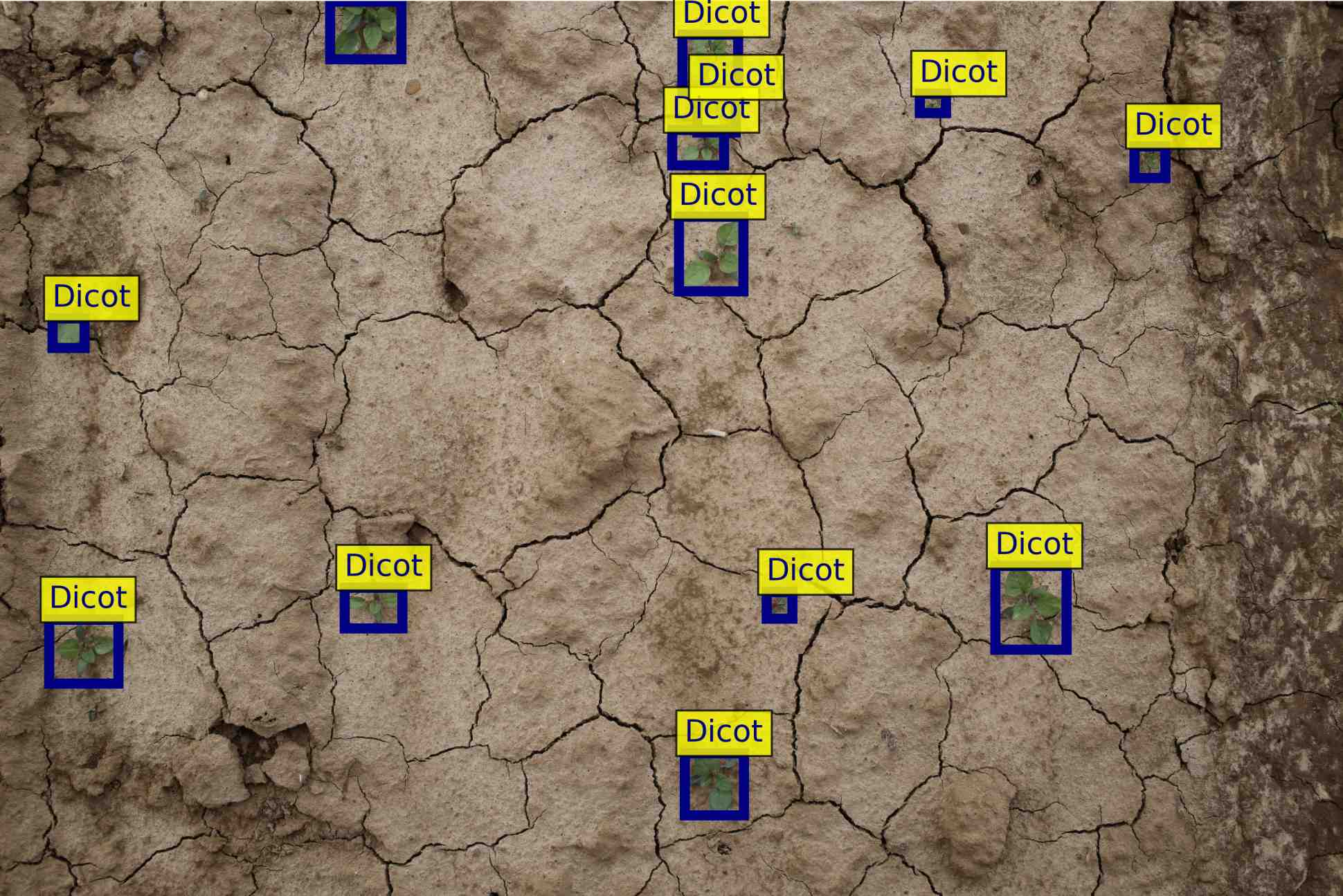}
        \caption{Dataset 2}
    \end{subfigure}

        \centering
    \begin{subfigure}{0.3\textwidth}
        \centering
        \includegraphics[width=\textwidth]{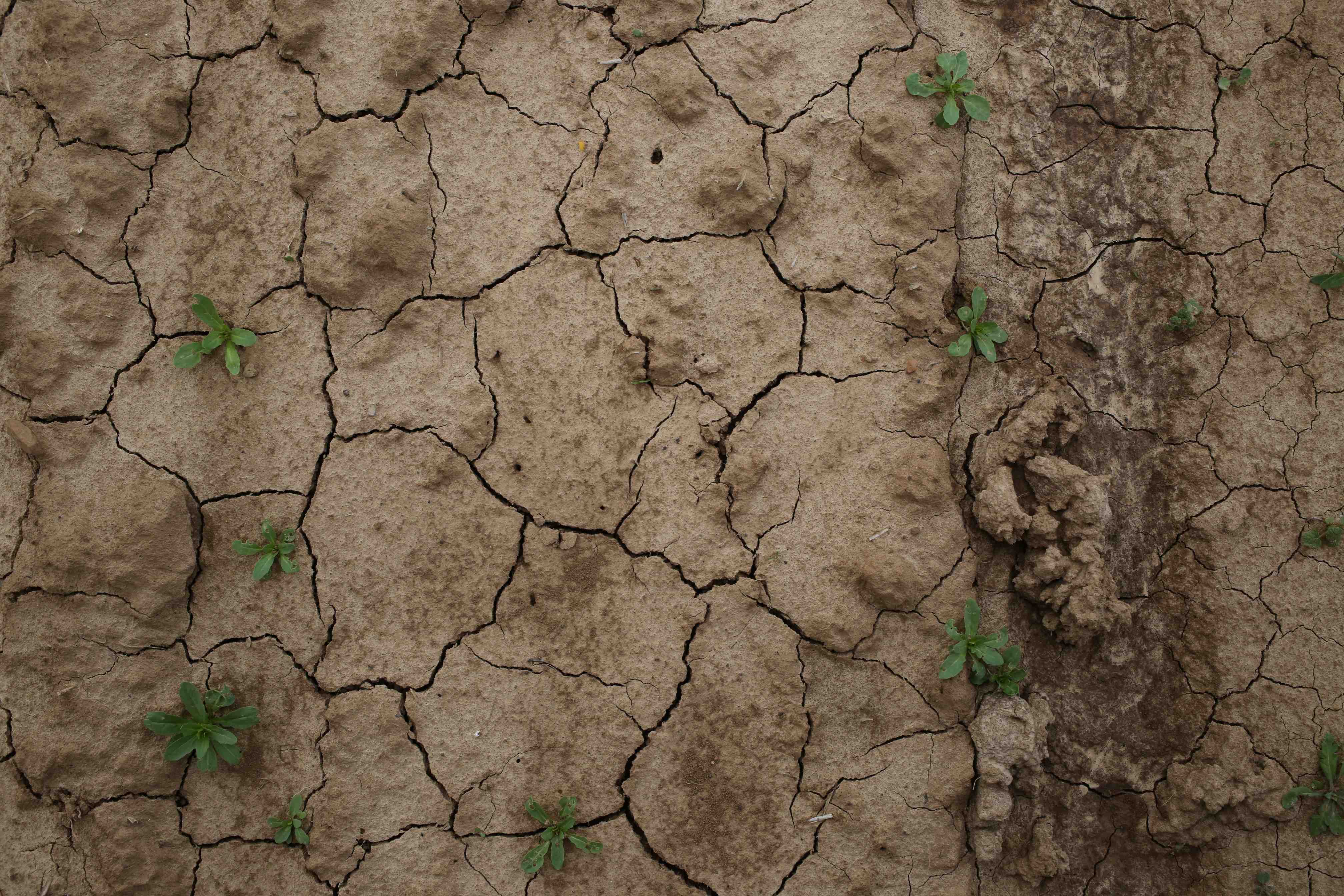}
        \caption{\textit{Thlaspi arvense} L.}
    \end{subfigure}
    \begin{subfigure}{0.3\textwidth}
        \centering
        \includegraphics[width=\textwidth]{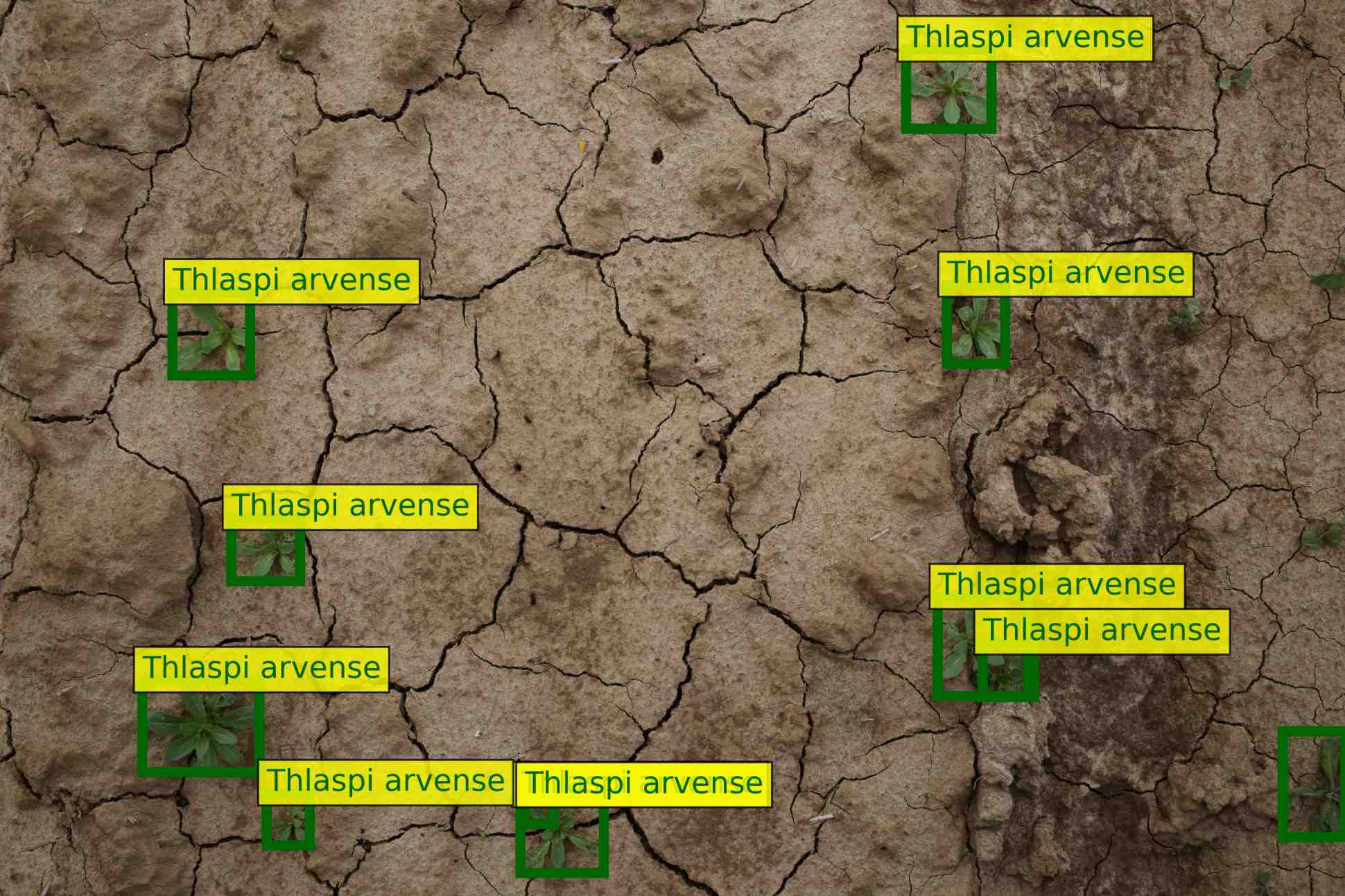}
        \caption{Dataset 1}
    \end{subfigure}
    \begin{subfigure}{0.3\textwidth}
        \centering
        \includegraphics[width=\textwidth]{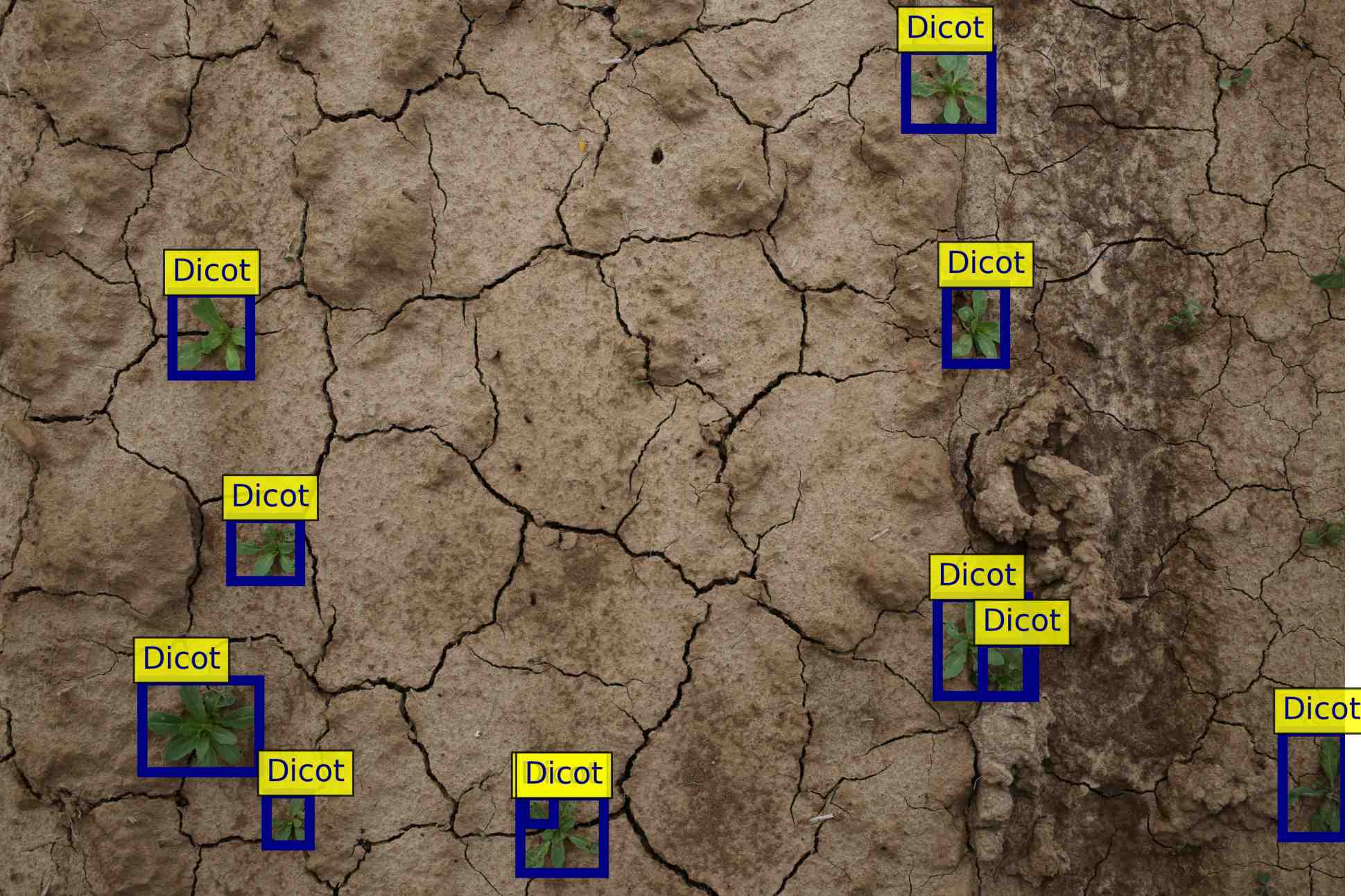}
        \caption{Dataset 2}
    \end{subfigure}

  \caption{Comparison of captured and annotated ground truth images for dataset 1 and dataset 2.}
  \label{fig:comparison_groundtruth_annotated2}
\end{figure}

\begin{figure}[H]

    \begin{subfigure}{0.3\textwidth}
        \centering
        \includegraphics[width=\textwidth]{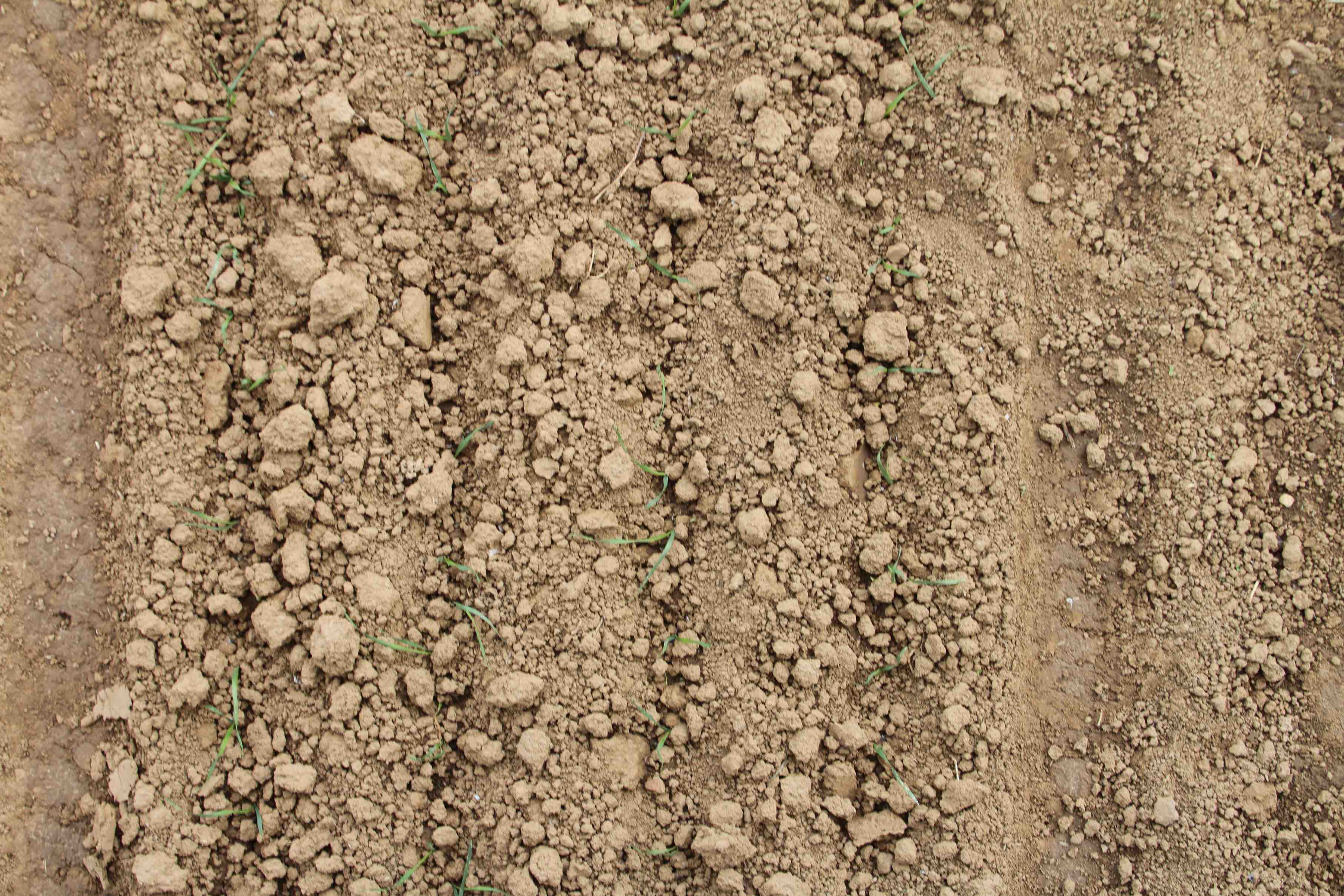}
        \caption{\textit{Triticum aestivum} L.}
    \end{subfigure}
    \begin{subfigure}{0.3\textwidth}
        \centering
        \includegraphics[width=\textwidth]{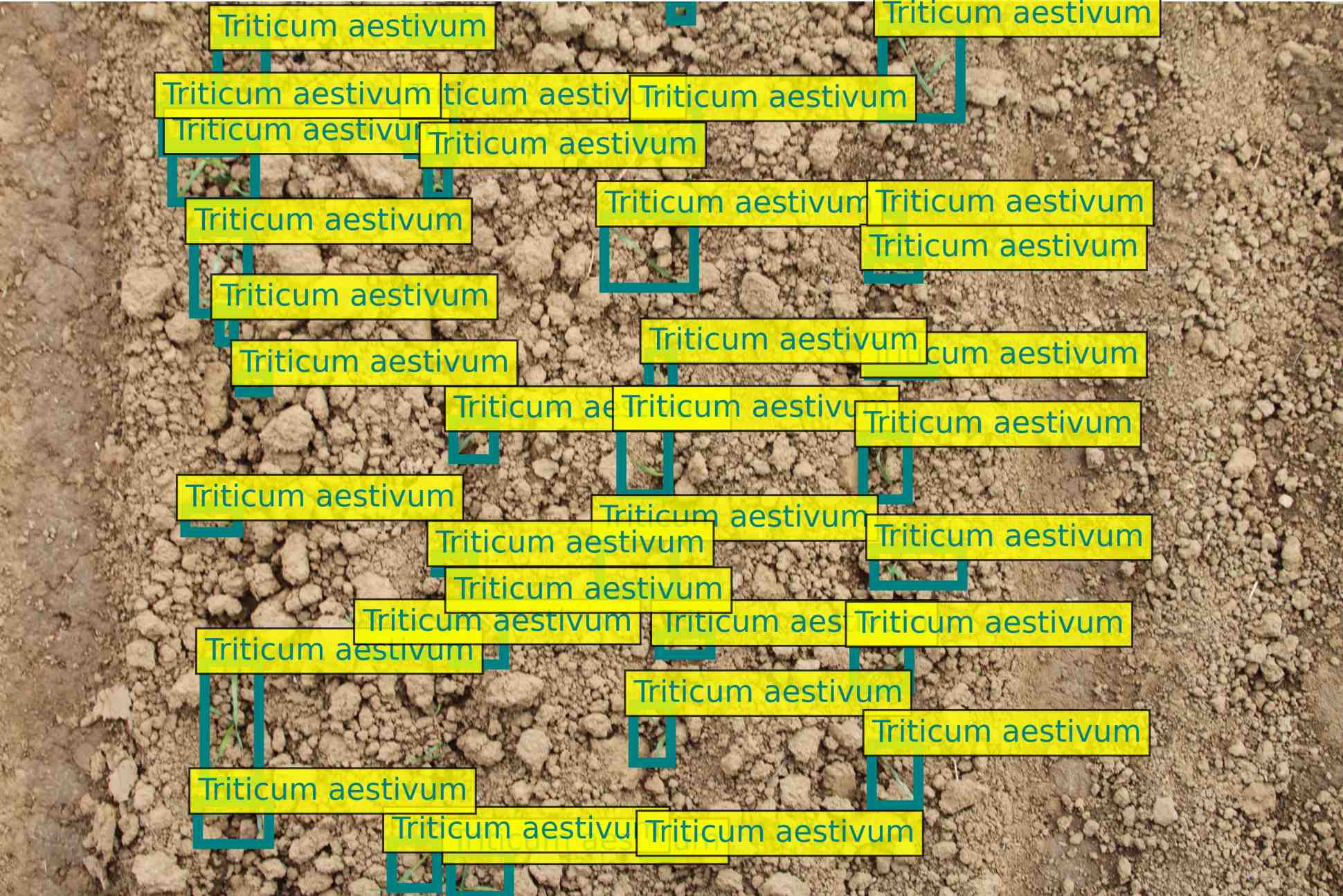}
        \caption{Dataset 1}
    \end{subfigure}
    \begin{subfigure}{0.3\textwidth}
        \centering
        \includegraphics[width=\textwidth]{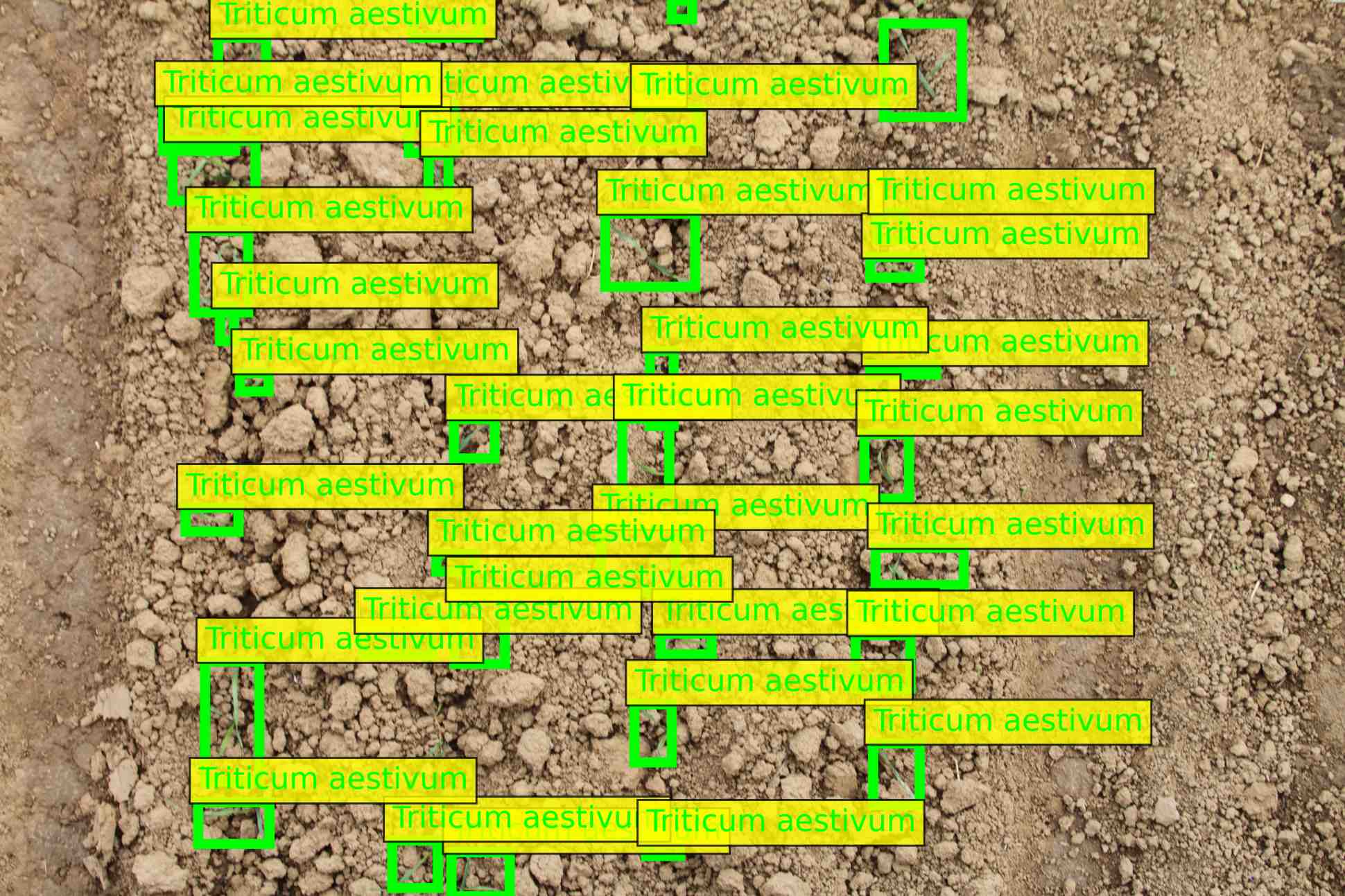}
        \caption{Dataset 2}
    \end{subfigure}

    \begin{subfigure}{0.3\textwidth}
        \centering
        \includegraphics[width=\textwidth]{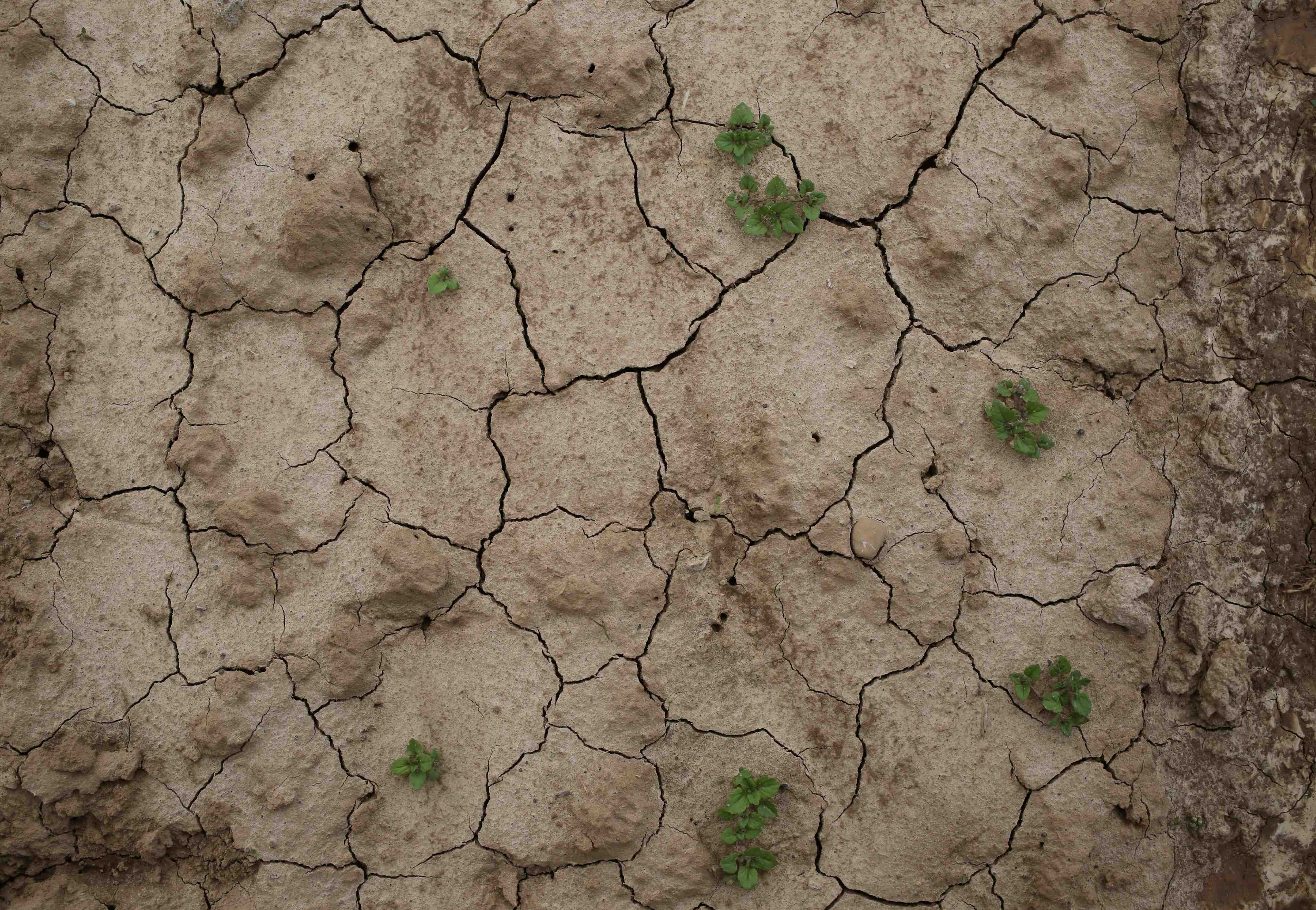}
        \caption{\textit{Veronica persica} Poir.}
    \end{subfigure}
    \begin{subfigure}{0.3\textwidth}
        \centering
        \includegraphics[width=\textwidth]{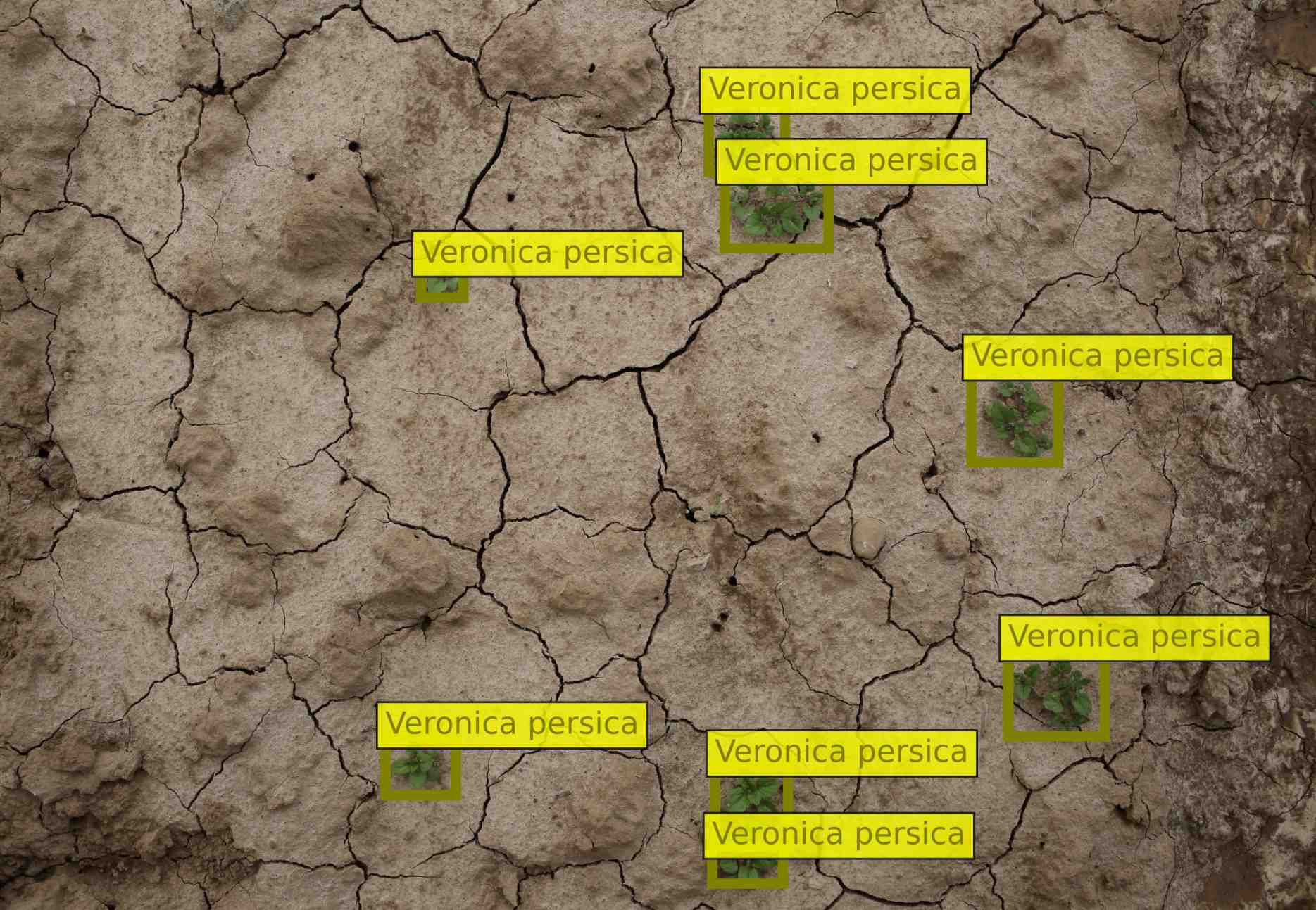}
        \caption{Dataset 1}
    \end{subfigure}
    \begin{subfigure}{0.3\textwidth}
        \centering
        \includegraphics[width=\textwidth]{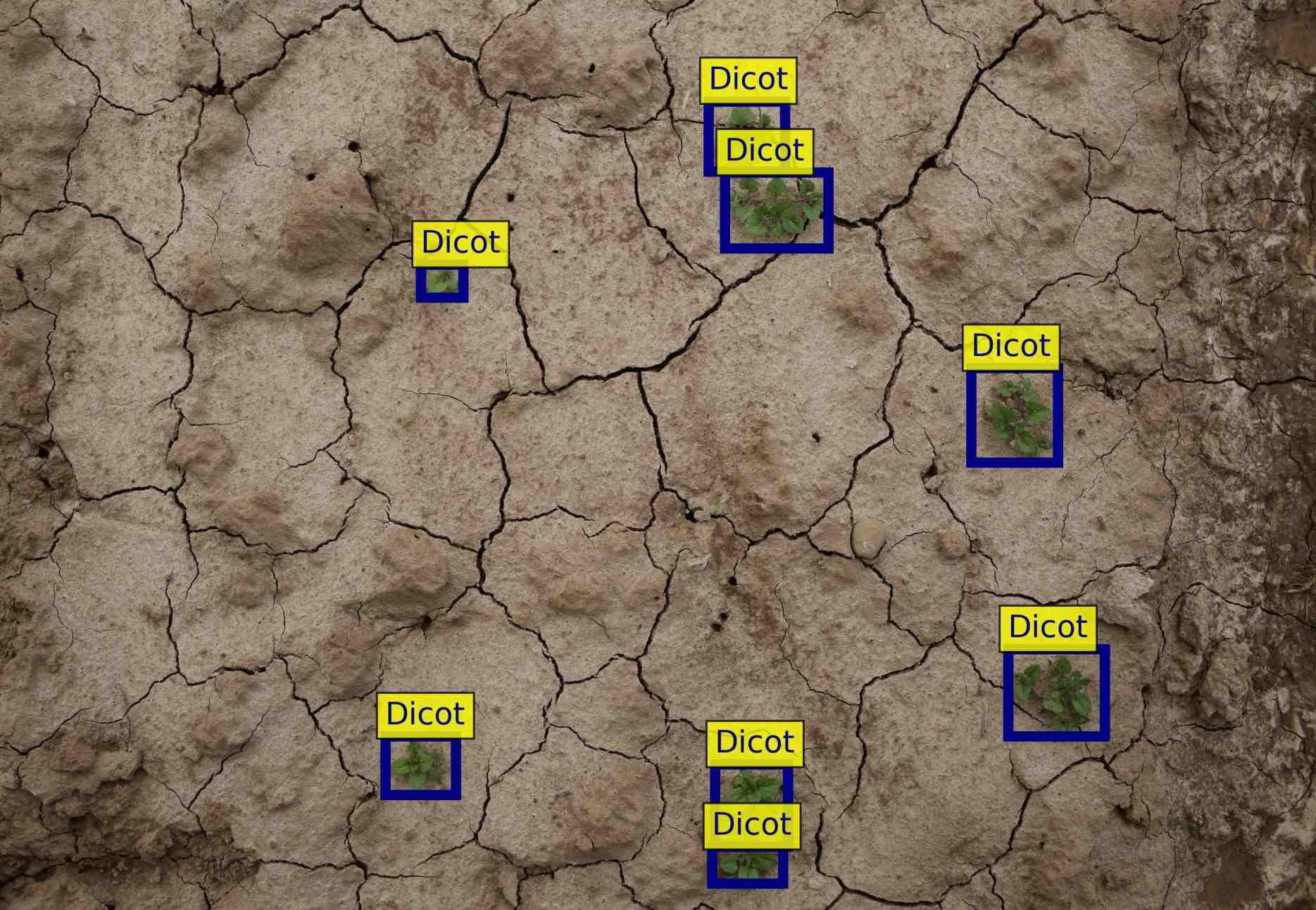}
        \caption{Dataset 2}
    \end{subfigure}

    \begin{subfigure}{0.3\textwidth}
        \centering
        \includegraphics[width=\textwidth]{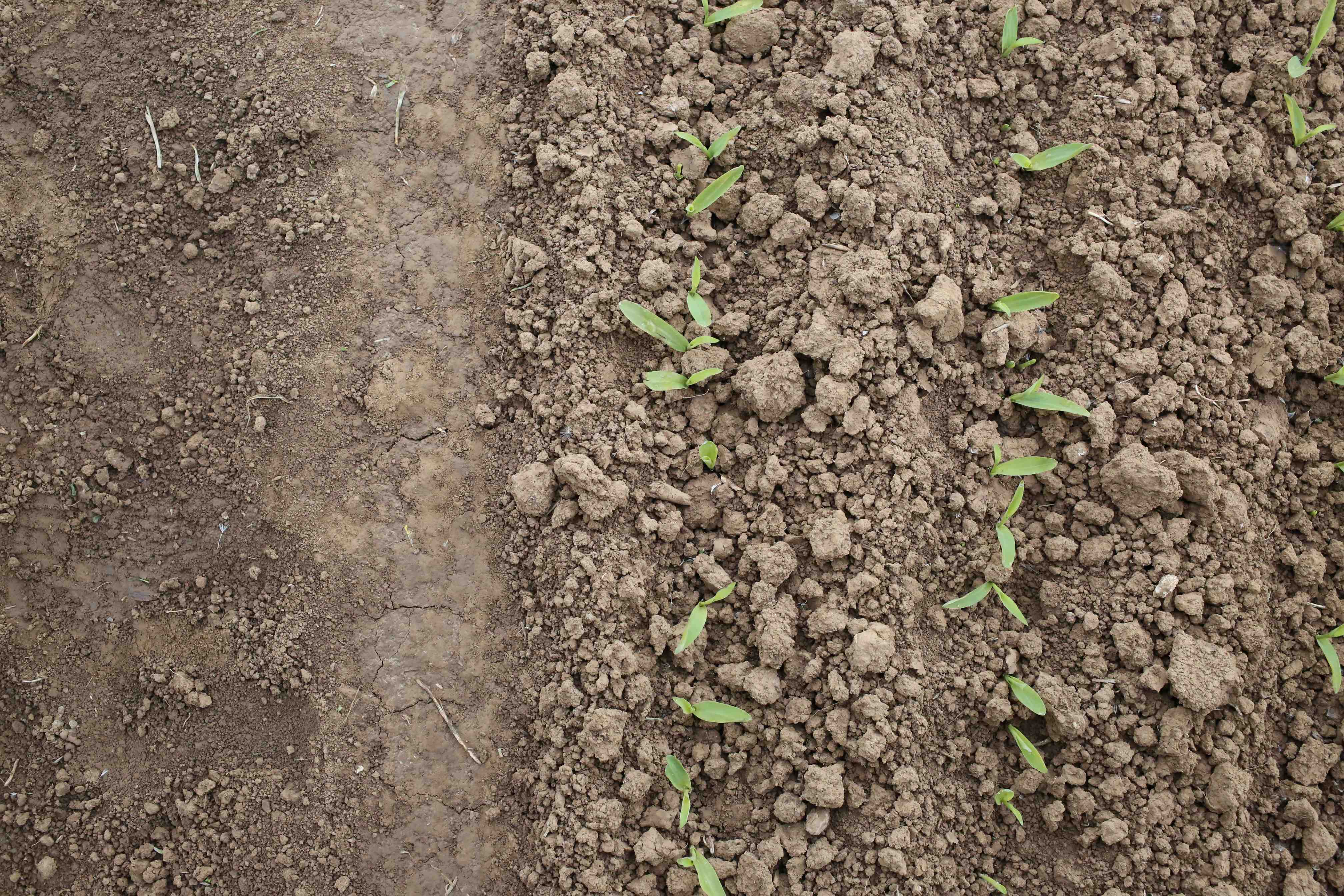}
        \caption{\textit{Zea mays} L. }
    \end{subfigure}
    \begin{subfigure}{0.3\textwidth}
        \centering
        \includegraphics[width=\textwidth]{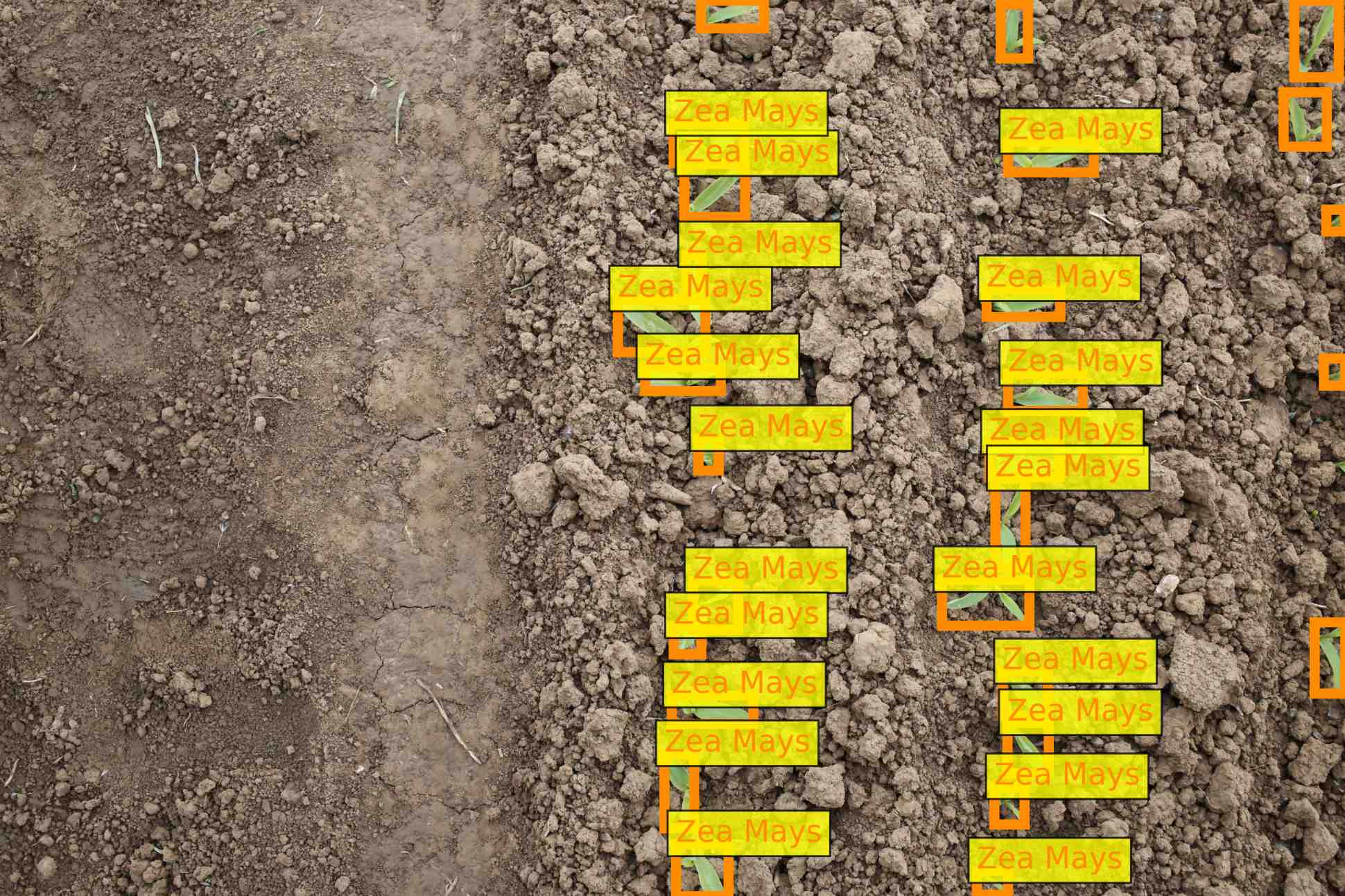}
        \caption{Dataset 1}
    \end{subfigure}
    \begin{subfigure}{0.3\textwidth}
        \centering
        \includegraphics[width=\textwidth]{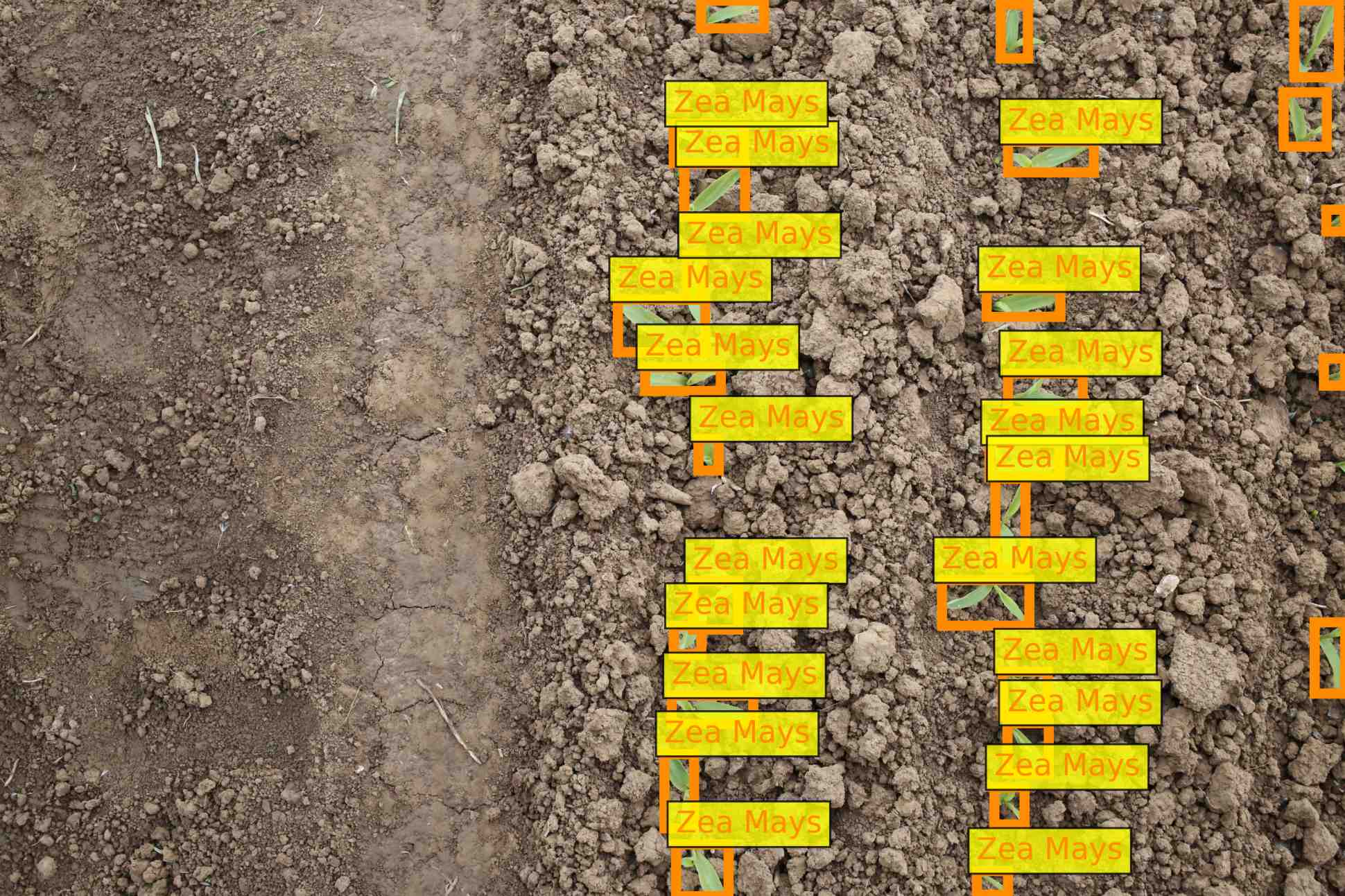}
        \caption{Dataset 2}
    \end{subfigure}

  \caption{Comparison of captured and annotated ground truth images for dataset 1 and dataset 2.}
  \label{fig:comparison_groundtruth_annotated3}
\end{figure}

\subsection{Supplementary B: Detailed Class-wise Results}

\subsubsection{Dataset 1}
\subsubsection{Performance of YOLOv8}

\begin{table}[H]
\caption{Performance of YOLOv8x, Images and Instances per species, Precision, Recall, AP50 and AP50-95, with standard deviations (+/- SD) indicated. Bold values represent the highest values.}
\resizebox{\textwidth}{!}{%
\begin{tabular}{c|c|c|c|c|c|c}
\toprule
\textbf{Class} & \textbf{Images} & \textbf{Instances} & \textbf{Precision} & \textbf{Recall} & \textbf{AP50} & \textbf{AP50-95} \\
\midrule
\textit{Abutilon theophrasti} Medik. & 48 & 1166 & 74.66 ± 1.38 & 76.90 ± 1.06 & 79.06 ± 0.71 & 43.24 ± 0.79 \\ 
\textit{Elymus repens} (L.) Gould & 53 & 309 & 69.44 ± 2.13 & 72.62 ± 3.52 & 73.56 ± 2.87 & 34.38 ± 1.62 \\ 
\textit{Alopecurus myosuroides} Huds. & 77 & 635 & \textbf{91.58 ± 0.66} & \textbf{81.86 ± 0.71} & \textbf{88.18 ± 0.36} & 49.12 ± 0.95 \\ 
\textit{Amaranthus retroflexus} L. & 94 & 418 & 86.58 ± 1.36 & 59.14 ± 0.76 & 69.40 ± 0.51 & 51.66 ± 0.53 \\ 
\textit{Avena fatua} L. & 77 & 1131 & 78.22 ± 2.04 & 66.74 ± 1.85 & 74.14 ± 1.97 & 40.22 ± 1.68 \\ 
\textit{Chenopodium album} L. & 101 & 329 & 78.34 ± 3.74 & 54.68 ± 1.37 & 65.14 ± 1.11 & 39.14 ± 1.16 \\ 
\textit{Geranium spp.} & 53 & 1347 & 79.60 ± 2.62 & 59.98 ± 1.14 & 67.22 ± 3.19 & 32.00 ± 1.57 \\ 
\textit{Helianthus annuus} L. & 56 & 1873 & 77.48 ± 2.99 & 75.26 ± 2.52 & 82.54 ± 1.52 & \textbf{59.80 ± 1.13} \\ 
\textit{Lamium purpureum} L. & 214 & 1069 & 90.78 ± 0.84 & 58.12 ± 1.26 & 69.50 ± 0.76 & 40.90 ± 0.44 \\ 
\textit{Fallopia convolvulus} (L.) Á.  Löve & 78 & 115 & 76.88 ± 4.18 & 57.40 ± 1.59 & 62.56 ± 0.86 & 44.60 ± 1.27 \\ 
\textit{Setaria spp.} & 134 & 275 & 76.44 ± 4.08 & 43.66 ± 2.32 & 50.78 ± 1.56 & 30.30 ± 0.79 \\ 
\textit{Solanum nigrum} L. & 57 & 331 & 80.44 ± 1.39 & 81.62 ± 1.00 & 86.14 ± 0.59 & 52.80 ± 0.73 \\ 
\textit{Thlaspi arvense} L. & 273 & 1208 & 88.34 ± 1.08 & 51.10 ± 1.20 & 65.34 ± 0.66 & 41.94 ± 0.29 \\ 
\textit{Triticum aestivum} L. & 58 & 1772 & 74.48 ± 1.99 & 68.00 ± 0.16 & 72.26 ± 0.69 & 32.12 ± 0.54 \\ 
\textit{Veronica persica} Poir. & 108 & 322 & 81.36 ± 2.60 & 67.94 ± 0.65 & 71.94 ± 0.67 & 38.72 ± 0.28 \\ 
\textit{Zea mays} L. & 65 & 1327 & 84.14 ± 1.73 & 76.98 ± 1.22 & 84.78 ± 0.61 & 51.78 ± 0.57 \\ 
\bottomrule
\end{tabular}

}
\end{table}

\begin{table}[H]
\caption{Performance of YOLOv8m, Images and Instances per species, Precision, Recall, AP50 and AP50-95, with standard deviations (+/- SD) indicated. Bold values represent the highest values.}
\resizebox{\textwidth}{!}{%
\begin{tabular}{c|c|c|c|c|c|c}
\toprule
\textbf{Class} & \textbf{Images} & \textbf{Instances} & \textbf{Precision} & \textbf{Recall} & \textbf{AP50} & \textbf{AP50-95} \\
\midrule
\textit{Abutilon theophrasti} Medik. & 48 & 1166 & 74.62 ± 1.41 & 75.50 ± 1.83 & 78.48 ± 0.89 & 43.48 ± 1.21 \\ 
\textit{Elymus repens} (L.) Gould & 53 & 309 & 68.76 ± 3.37 & 70.42 ± 2.20 & 71.76 ± 1.57 & 33.44 ± 1.90 \\ 
\textit{Alopecurus myosuroides} Huds. & 77 & 635 & 87.14 ± 3.60 & \textbf{81.52 ± 0.50} & \textbf{86.72 ± 1.47} & 47.46 ± 1.69 \\ 
\textit{Amaranthus retroflexus} L. & 94 & 418 & 86.94 ± 2.46 & 57.80 ± 1.60 & 68.40 ± 1.83 & 50.62 ± 1.72 \\ 
\textit{Avena fatua} L. & 77 & 1131 & 78.84 ± 1.08 & 62.58 ± 2.31 & 72.04 ± 1.41 & 39.18 ± 1.15 \\ 
\textit{Chenopodium album} L. & 101 & 329 & 79.06 ± 2.45 & 53.34 ± 1.70 & 65.06 ± 1.11 & 39.44 ± 0.67 \\ 
\textit{Geranium spp.} & 53 & 1347 & 78.70 ± 1.93 & 58.10 ± 1.75 & 65.08 ± 1.68 & 29.96 ± 1.05 \\ 
\textit{Helianthus annuus} L. & 56 & 1873 & 78.98 ± 1.27 & 75.68 ± 0.97 & 83.18 ± 1.05 & \textbf{60.60 ± 1.12} \\ 
\textit{Lamium purpureum} L. & 214 & 1069 & \textbf{89.82 ± 1.78} & 58.38 ± 1.39 & 69.62 ± 0.45 & 40.64 ± 0.35 \\ 
\textit{Fallopia convolvulus} (L.) Á.  Löve & 78 & 115 & 78.88 ± 4.70 & 55.30 ± 2.63 & 60.80 ± 0.89 & 42.42 ± 1.98 \\ 
\textit{Setaria spp.} & 134 & 275 & 76.92 ± 3.78 & 41.02 ± 1.68 & 47.40 ± 2.59 & 27.48 ± 1.55 \\ 
\textit{Solanum nigrum} L. & 57 & 331 & 78.04 ± 1.87 & 79.52 ± 2.17 & 83.50 ± 0.72 & 50.44 ± 0.47 \\ 
\textit{Thlaspi arvense} L. & 273 & 1208 & 86.64 ± 2.04 & 52.00 ± 2.25 & 65.22 ± 1.34 & 41.84 ± 0.74 \\ 
\textit{Triticum aestivum} L. & 58 & 1772 & 72.84 ± 1.49 & 67.60 ± 2.08 & 72.00 ± 0.80 & 32.26 ± 0.63 \\ 
\textit{Veronica persica} Poir. & 108 & 322 & 85.66 ± 1.80 & 68.42 ± 1.12 & 72.52 ± 0.99 & 39.66 ± 1.15 \\ 
\textit{Zea mays} L. & 65 & 1327 & 83.52 ± 1.00 & 77.98 ± 1.48 & 85.04 ± 1.24 & 51.44 ± 0.36 \\
\bottomrule
\end{tabular}

}
\end{table}

\begin{table}[H]
\caption{Performance of YOLOv8s, Images and Instances per species, Precision, Recall, AP50 and AP50-95, with standard deviations (+/- SD) indicated. Bold values represent the highest values.}
\resizebox{\textwidth}{!}{%
\begin{tabular}{c|c|c|c|c|c|c}
\toprule
\textbf{Class} & \textbf{Images} & \textbf{Instances} & \textbf{Precision} & \textbf{Recall} & \textbf{AP50} & \textbf{AP50-95} \\
\midrule
\textit{Abutilon theophrasti} Medik. & 48 & 1166 & 76.96 ± 0.98 & 75.48 ± 2.76 & 79.22 ± 0.83 & 44.42 ± 0.96 \\ 
\textit{Elymus repens} (L.) Gould & 53 & 309 & 70.90 ± 1.75 & 73.88 ± 4.37 & 75.66 ± 1.66 & 34.86 ± 1.78 \\ 
\textit{Alopecurus myosuroides} Huds. & 77 & 635 & \textbf{90.78 ± 0.83} & \textbf{81.92 ± 10.17} & 88.02 ± 0.15 & 48.92 ± 0.83 \\ 
\textit{Amaranthus retroflexus} L. & 94 & 418 & 85.38 ± 2.24 & 58.76 ± 2.80 & 69.10 ± 1.05 & 50.92 ± 0.46 \\ 
\textit{Avena fatua} L. & 77 & 1131 & 78.28 ± 1.06 & 65.60 ± 4.52 & 74.28 ± 1.52 & 40.20 ± 0.92 \\ 
\textit{Chenopodium album} L. & 101 & 329 & 77.00 ± 1.89 & 55.62 ± 1.38 & 64.18 ± 1.02 & 39.78 ± 0.91 \\ 
\textit{Geranium spp.} & 53 & 1347 & 81.22 ± 0.64 & 58.20 ± 7.30 & 66.58 ± 0.93 & 31.12 ± 0.36 \\ 
\textit{Helianthus annuus} L. & 56 & 1873 & 79.32 ± 1.36 & 75.26 ± 7.58 & 83.76 ± 1.20 & \textbf{60.64 ± 1.13} \\ 
\textit{Lamium purpureum} L. & 214 & 1069 & 90.38 ± 0.85 & 58.84 ± 1.24 & 69.36 ± 0.62 & 40.78 ± 0.67 \\ 
\textit{Fallopia convolvulus} (L.) Á.  Löve & 78 & 115 & 71.64 ± 5.91 & 58.44 ± 9.36 & 61.34 ± 1.05 & 42.50 ± 1.18 \\ 
\textit{Setaria spp.} & 134 & 275 & 75.06 ± 2.75 & 41.44 ± 16.09 & 47.88 ± 1.33 & 27.72 ± 0.67 \\ 
\textit{Solanum nigrum} L. & 57 & 331 & 77.42 ± 3.16 & 79.90 ± 14.49 & \textbf{83.42 ± 0.82} & 49.98 ± 1.16 \\ 
\textit{Thlaspi arvense} L. & 273 & 1208 & 87.16 ± 1.44 & 49.96 ± 6.57 & 64.48 ± 0.55 & 41.32 ± 0.28 \\ 
\textit{Triticum aestivum} L. & 58 & 1772 & 74.24 ± 1.63 & 66.18 ± 0.77 & 71.40 ± 1.47 & 31.24 ± 0.89 \\ 
\textit{Veronica persica} Poir. & 108 & 322 & 84.14 ± 1.76 & 67.96 ± 3.71 & 71.68 ± 1.36 & 39.98 ± 1.04 \\ 
\textit{Zea mays} L. & 65 & 1327 & 85.34 ± 0.76 & 76.22 ± 0.62 & 84.68 ± 0.86 & 51.44 ± 0.40 \\
\bottomrule
\end{tabular}

}
\end{table}

\begin{table}[H]
\caption{Performance of YOLOv8n, Images and Instances per species, Precision, Recall, AP50 and AP50-95, with standard deviations (+/- SD) indicated. Bold values represent the highest values.}
\resizebox{\textwidth}{!}{%
\begin{tabular}{c|c|c|c|c|c|c}
\toprule
\textbf{Class} & \textbf{Images} & \textbf{Instances} & \textbf{Precision} & \textbf{Recall} & \textbf{AP50} & \textbf{AP50-95} \\
\midrule
\textit{Abutilon theophrasti} Medik. & 48 & 1166 & 75.04 ± 1.28 & 74.44 ± 0.81 & 78.90 ± 0.53 & 44.04 ± 0.37 \\ 
\textit{Elymus repens} (L.) Gould & 53 & 309 & 71.74 ± 2.22 & 72.24 ± 0.29 & 75.18 ± 2.15 & 34.62 ± 1.55 \\ 
\textit{Alopecurus myosuroides} Huds. & 77 & 635 & \textbf{91.12 ± 0.56} & \textbf{81.02 ± 1.36} & \textbf{87.46 ± 0.47} & 48.30 ± 0.39 \\ 
\textit{Amaranthus retroflexus} L. & 94 & 418 & 85.18 ± 1.57 & 59.28 ± 0.72 & 69.26 ± 0.86 & 50.98 ± 0.59 \\ 
\textit{Avena fatua} L. & 77 & 1131 & 77.72 ± 1.56 & 65.18 ± 0.48 & 73.60 ± 0.97 & 39.56 ± 0.71 \\ 
\textit{Chenopodium album} L. & 101 & 329 & 80.08 ± 1.52 & 53.08 ± 1.40 & 65.04 ± 0.86 & 39.94 ± 0.94 \\ 
\textit{Geranium spp.} & 53 & 1347 & 81.82 ± 1.17 & 57.16 ± 0.74 & 66.22 ± 1.61 & 31.98 ± 0.82 \\ 
\textit{Helianthus annuus} L. & 56 & 1873 & 78.56 ± 0.70 & 75.28 ± 0.71 & 84.18 ± 0.41 & \textbf{60.80 ± 0.37} \\ 
\textit{Lamium purpureum} L. & 214 & 1069 & 90.06 ± 0.85 & 58.62 ± 0.50 & 68.98 ± 0.40 & 40.76 ± 0.51 \\ 
\textit{Fallopia convolvulus} (L.) Á.  Löve & 78 & 115 & 65.06 ± 2.01 & 57.82 ± 1.19 & 60.70 ± 1.50 & 41.58 ± 0.76 \\ 
\textit{Setaria spp.} & 134 & 275 & 79.22 ± 3.78 & 42.50 ± 2.81 & 50.70 ± 2.08 & 28.68 ± 1.18 \\ 
\textit{Solanum nigrum} L. & 57 & 331 & 75.60 ± 1.91 & 77.82 ± 1.44 & 82.86 ± 0.84 & 50.72 ± 0.69 \\ 
\textit{Thlaspi arvense} L. & 273 & 1208 & 86.72 ± 0.77 & 48.14 ± 0.89 & 63.40 ± 0.70 & 40.88 ± 0.42 \\ 
\textit{Triticum aestivum} L. & 58 & 1772 & 74.90 ± 1.06 & 65.44 ± 0.34 & 72.80 ± 0.99 & 32.02 ± 0.64 \\ 
\textit{Veronica persica} Poir. & 108 & 322 & 84.20 ± 1.14 & 66.00 ± 0.89 & 70.78 ± 0.79 & 39.90 ± 0.66 \\ 
\textit{Zea mays} L. & 65 & 1327 & 84.72 ± 0.75 & 75.68 ± 0.90 & 84.88 ± 0.59 & 51.76 ± 0.53 \\
\bottomrule
\end{tabular}

}
\end{table}

\subsubsection{Performance of YOLOv9}

\begin{table}[H]
\caption{Performance of YOLOv9e, Images and Instances per species, Precision, Recall, AP50 and AP50-95, with standard deviations (+/- SD) indicated. Bold values represent the highest values.}
\resizebox{\textwidth}{!}{%
\begin{tabular}{c|c|c|c|c|c|c}
\toprule
\textbf{Class} & \textbf{Images} & \textbf{Instances} & \textbf{Precision} & \textbf{Recall} & \textbf{AP50} & \textbf{AP50-95} \\
\midrule
\textit{Abutilon theophrasti} Medik. & 48 & 1166 & 76.86 ± 0.99 & 77.80 ± 0.64 & 80.32 ± 0.55 & 45.90 ± 0.82 \\ 
\textit{Elymus repens} (L.) Gould & 53 & 309 & 71.30 ± 2.24 & 75.28 ± 2.42 & 77.26 ± 2.47 & 37.36 ± 1.64 \\ 
\textit{Alopecurus myosuroides} Huds. & 77 & 635 & \textbf{91.66 ± 0.77} & 82.10 ± 1.13 & 88.30 ± 0.78 & 49.34 ± 1.19 \\ 
\textit{Amaranthus retroflexus} L. & 94 & 418 & 86.04 ± 1.57 & 59.34 ± 1.01 & 68.90 ± 0.48 & 51.84 ± 0.81 \\ 
\textit{Avena fatua} L. & 77 & 1131 & 79.38 ± 1.58 & 65.34 ± 0.63 & 74.28 ± 0.22 & 40.68 ± 0.51 \\ 
\textit{Chenopodium album} L. & 101 & 329 & 77.70 ± 2.67 & 56.54 ± 1.27 & 66.34 ± 1.94 & 39.80 ± 1.36 \\ 
\textit{Geranium spp.} & 53 & 1347 & 79.94 ± 1.22 & 59.18 ± 1.05 & 66.34 ± 1.54 & 31.28 ± 0.82 \\ 
\textit{Helianthus annuus} L. & 56 & 1873 & 79.38 ± 1.54 & 75.62 ± 1.58 & 83.48 ± 1.89 & \textbf{60.90 ± 1.59} \\ 
\textit{Lamium purpureum} L. & 214 & 1069 & 90.64 ± 1.37 & 58.24 ± 0.69 & 69.84 ± 0.94 & 40.76 ± 0.74 \\ 
\textit{Fallopia convolvulus} (L.) Á.  Löve & 78 & 115 & 81.96 ± 3.38 & 58.30 ± 1.35 & 63.50 ± 0.69 & 46.86 ± 1.10 \\ 
\textit{Setaria spp.} & 134 & 275 & 79.78 ± 1.57 & 44.74 ± 1.52 & 52.28 ± 1.68 & 32.04 ± 0.48 \\ 
\textit{Solanum nigrum} L. & 57 & 331 & 81.10 ± 2.94 & \textbf{82.84 ± 1.53} & \textbf{86.14 ± 0.91} & 51.30 ± 0.90 \\ 
\textit{Thlaspi arvense} L. & 273 & 1208 & 86.18 ± 2.03 & 52.98 ± 0.98 & 65.90 ± 1.17 & 42.74 ± 0.38 \\ 
\textit{Triticum aestivum} L. & 58 & 1772 & 73.94 ± 1.03 & 65.68 ± 1.28 & 71.06 ± 1.20 & 31.34 ± 0.65 \\ 
\textit{Veronica persica} Poir. & 108 & 322 & 83.90 ± 1.72 & 68.46 ± 1.38 & 72.28 ± 0.45 & 39.20 ± 1.02 \\ 
\textit{Zea mays} L. & 65 & 1327 & 84.02 ± 0.56 & 78.08 ± 1.01 & 84.70 ± 0.49 & 51.78 ± 0.58 \\
\bottomrule
\end{tabular}

}
\end{table}

\begin{table}[H]
\caption{Performance of YOLOv9m, Images and Instances per species, Precision, Recall, AP50 and AP50-95, with standard deviations (+/- SD) indicated. Bold values represent the highest values.}
\resizebox{\textwidth}{!}{%
\begin{tabular}{c|c|c|c|c|c|c}
\toprule
\textbf{Class} & \textbf{Images} & \textbf{Instances} & \textbf{Precision} & \textbf{Recall} & \textbf{AP50} & \textbf{AP50-95} \\
\midrule
\textit{Abutilon theophrasti} Medik. & 48 & 1166 & 74.12 ± 1.49 & 76.36 ± 1.85 & 79.08 ± 0.67 & 45.08 ± 0.83 \\ 
\textit{Elymus repens} (L.) Gould & 53 & 309 & 68.90 ± 2.51 & 74.14 ± 2.15 & 75.32 ± 3.18 & 36.12 ± 1.58 \\ 
\textit{Alopecurus myosuroides} Huds. & 77 & 635 & 90.30 ± 1.61 & \textbf{82.26 ± 1.17} & \textbf{87.88 ± 0.37} & 49.60 ± 0.78 \\ 
\textit{Amaranthus retroflexus} L. & 94 & 418 & 86.32 ± 0.78 & 59.52 ± 1.26 & 69.96 ± 1.71 & 52.18 ± 1.55 \\ 
\textit{Avena fatua} L. & 77 & 1131 & 78.16 ± 1.04 & 65.84 ± 1.18 & 74.32 ± 0.99 & 40.76 ± 0.64 \\ 
\textit{Chenopodium album} L. & 101 & 329 & 78.16 ± 3.56 & 57.20 ± 1.78 & 66.82 ± 1.54 & 40.76 ± 1.15 \\ 
\textit{Geranium spp.} & 53 & 1347 & 79.48 ± 0.86 & 60.06 ± 0.98 & 66.96 ± 1.27 & 31.90 ± 0.58 \\ 
\textit{Helianthus annuus} L. & 56 & 1873 & 79.34 ± 1.68 & 75.38 ± 1.58 & 83.80 ± 1.25 & \textbf{61.50 ± 1.00} \\ 
\textit{Lamium purpureum} L. & 214 & 1069 & \textbf{90.82 ± 0.87} & 58.84 ± 0.59 & 70.66 ± 0.87 & 41.40 ± 0.74 \\ 
\textit{Fallopia convolvulus} (L.) Á.  Löve & 78 & 115 & 75.18 ± 6.00 & 57.36 ± 2.48 & 62.16 ± 0.71 & 45.04 ± 0.63 \\ 
\textit{Setaria spp.} & 134 & 275 & 77.48 ± 3.99 & 42.06 ± 2.79 & 50.10 ± 1.56 & 30.62 ± 0.69 \\ 
\textit{Solanum nigrum} L. & 57 & 331 & 81.78 ± 1.64 & 79.82 ± 1.36 & 85.50 ± 0.83 & 52.12 ± 0.66 \\ 
\textit{Thlaspi arvense} L. & 273 & 1208 & 87.08 ± 1.53 & 53.12 ± 1.23 & 66.08 ± 0.36 & 42.46 ± 0.38 \\ 
\textit{Triticum aestivum} L. & 58 & 1772 & 73.56 ± 1.32 & 68.22 ± 1.74 & 72.52 ± 1.52 & 32.42 ± 1.19 \\ 
\textit{Veronica persica} Poir. & 108 & 322 & 85.52 ± 2.45 & 68.64 ± 0.80 & 71.68 ± 1.26 & 39.50 ± 0.80 \\ 
\textit{Zea mays} L. & 65 & 1327 & 83.76 ± 1.28 & 77.30 ± 1.29 & 84.54 ± 0.75 & 51.52 ± 0.94 \\
\bottomrule
\end{tabular}

}
\end{table}

\begin{table}[H]
\caption{Performance of YOLOv9s, Images and Instances per species, Precision, Recall, AP50 and AP50-95, with standard deviations (+/- SD) indicated. Bold values represent the highest values.}
\resizebox{\textwidth}{!}{%
\begin{tabular}{c|c|c|c|c|c|c}
\toprule
\textbf{Class} & \textbf{Images} & \textbf{Instances} & \textbf{Precision} & \textbf{Recall} & \textbf{AP50} & \textbf{AP50-95} \\
\midrule
\textit{Abutilon theophrasti} Medik. & 48 & 1166 & 73.76 ± 1.28 & 77.88 ± 0.90 & 79.38 ± 0.73 & 45.22 ± 0.80 \\ 
\textit{Elymus repens} (L.) Gould & 53 & 309 & 73.52 ± 2.39 & 75.76 ± 1.91 & 78.00 ± 1.76 & 37.16 ± 1.62 \\ 
\textit{Alopecurus myosuroides} Huds. & 77 & 635 & \textbf{91.84 ± 1.29} & \textbf{83.12 ± 0.90} & \textbf{89.08 ± 0.93} & 50.60 ± 0.91 \\ 
\textit{Amaranthus retroflexus} L. & 94 & 418 & 85.98 ± 1.60 & 59.86 ± 1.63 & 70.86 ± 1.75 & 52.58 ± 0.96 \\ 
\textit{Avena fatua} L. & 77 & 1131 & 78.32 ± 1.13 & 67.64 ± 0.47 & 75.28 ± 0.73 & 41.82 ± 0.77 \\ 
\textit{Chenopodium album} L. & 101 & 329 & 78.94 ± 2.98 & 56.66 ± 0.65 & 67.28 ± 1.60 & 41.16 ± 1.10 \\ 
\textit{Geranium spp.} & 53 & 1347 & 80.76 ± 1.15 & 58.96 ± 1.40 & 66.24 ± 1.59 & 31.48 ± 1.02 \\ 
\textit{Helianthus annuus} L. & 56 & 1873 & 79.90 ± 0.41 & 75.34 ± 0.68 & 83.92 ± 0.70 & \textbf{61.34 ± 0.92} \\ 
\textit{Lamium purpureum} L. & 214 & 1069 & 90.60 ± 1.06 & 58.98 ± 0.23 & 69.76 ± 0.90 & 41.00 ± 0.90 \\ 
\textit{Fallopia convolvulus} (L.) Á. Löve & 78 & 115 & 75.26 ± 4.88 & 59.28 ± 2.14 & 61.66 ± 0.93 & 46.42 ± 1.37 \\ 
\textit{Setaria spp.} & 134 & 275 & 78.16 ± 3.04 & 44.72 ± 0.78 & 52.28 ± 1.24 & 32.88 ± 0.95 \\ 
\textit{Solanum nigrum} L. & 57 & 331 & 80.38 ± 2.83 & 81.92 ± 1.85 & 85.98 ± 1.04 & 52.50 ± 1.08 \\ 
\textit{Thlaspi arvense} L. & 273 & 1208 & 87.56 ± 1.36 & 51.96 ± 1.44 & 65.00 ± 1.34 & 41.78 ± 0.68 \\ 
\textit{Triticum aestivum} L. & 58 & 1772 & 75.84 ± 1.18 & 67.60 ± 0.54 & 74.42 ± 0.82 & 33.08 ± 0.43 \\ 
\textit{Veronica persica} Poir. & 108 & 322 & 83.46 ± 1.81 & 68.64 ± 0.87 & 72.46 ± 1.30 & 40.62 ± 0.81 \\ 
\textit{Zea mays} L. & 65 & 1327 & 84.56 ± 0.73 & 77.18 ± 1.05 & 84.72 ± 0.24 & 51.58 ± 0.44 \\
\bottomrule
\end{tabular}

}
\end{table}

\begin{table}[H]
\caption{Performance of YOLOv9t, Images and Instances per species, Precision, Recall, AP50 and AP50-95, with standard deviations (+/- SD) indicated. Bold values represent the highest values.}
\resizebox{\textwidth}{!}{%
\begin{tabular}{c|c|c|c|c|c|c}
\toprule
\textbf{Class} & \textbf{Images} & \textbf{Instances} & \textbf{Precision} & \textbf{Recall} & \textbf{AP50} & \textbf{AP50-95} \\
\midrule
\textit{Abutilon theophrasti} Medik. & 48 & 1166 & 76.76 ± 0.95 & 75.48 ± 0.99 & 80.24 ± 0.59 & 45.00 ± 0.50 \\ 
\textit{Elymus repens} (L.) Gould & 53 & 309 & 77.64 ± 2.59 & 76.18 ± 2.49 & 79.62 ± 1.81 & 38.12 ± 0.61 \\ 
\textit{Alopecurus myosuroides} Huds. & 77 & 635 & \textbf{91.78 ± 0.53} & \textbf{81.80 ± 0.83} & \textbf{88.44 ± 0.54} & 50.46 ± 0.45 \\ 
\textit{Amaranthus retroflexus} L. & 94 & 418 & 83.42 ± 2.10 & 59.92 ± 1.85 & 70.88 ± 0.82 & 52.70 ± 0.94 \\ 
\textit{Avena fatua} L. & 77 & 1131 & 78.74 ± 0.97 & 67.86 ± 0.86 & 74.84 ± 0.57 & 40.40 ± 0.31 \\ 
\textit{Chenopodium album} L. & 101 & 329 & 83.36 ± 2.42 & 53.16 ± 0.86 & 66.60 ± 0.89 & 41.24 ± 0.36 \\ 
\textit{Geranium spp.} & 53 & 1347 & 81.60 ± 1.23 & 57.54 ± 0.77 & 66.16 ± 0.83 & 32.12 ± 0.66 \\ 
\textit{Helianthus annuus} L. & 56 & 1873 & 78.30 ± 1.67 & 75.32 ± 0.23 & 83.90 ± 0.72 & \textbf{61.16 ± 0.42} \\ 
\textit{Lamium purpureum} L. & 214 & 1069 & 89.36 ± 0.75 & 57.80 ± 0.81 & 68.34 ± 0.62 & 41.02 ± 0.33 \\ 
\textit{Fallopia convolvulus} (L.) Á.  Löve & 78 & 115 & 73.62 ± 4.61 & 58.24 ± 1.55 & 61.04 ± 1.05 & 45.50 ± 0.62 \\ 
\textit{Setaria spp.} & 134 & 275 & 79.62 ± 1.83 & 45.56 ± 1.26 & 54.24 ± 1.07 & 33.44 ± 0.41 \\ 
\textit{Solanum nigrum} L. & 57 & 331 & 80.22 ± 0.94 & 80.40 ± 1.54 & 85.90 ± 0.54 & 53.12 ± 0.26 \\ 
\textit{Thlaspi arvense} L. & 273 & 1208 & 84.42 ± 0.72 & 50.44 ± 0.47 & 62.10 ± 0.86 & 40.34 ± 0.72 \\ 
\textit{Triticum aestivum} L. & 58 & 1772 & 75.20 ± 0.66 & 66.12 ± 1.10 & 73.14 ± 0.44 & 32.90 ± 0.50 \\ 
\textit{Veronica persica} Poir. & 108 & 322 & 81.70 ± 2.61 & 68.88 ± 1.23 & 72.08 ± 0.57 & 41.32 ± 0.53 \\ 
\textit{Zea mays} L. & 65 & 1327 & 83.68 ± 1.29 & 77.20 ± 1.05 & 85.56 ± 0.73 & 52.08 ± 0.38 \\
\bottomrule
\end{tabular}

}
\end{table}

\subsubsection{Performance of YOLOv10}

\begin{table}[H]
\caption{Performance of YOLOv10x, Images and Instances per species, Precision, Recall, AP50 and AP50-95, with standard deviations (+/- SD) indicated. Bold values represent the highest values.}
\resizebox{\textwidth}{!}{%
\begin{tabular}{c|c|c|c|c|c|c}
\toprule
\textbf{Class} & \textbf{Images} & \textbf{Instances} & \textbf{Precision} & \textbf{Recall} & \textbf{AP50} & \textbf{AP50-95} \\
\midrule
\textit{Abutilon theophrasti} Medik. & 48 & 1166 & 77.32 ± 1.78 & 74.90 ± 1.66 & 80.14 ± 1.31 & 45.76 ± 1.76 \\ 
\textit{Elymus repens} (L.) Gould & 53 & 309 & 71.66 ± 2.00 & 72.94 ± 1.80 & 75.08 ± 0.49 & 36.34 ± 0.39 \\ 
\textit{Alopecurus myosuroides} Huds. & 77 & 635 & \textbf{90.10 ± 2.10} & 82.86 ± 0.82 & 88.26 ± 0.54 & 49.50 ± 1.38 \\ 
\textit{Amaranthus retroflexus} L. & 94 & 418 & 85.20 ± 1.96 & 58.30 ± 1.32 & 68.60 ± 1.00 & 51.28 ± 0.48 \\ 
\textit{Avena fatua} L. & 77 & 1131 & 78.88 ± 1.35 & 66.30 ± 1.04 & 73.52 ± 0.61 & 41.34 ± 0.99 \\ 
\textit{Chenopodium album} L. & 101 & 329 & 81.62 ± 3.27 & 56.34 ± 1.34 & 66.72 ± 0.54 & 40.72 ± 0.51 \\ 
\textit{Geranium spp.} & 53 & 1347 & 79.90 ± 2.33 & 57.56 ± 0.76 & 66.86 ± 2.23 & 31.86 ± 1.42 \\ 
\textit{Helianthus annuus} L. & 56 & 1873 & 80.54 ± 1.22 & 74.82 ± 0.82 & 83.00 ± 0.75 & \textbf{60.82 ± 0.61} \\ 
\textit{Lamium purpureum} L. & 214 & 1069 & 89.44 ± 1.09 & 57.18 ± 0.70 & 67.38 ± 0.86 & 39.60 ± 1.23 \\ 
\textit{Fallopia convolvulus} (L.) Á.  Löve & 78 & 115 & 76.06 ± 7.56 & 57.34 ± 1.78 & 61.82 ± 2.79 & 44.14 ± 1.15 \\ 
\textit{Setaria spp.} & 134 & 275 & 77.40 ± 1.23 & 45.32 ± 2.00 & 50.56 ± 1.11 & 30.82 ± 1.28 \\ 
\textit{Solanum nigrum} L. & 57 & 331 & 80.52 ± 2.79 & \textbf{86.64 ± 1.28} & \textbf{89.16 ± 0.82} & 54.50 ± 0.70 \\ 
\textit{Thlaspi arvense} L. & 273 & 1208 & 89.28 ± 0.41 & 51.46 ± 1.26 & 63.92 ± 0.88 & 41.86 ± 0.40 \\ 
\textit{Triticum aestivum} L. & 58 & 1772 & 75.58 ± 2.21 & 66.62 ± 0.86 & 71.88 ± 1.14 & 32.02 ± 0.83 \\ 
\textit{Veronica persica} Poir. & 108 & 322 & 83.90 ± 1.74 & 67.88 ± 2.29 & 72.22 ± 1.27 & 38.92 ± 0.75 \\ 
\textit{Zea mays} L. & 65 & 1327 & 83.56 ± 1.46 & 76.70 ± 0.63 & 84.24 ± 0.51 & 51.88 ± 0.27 \\
\bottomrule
\end{tabular}

}
\end{table}

\begin{table}[H]
\caption{Performance of YOLOv10m, Images and Instances per species, Precision, Recall, AP50 and AP50-95, with standard deviations (+/- SD) indicated. Bold values represent the highest values.}
\resizebox{\textwidth}{!}{%
\begin{tabular}{c|c|c|c|c|c|c}
\toprule
\textbf{Class} & \textbf{Images} & \textbf{Instances} & \textbf{Precision} & \textbf{Recall} & \textbf{AP50} & \textbf{AP50-95} \\
\midrule
\textit{Abutilon theophrasti} Medik. & 48 & 1166 & 76.22 ± 0.91 & 75.68 ± 0.85 & 79.94 ± 0.45 & 45.18 ± 0.71 \\ 
\textit{Elymus repens} (L.) Gould & 53 & 309 & 71.68 ± 2.55 & 71.16 ± 1.86 & 74.56 ± 1.17 & 34.72 ± 1.06 \\ 
\textit{Alopecurus myosuroides} Huds. & 77 & 635 & \textbf{90.18 ± 0.68} & \textbf{82.62 ± 0.88} & \textbf{88.32 ± 0.76} & 49.92 ± 0.83 \\ 
\textit{Amaranthus retroflexus} L. & 94 & 418 & 86.38 ± 0.97 & 57.52 ± 2.06 & 68.48 ± 1.31 & 51.76 ± 0.99 \\ 
\textit{Avena fatua} L. & 77 & 1131 & 80.10 ± 1.41 & 65.20 ± 0.90 & 73.54 ± 0.85 & 40.38 ± 0.48 \\ 
\textit{Chenopodium album} L. & 101 & 329 & 78.40 ± 2.44 & 55.96 ± 1.24 & 66.70 ± 0.92 & 40.66 ± 0.93 \\ 
\textit{Geranium spp.} & 53 & 1347 & 79.04 ± 1.54 & 57.18 ± 0.98 & 66.20 ± 1.18 & 31.90 ± 1.18 \\ 
\textit{Helianthus annuus} L. & 56 & 1873 & 80.36 ± 0.94 & 73.96 ± 0.34 & 82.62 ± 1.15 & \textbf{60.58 ± 0.93} \\ 
\textit{Lamium purpureum} L. & 214 & 1069 & 90.24 ± 1.10 & 57.26 ± 1.20 & 67.74 ± 0.90 & 40.04 ± 0.81 \\ 
\textit{Fallopia convolvulus} (L.) Á.  Löve & 78 & 115 & 71.72 ± 5.51 & 57.74 ± 1.42 & 60.32 ± 1.54 & 43.24 ± 1.94 \\ 
\textit{Setaria spp.} & 134 & 275 & 76.28 ± 4.48 & 43.24 ± 2.35 & 49.00 ± 1.52 & 29.96 ± 1.14 \\ 
\textit{Solanum nigrum} L. & 57 & 331 & 79.54 ± 1.17 & 84.92 ± 1.21 & 88.10 ± 0.35 & 53.44 ± 0.61 \\ 
\textit{Thlaspi arvense} L. & 273 & 1208 & 88.00 ± 1.42 & 52.04 ± 0.97 & 64.12 ± 1.10 & 41.56 ± 0.36 \\ 
\textit{Triticum aestivum} L. & 58 & 1772 & 73.74 ± 1.24 & 66.12 ± 1.45 & 71.86 ± 1.11 & 31.86 ± 1.11 \\ 
\textit{Veronica persica} Poir. & 108 & 322 & 83.32 ± 1.87 & 67.36 ± 1.94 & 71.78 ± 1.23 & 39.40 ± 0.78 \\ 
\textit{Zea mays} L. & 65 & 1327 & 83.58 ± 1.68 & 77.80 ± 0.83 & 84.64 ± 0.82 & 51.66 ± 0.59 \\
\bottomrule
\end{tabular}

}
\end{table}

\begin{table}[H]
\caption{Performance of YOLOv10s, Images and Instances per species, Precision, Recall, AP50 and AP50-95, with standard deviations (+/- SD) indicated. Bold values represent the highest values.}
\resizebox{\textwidth}{!}{%
\begin{tabular}{c|c|c|c|c|c|c}
\toprule
\textbf{Class} & \textbf{Images} & \textbf{Instances} & \textbf{Precision} & \textbf{Recall} & \textbf{AP50} & \textbf{AP50-95} \\
\midrule
\textit{Abutilon theophrasti} Medik. & 48 & 1166 & 78.20 ± 1.23 & 73.54 ± 1.17 & 80.20 ± 0.70 & 46.14 ± 0.66 \\ 
\textit{Elymus repens} (L.) Gould & 53 & 309 & 74.94 ± 2.04 & 72.72 ± 2.32 & 76.82 ± 0.76 & 35.82 ± 0.59 \\ 
\textit{Alopecurus myosuroides} Huds. & 77 & 635 & \textbf{92.04 ± 0.67} & 81.74 ± 1.01 & \textbf{88.20 ± 1.01} & 49.34 ± 1.24 \\ 
\textit{Amaranthus retroflexus} L. & 94 & 418 & 85.94 ± 1.42 & 57.78 ± 0.65 & 68.48 ± 1.58 & 51.70 ± 0.79 \\ 
\textit{Avena fatua} L. & 77 & 1131 & 78.62 ± 2.35 & 64.88 ± 0.90 & 72.84 ± 0.68 & 39.34 ± 0.81 \\ 
\textit{Chenopodium album} L. & 101 & 329 & 79.48 ± 1.71 & 53.58 ± 2.02 & 64.58 ± 1.11 & 39.90 ± 0.57 \\ 
\textit{Geranium spp.} & 53 & 1347 & 79.94 ± 0.78 & 57.48 ± 1.06 & 66.60 ± 1.37 & 31.92 ± 0.81 \\ 
\textit{Helianthus annuus} L. & 56 & 1873 & 78.56 ± 1.20 & 73.08 ± 1.83 & 81.04 ± 1.82 & \textbf{59.08 ± 1.42} \\ 
\textit{Lamium purpureum} L. & 214 & 1069 & 88.76 ± 1.46 & 56.88 ± 0.72 & 66.92 ± 0.69 & 39.20 ± 0.34 \\ 
\textit{Fallopia convolvulus} (L.) Á.  Löve & 78 & 115 & 73.06 ± 6.03 & 58.40 ± 1.69 & 61.36 ± 1.03 & 43.90 ± 0.78 \\ 
\textit{Setaria spp.} & 134 & 275 & 76.10 ± 3.06 & 44.88 ± 1.18 & 49.28 ± 0.68 & 29.88 ± 0.60 \\ 
\textit{Solanum nigrum} L. & 57 & 331 & 80.30 ± 4.09 & \textbf{84.36 ± 2.34} & 87.34 ± 0.86 & 53.44 ± 0.61 \\ 
\textit{Thlaspi arvense} L. & 273 & 1208 & 84.54 ± 2.17 & 50.90 ± 0.96 & 62.00 ± 0.81 & 40.06 ± 0.57 \\ 
\textit{Triticum aestivum} L. & 58 & 1772 & 75.18 ± 1.23 & 63.88 ± 1.16 & 70.68 ± 0.77 & 31.22 ± 0.41 \\ 
\textit{Veronica persica} Poir. & 108 & 322 & 81.40 ± 2.00 & 67.72 ± 1.36 & 71.32 ± 0.98 & 39.84 ± 0.57 \\ 
\textit{Zea mays} L. & 65 & 1327 & 84.98 ± 1.45 & 76.14 ± 1.14 & 85.12 ± 0.73 & 51.54 ± 0.30 \\
\bottomrule
\end{tabular}

}
\end{table}

\begin{table}[H]
\caption{Performance of YOLOv10n, Images and Instances per species, Precision, Recall, AP50 and AP50-95, with standard deviations (+/- SD) indicated. Bold values represent the highest values.}
\resizebox{\textwidth}{!}{%
\begin{tabular}{c|c|c|c|c|c|c}
\toprule
\textbf{Class} & \textbf{Images} & \textbf{Instances} & \textbf{Precision} & \textbf{Recall} & \textbf{AP50} & \textbf{AP50-95} \\
\midrule
\textit{Abutilon theophrasti} Medik. & 48 & 1166 & 78.00 ± 0.77 & 70.78 ± 0.93 & 77.88 ± 0.83 & 43.70 ± 0.52 \\ 
\textit{Elymus repens} (L.) Gould & 53 & 309 & 76.80 ± 1.56 & 71.90 ± 2.32 & 77.02 ± 1.42 & 36.82 ± 1.97 \\ 
\textit{Alopecurus myosuroides} Huds. & 77 & 635 & \textbf{91.50 ± 0.66} & 79.76 ± 0.80 & \textbf{87.12 ± 0.68} & 48.46 ± 1.32 \\ 
\textit{Amaranthus retroflexus} L. & 94 & 418 & 87.82 ± 0.90 & 56.98 ± 0.77 & 67.82 ± 0.54 & 50.20 ± 0.35 \\ 
\textit{Avena fatua} L. & 77 & 1131 & 79.62 ± 0.65 & 63.82 ± 0.88 & 71.60 ± 0.69 & 38.22 ± 0.46 \\ 
\textit{Chenopodium album} L. & 101 & 329 & 78.88 ± 1.36 & 52.46 ± 2.09 & 62.36 ± 0.80 & 37.40 ± 1.08 \\ 
\textit{Geranium spp.} & 53 & 1347 & 81.06 ± 1.35 & 54.90 ± 0.98 & 65.68 ± 1.72 & 31.86 ± 0.88 \\ 
\textit{Helianthus annuus} L. & 56 & 1873 & 77.62 ± 1.08 & 71.58 ± 0.24 & 79.08 ± 0.74 & \textbf{57.38 ± 0.58} \\ 
\textit{Lamium purpureum} L. & 214 & 1069 & 87.20 ± 0.60 & 56.26 ± 0.62 & 64.96 ± 0.19 & 38.92 ± 0.35 \\ 
\textit{Fallopia convolvulus} (L.) Á.  Löve & 78 & 115 & 69.82 ± 3.63 & 59.44 ± 1.52 & 61.24 ± 1.75 & 44.86 ± 1.67 \\ 
\textit{Setaria spp.} & 134 & 275 & 71.84 ± 1.70 & 45.24 ± 1.64 & 48.60 ± 0.79 & 29.18 ± 0.58 \\ 
\textit{Solanum nigrum} L. & 57 & 331 & 80.14 ± 1.43 & \textbf{84.36 ± 1.40} & 86.34 ± 0.55 & 53.40 ± 0.66 \\ 
\textit{Thlaspi arvense} L. & 273 & 1208 & 84.74 ± 0.84 & 48.98 ± 0.70 & 60.10 ± 0.55 & 38.56 ± 0.38 \\ 
\textit{Triticum aestivum} L. & 58 & 1772 & 75.62 ± 0.68 & 62.24 ± 1.56 & 70.20 ± 0.27 & 30.60 ± 0.43 \\ 
\textit{Veronica persica} Poir. & 108 & 322 & 77.92 ± 1.26 & 67.12 ± 1.25 & 70.16 ± 0.83 & 39.64 ± 0.88 \\ 
\textit{Zea mays} L. & 65 & 1327 & 82.92 ± 1.14 & 74.22 ± 0.89 & 83.08 ± 0.66 & 50.18 ± 0.46 \\
\bottomrule
\end{tabular}

}
\end{table}

\subsubsection{Performance of RT-DETR}

\begin{table}[H]
\caption{Performance of RT-DETR-x, Images and Instances per species, Precision, Recall, AP50 and AP50-95, with standard deviations (+/- SD) indicated. Bold values represent the highest values.}
\resizebox{\textwidth}{!}{%
\begin{tabular}{c|c|c|c|c|c|c}
\toprule
\textbf{Class} & \textbf{Images} & \textbf{Instances} & \textbf{Precision} & \textbf{Recall} & \textbf{AP50} & \textbf{AP50-95} \\
\midrule
\textit{Abutilon theophrasti} Medik. & 48 & 1166 & 76.44 ± 2.00 & 77.78 ± 1.60 & 77.94 ± 1.86 & 45.62 ± 1.24 \\ 
\textit{Elymus repens} (L.) Gould & 53 & 309 & 76.14 ± 3.79 & 72.40 ± 1.37 & 75.74 ± 1.60 & 34.66 ± 1.16 \\ 
\textit{Alopecurus myosuroides} Huds. & 77 & 635 & 90.24 ± 1.40 & \textbf{81.30 ± 1.49} & 86.18 ± 1.44 & 48.42 ± 1.70 \\ 
\textit{Amaranthus retroflexus} L. & 94 & 418 & 85.24 ± 2.34 & 58.62 ± 0.75 & 68.46 ± 0.99 & 50.04 ± 0.53 \\ 
\textit{Avena fatua} L. & 77 & 1131 & 77.98 ± 1.56 & 66.04 ± 1.42 & 71.12 ± 1.63 & 39.60 ± 1.08 \\ 
\textit{Chenopodium album} L. & 101 & 329 & 82.00 ± 2.47 & 57.92 ± 2.88 & 67.00 ± 0.31 & 40.70 ± 0.75 \\ 
\textit{Geranium spp.} & 53 & 1347 & 80.18 ± 1.63 & 61.06 ± 1.24 & 66.96 ± 2.22 & 30.60 ± 1.12 \\ 
\textit{Helianthus annuus} L. & 56 & 1873 & 80.66 ± 1.40 & 74.86 ± 1.18 & 79.30 ± 1.90 & \textbf{57.04 ± 1.46} \\ 
\textit{Lamium purpureum} L. & 214 & 1069 & \textbf{90.56 ± 0.98} & 58.46 ± 2.01 & 65.02 ± 1.04 & 37.18 ± 1.16 \\ 
\textit{Fallopia convolvulus} (L.) Á.  Löve & 78 & 115 & 75.56 ± 1.35 & 60.92 ± 1.53 & 62.32 ± 1.13 & 48.00 ± 1.27 \\ 
\textit{Setaria spp.} & 134 & 275 & 80.16 ± 2.51 & 44.80 ± 2.53 & 48.00 ± 2.30 & 28.30 ± 1.18 \\ 
\textit{Solanum nigrum} L. & 57 & 331 & 79.92 ± 3.26 & 84.54 ± 0.75 & \textbf{86.68 ± 1.36} & 52.96 ± 1.93 \\ 
\textit{Thlaspi arvense} L. & 273 & 1208 & 87.74 ± 1.84 & 49.32 ± 1.75 & 59.34 ± 1.63 & 37.84 ± 0.88 \\ 
\textit{Triticum aestivum} L. & 58 & 1772 & 76.28 ± 2.30 & 68.20 ± 0.98 & 69.24 ± 1.46 & 30.50 ± 0.99 \\ 
\textit{Veronica persica} Poir. & 108 & 322 & 88.72 ± 1.68 & 68.14 ± 1.88 & 72.32 ± 0.49 & 39.40 ± 0.87 \\ 
\textit{Zea mays} L. & 65 & 1327 & 85.08 ± 0.91 & 77.26 ± 1.24 & 81.54 ± 1.26 & 49.30 ± 0.64 \\
\bottomrule
\end{tabular}

}
\end{table}

\subsubsection{Dataset 2}
\subsubsection{Performance of YOLOv8}

\begin{table}[H]
\caption{Performance of YOLOv8x, Images and Instances per species, Precision, Recall, AP50 and AP50-95, with standard deviations (+/- SD) indicated. Bold values represent the highest values.}
\resizebox{\textwidth}{!}{%
\begin{tabular}{c|c|c|c|c|c|c}
\toprule
\textbf{Class} & \textbf{Images} & \textbf{Instances} & \textbf{Precision} & \textbf{Recall} & \textbf{AP50} & \textbf{AP50-95} \\
\midrule
Dicot & 590 & 6305 & \textbf{84.54 ± 0.48} & 66.28 ± 0.46 & 76.64 ± 0.75 & 44.18 ± 0.65 \\ 
\textit{Helianthus annuus} L. & 56 & 1873 & 80.18 ± 1.25 & 76.62 ± 1.13 & 83.36 ± 0.95 & \textbf{60.72 ± 0.94} \\ 
Monocot & 312 & 2350 & 81.42 ± 0.52 & 70.78 ± 0.52 & 77.04 ± 0.81 & 42.30 ± 0.51 \\ 
\textit{Triticum aestivum} L. & 58 & 1772 & 74.18 ± 1.80 & 68.60 ± 1.22 & 72.30 ± 1.65 & 32.02 ± 1.41 \\ 
\textit{Zea mays} L. & 65 & 1327 & 82.26 ± 0.86 & \textbf{76.90 ± 0.73} & \textbf{84.32 ± 0.58} & 51.44 ± 0.64 \\
\bottomrule
\end{tabular}

}
\end{table}

\begin{table}[H]
\caption{Performance of YOLOv8m, Images and Instances per species, Precision, Recall, AP50 and AP50-95, with standard deviations (+/- SD) indicated. Bold values represent the highest values.}
\resizebox{\textwidth}{!}{%
\begin{tabular}{c|c|c|c|c|c|c}
\toprule
\textbf{Class} & \textbf{Images} & \textbf{Instances} & \textbf{Precision} & \textbf{Recall} & \textbf{AP50} & \textbf{AP50-95} \\
\midrule
Dicot & 590 & 6305 & \textbf{84.74 ± 0.63} & 65.86 ± 1.34 & 77.00 ± 0.77 & 44.30 ± 0.42 \\ 
\textit{Helianthus annuus} L. & 56 & 1873 & 78.38 ± 1.03 & 76.28 ± 1.15 & \textbf{84.28 ± 0.92} & \textbf{61.62 ± 0.89} \\ 
Monocot & 312 & 2350 & 79.98 ± 0.66 & 69.52 ± 0.98 & 76.36 ± 0.83 & 41.22 ± 0.68 \\ 
\textit{Triticum aestivum} L. & 58 & 1772 & 74.24 ± 1.05 & 69.02 ± 0.93 & 73.04 ± 1.27 & 32.26 ± 1.31 \\ 
\textit{Zea mays} L. & 65 & 1327 & 82.78 ± 1.17 & \textbf{77.06 ± 1.33} & 83.90 ± 0.89 & 51.34 ± 1.19 \\
\bottomrule
\end{tabular}

}
\end{table}

\begin{table}[H]
\caption{Performance of YOLOv8s, Images and Instances per species, Precision, Recall, AP50 and AP50-95, with standard deviations (+/- SD) indicated. Bold values represent the highest values.}
\resizebox{\textwidth}{!}{%
\begin{tabular}{c|c|c|c|c|c|c}
\toprule
\textbf{Class} & \textbf{Images} & \textbf{Instances} & \textbf{Precision} & \textbf{Recall} & \textbf{AP50} & \textbf{AP50-95} \\
\midrule
Dicot & 590 & 6305 & \textbf{85.28 ± 0.93} & 66.18 ± 1.55 & 77.42 ± 0.65 & 44.28 ± 0.41 \\ 
\textit{Helianthus annuus} L. & 56 & 1873 & 78.82 ± 2.14 & 76.24 ± 0.76 & 84.00 ± 0.80 & \textbf{60.88 ± 0.65} \\ 
Monocot & 312 & 2350 & 81.18 ± 1.85 & 69.72 ± 1.29 & 77.12 ± 0.39 & 41.68 ± 0.47 \\ 
\textit{Triticum aestivum} L. & 58 & 1772 & 74.40 ± 1.43 & 68.32 ± 0.99 & 72.80 ± 0.56 & 32.32 ± 0.47 \\ 
\textit{Zea mays} L. & 65 & 1327 & 84.26 ± 1.06 & \textbf{78.10 ± 0.97} & \textbf{85.40 ± 0.41} & 51.90 ± 0.48 \\
\bottomrule
\end{tabular}

}
\end{table}

\begin{table}[H]
\caption{Performance of YOLOv8n, Images and Instances per species, Precision, Recall, AP50 and AP50-95, with standard deviations (+/- SD) indicated. Bold values represent the highest values.}
\resizebox{\textwidth}{!}{%
\begin{tabular}{c|c|c|c|c|c|c}
\toprule
\textbf{Class} & \textbf{Images} & \textbf{Instances} & \textbf{Precision} & \textbf{Recall} & \textbf{AP50} & \textbf{AP50-95} \\
\midrule
Dicot & 590 & 6305 & \textbf{84.52 ± 0.53} & 66.28 ± 0.76 & 76.76 ± 0.49 & 44.32 ± 0.30 \\ 
\textit{Helianthus annuus} L. & 56 & 1873 & 77.04 ± 1.62 & 76.36 ± 0.60 & 83.68 ± 0.69 & \textbf{60.62 ± 0.74} \\ 
Monocot & 312 & 2350 & 80.74 ± 1.08 & 70.70 ± 0.77 & 77.32 ± 0.41 & 41.30 ± 0.17 \\ 
\textit{Triticum aestivum} L. & 58 & 1772 & 73.30 ± 1.56 & 66.78 ± 0.95 & 71.98 ± 1.82 & 31.58 ± 0.89 \\ 
\textit{Zea mays} L. & 65 & 1327 & 82.52 ± 0.92 & \textbf{77.80 ± 0.82} & \textbf{84.82 ± 0.83} & 51.84 ± 0.63 \\
\bottomrule
\end{tabular}

}
\end{table}

\subsubsection{Performance of YOLOv9}

\begin{table}[H]
\caption{Performance of YOLOv9e, Images and Instances per species, Precision, Recall, AP50 and AP50-95, with standard deviations (+/- SD) indicated. Bold values represent the highest values.}
\resizebox{\textwidth}{!}{%
\begin{tabular}{c|c|c|c|c|c|c}
\toprule
\textbf{Class} & \textbf{Images} & \textbf{Instances} & \textbf{Precision} & \textbf{Recall} & \textbf{AP50} & \textbf{AP50-95} \\
\midrule
Dicot & 590 & 6305 & \textbf{84.66 ± 67.44} & 67.44 ± 0.93 & 78.56 ± 0.46 & 45.82 ± 0.26 \\ 
\textit{Helianthus annuus} L. & 56 & 1873 & 77.68 ± 76.78 & 76.78 ± 0.72 & 84.30 ± 1.09 & \textbf{61.52 ± 0.84} \\ 
Monocot & 312 & 2350 & 80.94 ± 70.20 & 70.20 ± 0.98 & 78.10 ± 0.60 & 43.92 ± 0.27 \\ 
\textit{Triticum aestivum} L. & 58 & 1772 & 73.08 ± 68.34 & 68.34 ± 1.74 & 73.18 ± 1.44 & 32.78 ± 1.02 \\ 
\textit{Zea mays} L. & 65 & 1327 & 82.56 ± 77.88 & \textbf{77.88 ± 1.38} & \textbf{84.60 ± 0.68} & 52.22 ± 1.02 \\
\bottomrule
\end{tabular}

}
\end{table}

\begin{table}[H]
\caption{Performance of YOLOv9m, Images and Instances per species, Precision, Recall, AP50 and AP50-95, with standard deviations (+/- SD) indicated. Bold values represent the highest values.}
\resizebox{\textwidth}{!}{%
\begin{tabular}{c|c|c|c|c|c|c}
\toprule
\textbf{Class} & \textbf{Images} & \textbf{Instances} & \textbf{Precision} & \textbf{Recall} & \textbf{AP50} & \textbf{AP50-95} \\
\midrule
Dicot & 590 & 6305 & \textbf{84.70 ± 0.43} & 66.68 ± 0.60 & 77.68 ± 0.74 & 45.08 ± 0.29 \\ 
\textit{Helianthus annuus} L. & 56 & 1873 & 77.26 ± 1.66 & 76.06 ± 1.54 & 83.78 ± 1.11 & \textbf{61.16 ± 1.16} \\ 
Monocot & 312 & 2350 & 81.30 ± 1.68 & 70.04 ± 0.70 & 77.86 ± 1.14 & 42.56 ± 0.64 \\ 
\textit{Triticum aestivum} L. & 58 & 1772 & 74.42 ± 1.12 & 68.08 ± 0.92 & 73.12 ± 1.47 & 32.86 ± 0.76 \\ 
\textit{Zea mays} L. & 65 & 1327 & 81.92 ± 1.43 & \textbf{77.76 ± 0.46} & \textbf{84.52 ± 0.37} & 51.36 ± 0.61 \\
\bottomrule
\end{tabular}

}
\end{table}

\begin{table}[H]
\caption{Performance of YOLOv9s, Images and Instances per species, Precision, Recall, AP50 and AP50-95, with standard deviations (+/- SD) indicated. Bold values represent the highest values.}
\resizebox{\textwidth}{!}{%
\begin{tabular}{c|c|c|c|c|c|c}
\toprule
\textbf{Class} & \textbf{Images} & \textbf{Instances} & \textbf{Precision} & \textbf{Recall} & \textbf{AP50} & \textbf{AP50-95} \\
\midrule
Dicot & 590 & 6305 & \textbf{84.88 ± 0.62} & 67.36 ± 0.55 & 77.76 ± 0.30 & 45.28 ± 0.30 \\ 
\textit{Helianthus annuus} L. & 56 & 1873 & 77.92 ± 1.63 & 75.66 ± 1.29 & 83.66 ± 1.59 & \textbf{61.22 ± 1.24} \\ 
Monocot & 312 & 2350 & 82.36 ± 0.84 & 71.14 ± 0.82 & 78.70 ± 0.33 & 43.54 ± 0.33 \\ 
\textit{Triticum aestivum} L. & 58 & 1772 & 74.64 ± 0.86 & 68.84 ± 0.69 & 74.04 ± 0.27 & 33.00 ± 0.41 \\ 
\textit{Zea mays} L. & 65 & 1327 & 83.30 ± 0.76 & \textbf{78.82 ± 0.58} & \textbf{85.10 ± 0.32} & 52.02 ± 0.31 \\
\bottomrule
\end{tabular}

}
\end{table}

\begin{table}[H]
\caption{Performance of YOLOv9t, Images and Instances per species, Precision, Recall, AP50 and AP50-95, with standard deviations (+/- SD) indicated. Bold values represent the highest values.}
\resizebox{\textwidth}{!}{%
\begin{tabular}{c|c|c|c|c|c|c}
\toprule
\textbf{Class} & \textbf{Images} & \textbf{Instances} & \textbf{Precision} & \textbf{Recall} & \textbf{AP50} & \textbf{AP50-95} \\
\midrule
Dicot & 590 & 6305 & \textbf{85.66 ± 0.68} & 65.40 ± 1.23 & 76.32 ± 0.50 & 44.76 ± 0.27 \\ 
\textit{Helianthus annuus} L. & 56 & 1873 & 77.20 ± 1.88 & 75.16 ± 0.42 & 83.58 ± 0.72 & \textbf{61.00 ± 0.71} \\ 
Monocot & 312 & 2350 & 82.72 ± 0.92 & 71.04 ± 0.75 & 78.38 ± 0.40 & 43.24 ± 0.26 \\ 
\textit{Triticum aestivum} L. & 58 & 1772 & 74.68 ± 0.90 & 68.36 ± 0.55 & 73.84 ± 0.83 & 33.20 ± 0.41 \\ 
\textit{Zea mays} L. & 65 & 1327 & 83.94 ± 0.72 & \textbf{78.18 ± 0.98} & \textbf{85.64 ± 0.54} & 52.34 ± 0.32 \\
\bottomrule
\end{tabular}

}
\end{table}

\subsubsection{Performance of YOLOv10}

\begin{table}[H]
\caption{Performance of YOLOv10x, Images and Instances per species, Precision, Recall, AP50 and AP50-95, with standard deviations (+/- SD) indicated. Bold values represent the highest values.}
\resizebox{\textwidth}{!}{%
\begin{tabular}{c|c|c|c|c|c|c}
\toprule
\textbf{Class} & \textbf{Images} & \textbf{Instances} & \textbf{Precision} & \textbf{Recall} & \textbf{AP50} & \textbf{AP50-95} \\
\midrule
Dicot & 590 & 6305 & \textbf{84.60 ± 0.91} & 64.80 ± 0.99 & 76.08 ± 0.83 & 44.04 ± 0.56 \\ 
\textit{Helianthus annuus} L. & 56 & 1873 & 77.90 ± 1.20 & 74.46 ± 1.99 & 80.88 ± 1.70 & \textbf{59.28 ± 1.22} \\ 
Monocot & 312 & 2350 & 81.48 ± 1.40 & 69.66 ± 0.45 & 76.68 ± 0.47 & 41.84 ± 0.44 \\ 
\textit{Triticum aestivum} L. & 58 & 1772 & 75.42 ± 1.50 & 67.04 ± 0.72 & 71.62 ± 0.77 & 31.44 ± 0.53 \\ 
\textit{Zea mays} L. & 65 & 1327 & 82.88 ± 0.95 & \textbf{76.06 ± 1.43} & \textbf{83.78 ± 0.26} & 51.54 ± 0.59 \\

\bottomrule
\end{tabular}

}
\end{table}

\begin{table}[H]
\caption{Performance of YOLOv10m, Images and Instances per species, Precision, Recall, AP50 and AP50-95, with standard deviations (+/- SD) indicated. Bold values represent the highest values.}
\resizebox{\textwidth}{!}{%
\begin{tabular}{c|c|c|c|c|c|c}
\toprule
\textbf{Class} & \textbf{Images} & \textbf{Instances} & \textbf{Precision} & \textbf{Recall} & \textbf{AP50} & \textbf{AP50-95} \\
\midrule
Dicot & 590 & 6305 & \textbf{85.02 ± 0.99} & 65.98 ± 0.99 & 77.36 ± 0.59 & 45.22 ± 0.42 \\ 
\textit{Helianthus annuus} L. & 56 & 1873 & 77.86 ± 0.88 & 74.78 ± 0.78 & 82.50 ± 0.61 & \textbf{60.60 ± 0.25} \\ 
Monocot & 312 & 2350 & 80.70 ± 1.41 & 69.30 ± 0.85 & 76.16 ± 0.48 & 41.70 ± 0.45 \\ 
\textit{Triticum aestivum} L. & 58 & 1772 & 74.64 ± 1.25 & 65.54 ± 1.31 & 70.68 ± 1.13 & 31.24 ± 0.66 \\ 
\textit{Zea mays} L. & 65 & 1327 & 82.86 ± 1.68 & \textbf{75.34 ± 2.76} & \textbf{83.52 ± 1.20} & 51.50 ± 0.53 \\
\bottomrule
\end{tabular}

}
\end{table}

\begin{table}[H]
\caption{Performance of YOLOv10s, Images and Instances per species, Precision, Recall, AP50 and AP50-95, with standard deviations (+/- SD) indicated. Bold values represent the highest values.}
\resizebox{\textwidth}{!}{%
\begin{tabular}{c|c|c|c|c|c|c}
\toprule
\textbf{Class} & \textbf{Images} & \textbf{Instances} & \textbf{Precision} & \textbf{Recall} & \textbf{AP50} & \textbf{AP50-95} \\
\midrule
Dicot & 590 & 6305 & \textbf{84.74 ± 0.94} & 63.68 ± 0.59 & 75.16 ± 0.64 & 43.94 ± 0.34 \\ 
\textit{Helianthus annuus} L. & 56 & 1873 & 77.00 ± 1.40 & 73.76 ± 0.69 & 82.10 ± 0.47 & \textbf{60.18 ± 0.28} \\ 
Monocot & 312 & 2350 & 81.40 ± 1.50 & 69.28 ± 0.39 & 75.82 ± 0.62 & 41.08 ± 0.43 \\ 
\textit{Triticum aestivum} L. & 58 & 1772 & 74.62 ± 0.58 & 65.32 ± 0.92 & 70.44 ± 0.27 & 31.06 ± 0.46 \\ 
\textit{Zea mays} L. & 65 & 1327 & 83.48 ± 0.75 & \textbf{77.04 ± 0.76} & \textbf{84.54 ± 0.59} & 51.40 ± 0.19 \\
\bottomrule
\end{tabular}

}
\end{table}

\begin{table}[H]
\caption{Performance of YOLOv10n, Images and Instances per species, Precision, Recall, AP50 and AP50-95, with standard deviations (+/- SD) indicated. Bold values represent the highest values.}
\resizebox{\textwidth}{!}{%
\begin{tabular}{c|c|c|c|c|c|c}
\toprule
\textbf{Class} & \textbf{Images} & \textbf{Instances} & \textbf{Precision} & \textbf{Recall} & \textbf{AP50} & \textbf{AP50-95} \\
\midrule
Dicot & 590 & 6305 & \textbf{84.50 ± 0.81} & 63.06 ± 0.44 & 74.58 ± 0.27 & 43.50 ± 0.17 \\ 
\textit{Helianthus annuus} L. & 56 & 1873 & 76.12 ± 1.68 & 72.96 ± 0.82 & 79.64 ± 0.80 & \textbf{57.68 ± 0.59} \\ 
Monocot & 312 & 2350 & 81.82 ± 0.88 & 68.78 ± 0.82 & 76.14 ± 0.61 & 41.24 ± 0.25 \\ 
\textit{Triticum aestivum} L. & 58 & 1772 & 74.90 ± 1.90 & 64.58 ± 0.18 & 71.20 ± 1.12 & 31.32 ± 0.86 \\ 
\textit{Zea mays} L. & 65 & 1327 & 82.74 ± 1.45 & \textbf{76.16 ± 0.72} & \textbf{84.10 ± 0.82} & 50.98 ± 0.15 \\
\bottomrule
\end{tabular}

}
\end{table}

\subsubsection{Performance of RT-DETR}

\begin{table}[H]
\caption{Performance of RT-DETR-x, Images and Instances per species, Precision, Recall, AP50 and AP50-95, with standard deviations (+/- SD) indicated. Bold values represent the highest values.}
\resizebox{\textwidth}{!}{%
\begin{tabular}{c|c|c|c|c|c|c}
\toprule
\textbf{Class} & \textbf{Images} & \textbf{Instances} & \textbf{Precision} & \textbf{Recall} & \textbf{AP50} & \textbf{AP50-95} \\
\midrule
Dicot & 590 & 6305 & 84.22 ± 0.48 & 67.46 ± 0.86 & 76.60 ± 1.66 & 43.42 ± 0.85 \\ 
\textit{Helianthus annuus} L. & 56 & 1873 & 79.42 ± 1.52 & 75.14 ± 1.05 & 78.34 ± 1.11 & \textbf{56.38 ± 0.48} \\ 
Monocot & 312 & 2350 & 80.80 ± 1.34 & 71.36 ± 1.05 & 75.52 ± 1.12 & 40.88 ± 0.54 \\ 
\textit{Triticum aestivum} L. & 58 & 1772 & 76.58 ± 1.98 & 68.76 ± 2.00 & 69.44 ± 3.31 & 30.36 ± 1.84 \\ 
\textit{Zea mays} L. & 65 & 1327 & \textbf{84.74 ± 1.39} & \textbf{77.88 ± 1.07} & \textbf{82.08 ± 0.72} & 49.76 ± 0.50 \\
\bottomrule
\end{tabular}

}
\end{table}

\end{document}